\documentclass{osa-article}
\journal{oe}
\articletype{Research Article}
\DeclareMathAlphabet{\mathcal}{OMS}{cmsy}{m}{n}
\usepackage{amsmath, bbm, breqn}
\usepackage{graphicx}
\usepackage[export]{adjustbox}
\usepackage{subcaption, caption, makecell, multirow}
\usepackage{float}
\usepackage{textcomp}
\usepackage{xcolor, soul}
\usepackage{changes, nameref}
\usepackage{txfonts}
\usepackage{hyperref}

\begin{document}

    \title{SiSPRNet: End-to-End Learning for Single-Shot Phase Retrieval}
    
    \author{Qiuliang Ye\authormark{1}\authormark{*},  Li-Wen Wang\authormark{1}, and Daniel P.K. Lun \authormark{1, 2}\authormark{*}}
    
    \address{ \authormark{1} Department of Electronic and Information Engineering, The Hong Kong Polytechnic University, Hong Kong SAR, China\\
    \authormark{2} The Centre for Advances in Reliability and Safety (CAiRS), Hong Kong SAR, China}
    \email{\authormark{*}qiu-liang.ye@connect.polyu.hk;  pak.kong.lun@polyu.edu.hk}

        \begin{abstract}
    	With the success of deep learning methods in many image processing tasks, deep learning approaches have also been introduced to the phase retrieval problem recently. These approaches are different from the traditional iterative optimization methods in that they usually require only one intensity measurement and can reconstruct phase images in real-time. However, because of tremendous domain discrepancy, the quality of the reconstructed images given by these approaches still has much room to improve to meet the general application requirements. In this paper, we design a novel deep neural network structure named \textit{SiSPRNet} for phase retrieval based on a single Fourier intensity measurement. To effectively utilize the spectral information of the measurements, we propose a new feature extraction unit using the Multi-Layer Perceptron (MLP) as the front end. It allows all pixels of the input intensity image to be considered together for exploring their global representation. The size of the MLP is carefully designed to facilitate the extraction of the representative features while reducing noises and outliers. A dropout layer is also equipped to mitigate the possible overfitting problem in training the MLP. To promote the global correlation in the reconstructed images, a self-attention mechanism is introduced to the Up-sampling and Reconstruction (UR) blocks of the proposed \textit{SiSPRNet}. These UR blocks are inserted into a residual learning structure to prevent the weak information flow and vanishing gradient problems due to their complex layer structure. Extensive evaluations of the proposed model are performed using different testing datasets of phase-only images and images with linearly related magnitude and phase. Experiments were conducted on an optical experimentation platform (with defocusing to reduce the saturation problem) to understand the performance of different deep learning methods when working in a practical environment. The results demonstrate that the proposed approach consistently outperforms other deep learning methods in single-shot maskless phase retrieval. The source codes of the proposed method have been released in Github \cite{Ye_SiSPRNet}.
    \end{abstract}

        \section{Introduction}
    Phase retrieval aims to reconstruct a complex-valued signal from its intensity-only measurements. It is a crucial problem in crystallography, optical imaging, astronomy, X-ray, electronic imaging, etc., because most existing measurement systems can only detect the relative low-frequency magnitude information of the wave fields. The study of the phase retrieval problems originated in the $1970$s. The researchers of the optics community developed many reconstruction algorithms \cite{Gerchberg1972APA, Fienup:82, rodenburg2008ptychography}. 
    
    Mathematically, the phase retrieval problem can be described by the following equation:
    \begin{equation}
    	\label{CDP}
    	\begin{aligned}
    		\text { Find } \mathbf{x} \in \mathbb{C}^{N} \quad \text { s.t. } \ \mathbf{Y}_{i}=\left|\mathcal{F}\left(\mathbf{I}_{i} \circ \mathbf{x}\right)\right|^{2}, i=1, \ldots, M,
    	\end{aligned}
    \end{equation}
    where $\mathbf{x}\in \mathbb{C}^N$ is the complex-valued signal of interest, and $ \mathbf{Y}$ is its Fourier intensity measurements; $\circ $ and $\mathcal{F}$ refer to elementwise multiplication and Fourier transform operator, respectively. The pre-defined optical masks $ \mathbf{I} $ provide the constraints to reduce the ill-posedness of the problem. It is implemented in many different ways. For instance, early-stage phase retrieval algorithms treat the support of the signal as the optical mask \cite{Fienup:82}. Specifically, $|\mathbf{x}_i| \neq 0$ when $i \in \mathcal{S}$, where $\mathcal{S}$ denotes the support of the signal. However, these early-stage algorithms cannot guarantee the convergence of the optimization process nor provide globally optimal solutions. Recently, pre-defined random optical masks have been used as the constraints to improve the reconstruction performance \cite{Anand07, Horisaki2014SingleshotPI, candes2015phase, YE2022106808}. The random masks can be implemented using a spatial light modulator (SLM) or digital micromirror device (DMD) \cite{Horisaki2014SingleshotPI, Zheng2017DigitalMD}. Although the usage of the random masks can lead to better reconstruction performance, there are several disadvantages. The cost of DMD and SLM is one concern, the error due to the global gray-scale phase mismatch and spatial non-uniformity from the devices is another. Besides, it was empirically shown that about $ 4 - 6 $ measurements are required for exact recovery with random binary masks \cite{candes2015code, candes2015phase}. It increases the measurement time and can affect the quality of the reconstructed image, particularly for dynamic objects.
    
    Recently, data-driven methods, like deep neural networks, have been widely applied to explore the data distributions through extensive training samples \cite{LeCun_2015}. Among them, the so-called Convolutional Neural Network (CNN) that exploits the features based on convolution kernels has been successfully adopted in image processing tasks, like image restoration \cite{He_2016_CVPR, zhang2017beyond, wang2020lightening}. Besides, an important technique named attention mechanism that mimics cognitive attention became popular in the CNN structure to enhance some parts of the feature maps and thus improve the performance \cite{attention_17}. Since CNN can learn discriminative representations from large training datasets, researchers have spent effort in using CNN for solving phase retrieval problems \cite{deng_li_goy_kang_barbastathis_2020, White:19, Shi:19, Sinha17}. For instance, Chen et al. reconstructed signals from Fresnel diffraction patterns with CNN and the contrast transfer function \cite{Bai:19}. Kumar et al. proposed the deep unrolling networks that solved the iterative phase retrieval problem with CNN \cite{Kumar:21}. Uelwer et al. proposed a cascaded neural network for phase retrieval \cite{Uelwer_PhaseRetrieval}. Besides, the conditional generative adversarial network (GAN), a kind of data-driven generative method based on specific conditions \cite{cond_gan_14}, was adopted for reconstructing the phase information from Fourier intensity measurements \cite{uelwer2021phase}. Wu et al. also proposed a CNN to reconstruct the crystal structures with coherent X-ray diffraction patterns \cite{Wu_cw5029, Wu_2021}. The above  methods adopt supervised learning that trains the networks with the paired ground-truth data. In recent years, training with unpaired or unsupervised (no ground-truth data) datasets has been a popular topic. For example, Fus et al. proposed the unsupervised learning for in-line holography phase retrieval \cite{Fus_18}, while Zhang et al. developed the GAN-based phase retrieval with unpaired datasets \cite{Zhang_21}. While the above approaches have achieved some success, phase retrieval with Fraunhofer diffraction patterns (i.e., Fourier intensity) for general applications is still a challenge. It is partly because of the large domain discrepancy between the Fourier plane and image plane; the large variation in data distributions of general applications also introduces much difficulty in designing a model structure that is optimal for different applications.  In this paper, we propose a single-shot maskless phase retrieval method named \textit{SiSPRNet} that is benefited from the advanced deep learning technology. Similar to the existing deep learning approaches, the proposed SiSPRNet only requires a single intensity measurement for each reconstruction. It also does not require extra optical masks to impose constraints on the intensity measurements. The main novelties of SiSPRNet are two folded. First, it contains a new feature extraction unit using the Multi-Layer Perceptron (MLP) as the front end. Traditional deep learning phase retrieval approaches often use a CNN as the front end to extract the features from the Fourier intensity measurement. However, the convolution operation of CNN can only explore a small area of the intensity image at one time. It does not match with the property of the Fourier intensity images where the data are globally correlated. The proposed feature extraction unit starts with an MLP block that consists of three Fully Connected (FC) layers. They allow all pixels of the input intensity image to be considered together for exploring their global correlation. These FC layers have a size smaller than the input image, which can guide the backpropagation process to train the MLP to generate useful features of the reconstructed image while ignoring the less important information, such as noises and outliers. Such a design also reduces the complexity of the network. The proposed feature extraction unit is equipped with a Dropout layer in between the FC layers that mitigates the overfitting problem, which is common to MLP having many learnable parameters. Another novelty of SiSPRNet is that it is equipped with a new self-attention-based phase reconstruction unit to generate the required images from the extracted features. Traditional deep learning phase retrieval methods reconstruct the magnitude and phase images from the extracted features without considering the global correlation of the underlying objects. Most phase retrieval methods are applied to recover the structure of physical objects, which are often structural with global correlation. The proposed phase reconstruction unit is equipped with two UR blocks, in which two self-attention units are introduced to explore the global correlation of the feature maps. They are inserted into a residual learning structure to prevent the weak information flow and vanishing gradient problems due to the complex layer structure of the UR block. Compared with the traditional convolution operation that only considers local features, the self-attention plus residual learning mechanism help improve the phase retrieval performance.
    
    The proposed \textit{SiSPRNet} is designed to work with general image datasets without assuming a specific application domain. We verified the proposed method on an optical platform to show its practicality. It has demonstrated its generality by achieving state-of-the-art performance on three different datasets of phase-only images and images with linearly related magnitude and phase. It significantly outperforms the existing deep learning Fourier phase retrieval methods. The source codes of the proposed method have been released in Github \cite{Ye_SiSPRNet}.
    
    The rest of this paper is organized as follows. Section \ref{Sec:Method} introduces the defocused phase retrieval system we developed in this study and the proposed end-to-end deep learning phase retrieval network \textit{SiSPRNet}. Section \ref{Sec:simulation} presents the model analysis and simulation results. Section \ref{Sec:experiment} provides the experimental results on an optical system for validating the performance of the proposed method.

    \section{Proposed Method} 
    \label{Sec:Method}
    
    \subsection{Phase Retrieval System} \label{Sec:PRsystem}
    
\begin{figure}[htb]               
    	\centering\includegraphics[width=0.8\linewidth]{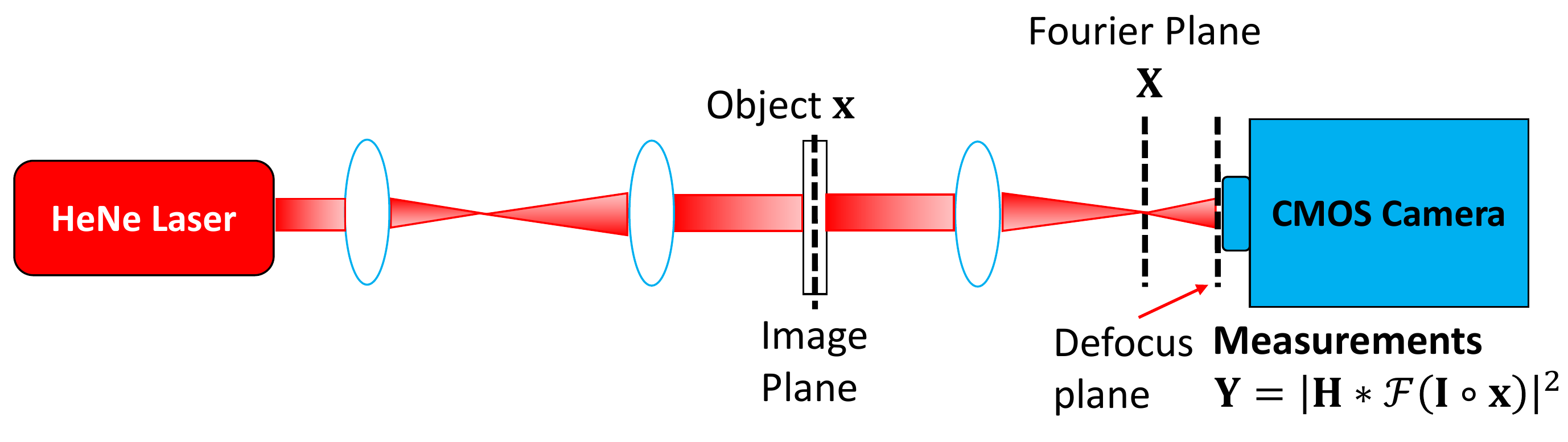}
    	\caption{Optical path of the defocus phase retrieval system. $\mathbf{x}$ represents the complex-valued object, and $\mathbf{Y}$ is its Fourier intensity measurement; $\mathbf{X} = \mathcal{F}\left(\mathbf{I}_{i} \circ \mathbf{x}\right)$,  $\mathbf{H}$ and $\ast$ denote the Fourier plane in complex-valued form, the defocus function generated via Fresnel diffraction and the convolution operation, respectively. }
    	\label{fig:opticalpath}
    	
    \end{figure}

    As indicated in Eq. \eqref{CDP}, a phase retrieval system aims to reconstruct a complex-valued signal $ \mathbf{x} $ from its Fourier intensity measurements. The proposed method does not require multiple measurements or optical masks, so $ M=1 $ and all the values of the mask $ \mathbf{I} $ are equal to one in Eq. \eqref{CDP}. In this case, the original complex-valued signal $ \mathbf{x} $ is reconstructed only from its Fourier intensity. This seemingly impossible task is known to have a solution, although with ambiguities. From \cite{hayes1982reconstruction}, it is known that $ \mathbf{x} $ can be uniquely defined, except for trivial ambiguities, by its Fourier intensity, with an oversampling factor over $ 2 $ in each dimension, if $ \mathbf{x} $ has a finite support and is non-symmetric. The above has an important implication to the optical system required for the proposed phase retrieval method. Fig. \ref{fig:opticalpath} shows the system we have constructed to implement the proposed method. In the system, we increase the resolution of the CMOS camera to make sure that the captured Fourier intensity image has a sampling rate at least two times in each dimension higher than that of the object image $ \mathbf{x} $. As a result, the number of samples in the zero diffraction order of the Fourier intensity image is $ 762\times762 $, while the number of samples of $ \mathbf{x} $ is $ 128\times128 $. Compared with other deep learning-based phase retrieval approaches that often accept very small intensity measurements (such as $64\times64$ or even $ 28\times 28 $ pixels), we choose to use normal-sized intensity measurements ($ 762\times 762 $ pixels) to allow a sufficiently large oversampling ratio and object images of relatively large size ($ 128\times 128 $ pixels). It thus enables more applications for the proposed \textit{SiSPRNet}. From the $ 762\times 762 $ pixels intensity measurement, we extract the central $ 128\times 128 $ pixels for feeding to the proposed \textit{SiSPRNet} model. Such a choice is optimal in balancing between complexity and performance, as shown in the ablation analysis in Section \ref{Sec:ablainputsize}.
    
    When capturing Fourier intensity images, it is common to have the saturation problem. It is because most structured signals have the energy concentrated in the low-frequency areas, in particular the zero frequency. Due to the limited analog-to-digital resolution of standard scientific cameras ($14$ - $16$ bits), the central regions of the Fourier intensity measurement are often seriously saturated. To lessen the effect of the saturation problem, we can defocus the measurement by moving the camera beyond the Fourier plane to reduce the dynamic range \cite{goodman2017introduction}. The operation can be mathematically modeled by the convolution of $\mathcal{F}(\mathbf{x})$ with a defocus function $\mathbf{H}$. In our simulation and experiment, we synthesized it by multiplying the inverse Fourier transform of $\mathbf{H}$ with the object. More details can be found in Section \ref{Sec:simulation} and \ref{Sec:experiment}. It can be implemented by moving the camera beyond the Fourier plane. Fig. \ref{fig:opticalpath} shows a typical optical path for implementing the defocus phase retrieval system. Fig. \ref{fig:compvisual} shows some examples of the Fourier intensity measurements obtained in our experiments. 
    
    Besides the lens-based system that we have implemented in this work, we can also find lensless phase retrieval systems \cite{Sinha17}. Different from the lens-based system that reconstructs images from their Fraunhofer diffraction patterns, the lensless system reconstructs images based on their Fresnel diffraction patterns \cite{Sinha17}. Compared with the lens-based system, the lensless system requires a much longer distance between the sensors and objects. For instance, in \cite{Sinha17}, the distance between the sensor and object is reported as $37.5 cm$  or longer. By using a lens, the lens-based system can generate high-quality Fraunhofer diffraction patterns at a short propagation distance between the lens and sensor ($3 cm$  or less).

    \subsection{Deep Learning Model: \textit{ SiSPRNet}}
    
    With the oversampling constraint as mentioned above, it is possible to reconstruct a complex-valued signal (with trivial ambiguities) from only its Fourier intensity without other constraints (such as optical masks). However, such a reconstruction problem is highly non-convex that is difficult to solve with traditional optimization methods. Due to their non-linear characteristic, deep neural networks have been successfully applied to many non-convex optimization problems. In this paper, we propose to regard the single-shot maskless phase retrieval challenge as an end-to-end learning task, and formulate the retrieval process directly by a deep neural network model (denoted as $P(\mathbf{Y};\theta)$) as:
    
    \begin{dmath}
    	\hat{\mathbf{x}} = P(\mathbf{Y};\theta),
    \end{dmath}
    where $\theta$ represents the learnable parameters of the deep learning model. The training target is to minimize the error between the predicted $P(\mathbf{Y};\theta)$ and the ground-truth complex-valued image $ \mathbf{x} $ regularized with the total variation (TV) norm $ TV(\hat{\mathbf{x}}) $. The training process can be written as:
    
    \begin{equation}
    	P(\mathbf{Y};\theta) = \mathop{argmin}\limits_{\hat{\mathbf{x}}} \left( \left\| \mathbf{x}-\hat{\mathbf{x}} \right\|_1 + \alpha TV\left(\hat{\mathbf{x}}\right)\right), 
    \end{equation}
    where $\left\| \cdot \right\|_1$ denotes the $\ell_1$-norm distance and $ \alpha $ is a constant determined empirically.
    
\begin{figure} [htb]
    	\centering
    	\includegraphics[width=\linewidth]{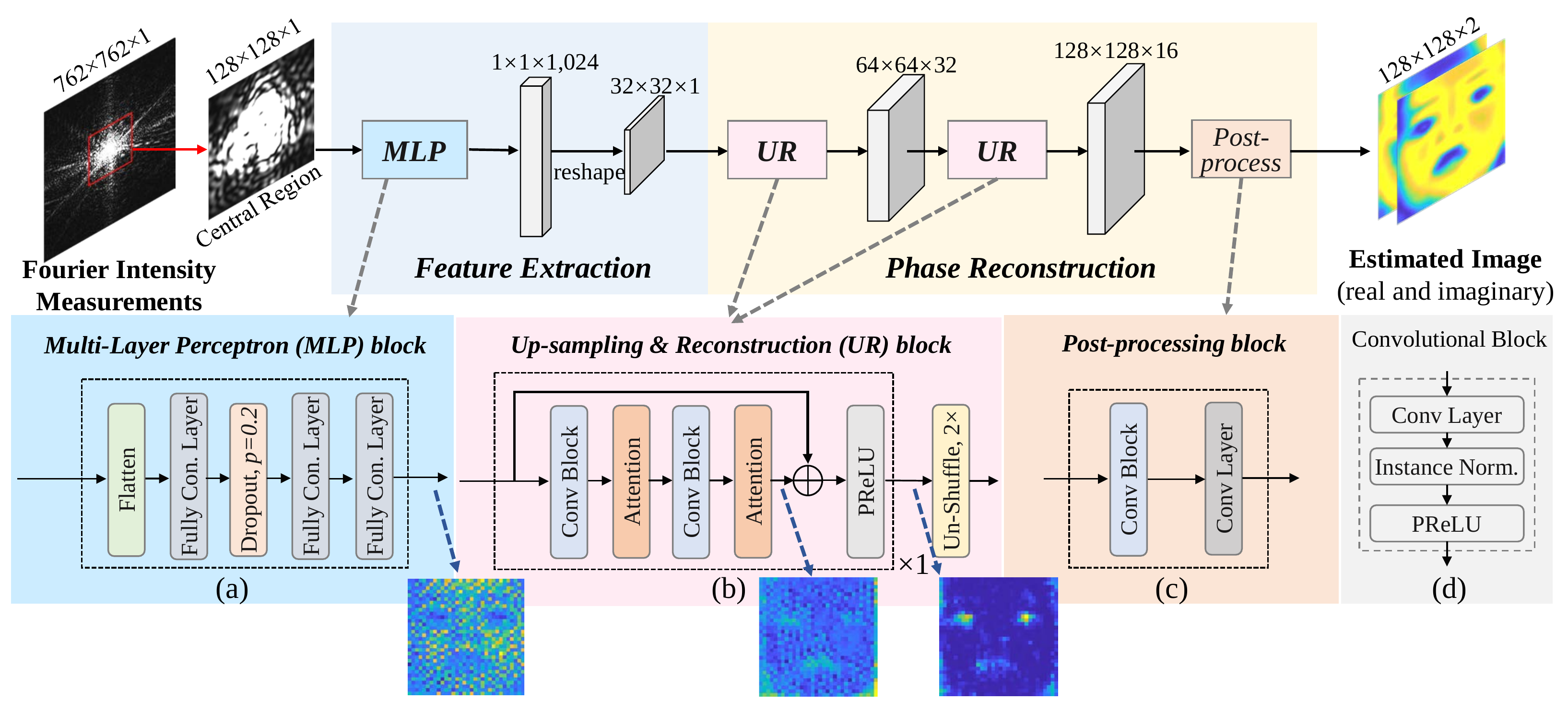}
    	\caption{Structure of the proposed network that mainly consists of two parts: Feature Extraction and Phase Reconstruction.}
    	\label{fig:structure}
    \end{figure}

    Fig. \ref{fig:structure} shows the overall structure of the proposed phase retrieval network: \textit{SiSPRNet}. It mainly consists of two parts: feature extraction and phase reconstruction. It takes a Fourier intensity measurement $ \mathbf{Y} $ as input and reconstructs a complex-valued image $ \mathbf{x} $ as its output.
    
    \subsubsection{\textbf{Feature Extraction}}
    Traditional learning-based approaches often use a CNN as the front end to extract the features of the Fourier intensity measurement. However, the convolution operation of CNN focuses on the local correlation of the input signal. It does not match with the property of Fourier intensity images that are globally correlated. Therefore, we propose to use a new Multi-Layer Perceptron (MLP) block to fuse the information of different frequency data as shown in Fig. \ref{fig:structure}(a). The MLP block consists of three Fully Connected (FC) and one Dropout layer. The input Fourier measurement is firstly reshaped from the size of $128\times 128\times 1$ to $1\times1\times16,384$, and then fed to the first FC layer of the MLP block. Each FC layer has $1,024$ neurons for extracting the features required for reconstructing the original complex-valued images. The number of neurons is smaller than the total pixel number of the input image. It can guide the backpropagation process to train the MLP to generate useful features of the reconstructed image while ignoring the less important information, such as noises and outliers. Such a design also reduces the complexity of the network. More analysis on this can be found in the ablation analysis in Section \ref{Sec:ablaMLPsize}. We also propose to embed a Dropout layer after the first FC layer. The Dropout layer randomly deactivates a small number (empirically $20\%$) of neurons at the training stage. At the inference stage, all neurons are activated that forms an ensemble to reduce the risk of overfitting \cite{srivastava2014dropout}. Our design with embedding a Dropout layer after the first FC layer constrains the behavior of the FC layers so that the features extracted by the MLP block do not depend on a particular input frequency measurement. The effectiveness of the Dropout layer has been evaluated and shown in Section \ref{Sec:abladropout}. Fig. \ref{fig:structure} shows an example of the original complex-valued image (real and imaginary), the Fourier intensity measurement, and the features extracted by the MLP block. It can be seen that the MLP block has performed the domain conversion. The extracted feature image contains the features of the original complex-valued image. It will be sent to the phase reconstruction part to reconstruct the real and imaginary images.

    \subsubsection{\textbf{Phase Reconstruction}}
    The phase reconstruction part contains two Up-sampling and Reconstruction (UR) blocks and one Post-processing block, as shown in Fig. \ref{fig:structure}(b) and (c). The UR block is responsible for reconstructing the complex-valued image from the extracted features and progressively upsampling it back to the original size. The basic processing unit of the UR block is the convolution block that has three components: convolutional, instance normalization, and Parametric Rectified Linear Unit (PReLU) layers. The convolutional layer contains $3\times 3$ learnable filters that aim to implicitly approximate the nonlinear backward mapping from the extracted features to the desired complex-valued image $ \mathbf{x} $. The instance normalization and PReLU layers are non-linear functions that increase the non-linear capacity and learning ability of the deep learning model. 
    
\begin{figure} [htb]
    	\centering
    	\includegraphics[width=0.5\linewidth]{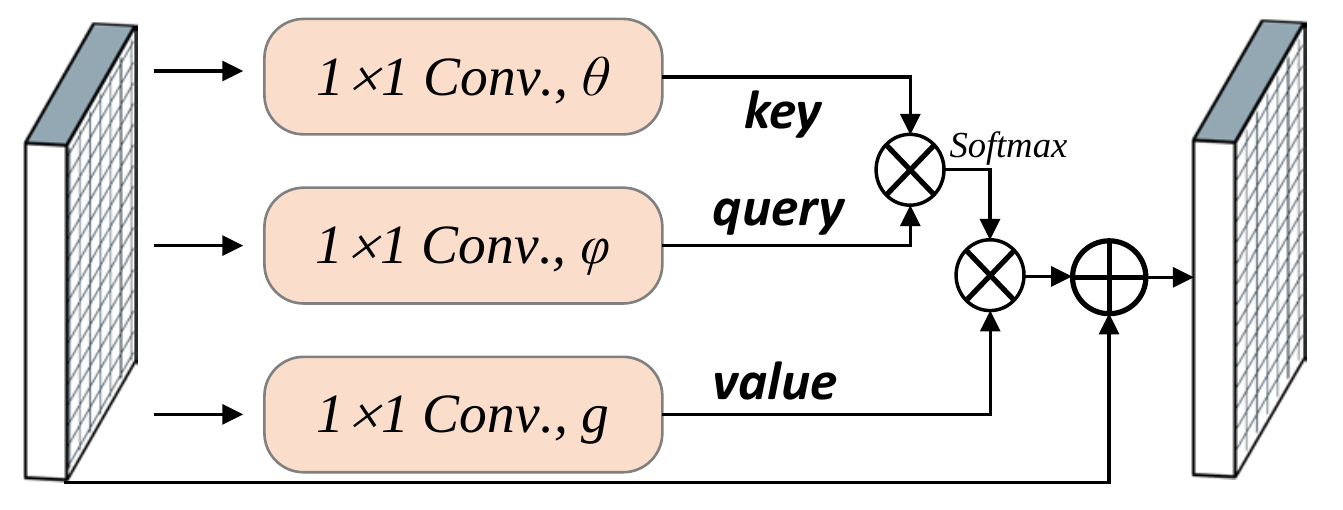}
    	\caption{Structure of the attention mechanism.}
    	\label{fig:attn}
    \end{figure}

    A self-attention unit is followed after each convolutional block. Recently, the self-attention technique has been popularly used in image reconstruction tasks, e.g., the scattering problem \cite{non-local, Wang:21}. Traditional CNN using small convolution kernels can only extract the local correlation of an image. However, natural images are often structural that exhibit global correlation. Self-attention is an effective way to exploit the global correlation in images. The structure of the self-attention layer used in the UR block is similar to that in \cite{non-local}, as shown in Fig. \ref{fig:attn}. At the first stage, the input features are fed to three $1\times1$ convolutional layers ($\theta, \phi$ and $g$). The outputs are denoted as key, query, and value, respectively, all with the same dimensions. Next, the cross-correlation matrix is computed between the key and query. It represents the dependency between any two pixels of the input features. The softmax function is then performed to normalize the correlation scores and build an attention map. Large values in the attention map denote strong correlations between the features concerned. Since the correlation is computed among all features, the attention is made not only to those with local correlations but also the global ones. The result is finally multiplied with the value to generate the self-attention feature maps. They are then added back to the original features to strengthen the features of interest. Such an attention mechanism is important to the training of the UR block. It allows the UR block to reconstruct images with globally correlated features. For instance, for the extracted features as shown in Fig. \ref{fig:structure}, the UR block will try to reconstruct both the left and right eyes together since they have a high correlation as indicated in the training face images in the dataset. Since the correlation of the features may change at different resolutions, another UR block is used after the first one on the features of higher resolution.
    
    Direct stacking covolutional and attention layers would lead to the obvious performance drop because of weak information flow and gradient vanishing \cite{He_2016_CVPR}. Therefore, we add a shortcut connection after the convolution block and attention mechanism that forms a residual structure (shown in Fig. \ref{fig:structure}(b)), which is widely adopted in image processing tasks \cite{Wang_2017_CVPR, zhang2021explainable, park2021dynamic}. The residual setting avoids the gradient vanishing problem and therefore reduces the learning difficulty. As shown in Fig. \ref{fig:structure}, the self-attention layers effectively estimate the residue between the input feature and desired one. Finally, we utilize the unshuffle layer to reshape the feature map to enlarge the spatial sizes. The effectiveness of the residual self-attention structure is further discussed in Section \ref{Sec:ablaresidual}.

    \subsubsection{\textbf{Post-processing block}}
    The Post-processing block contains two convolutional layers (shown in Fig. \ref{fig:structure}(c)). It estimates the real $\hat{\mathbf{x}}_{Re}$ and imaginary $\hat{\mathbf{x}}_{Im}$ parts of the image from the up-sampled features.

    \subsubsection{\textbf{Loss Function}}
    We regard the phase retrieval problem as a supervised learning task. At the training stage, we prepare a set of paired training data. Each Fourier intensity measurement has a paired complex-valued image as the ground-truth target. Then, the deep learning model can be optimized by minimizing the $\ell_1$-norm distance between the estimation and the ground-truth images plus the TV-norm to remove the noise while keeping the image edges. The loss function $\mathcal{L}$ is defined as: 
    \begin{equation}
    	\mathcal{L} = \left\| \mathbf{x}_{Re}-\hat{\mathbf{x}}_{Re} \right\|_1 + \left\| \mathbf{x}_{Im}-\hat{\mathbf{x}}_{Im} \right\|_1 + \alpha TV\left(\hat{\mathbf{x}}\right).
    \end{equation}
    The symbols $\mathbf{x}_{Re}$ and $\mathbf{x}_{Im}$ represent the ground-truth real and imaginary parts of the image, respectively. 
    
        \section{Simulation Results} \label{Sec:simulation}
    
    \subsection{Simulation Setup}
    
    We conducted the simulation based on the images from two different datasets: the Real-world Affective Faces (RAF) dataset \cite{li2017reliable} and Fashion-MNIST dateset \cite{xiao2017/online}. Specifically, the training data was converted to gray-scale image and resized to $128 \times 128$. Then the pre-processed images $\mathbf{x} \in [0, 1]$ were used to generate complex-valued images of two forms: phase-only and magnitude-phase. For generating the phase-only images, we mapped $\mathbf{x}$ to the $2\pi$ phase domain via an exponential function: $\exp(2\pi i \mathbf{x})$, with the magnitude kept as $1$. On the other hand, we used the images of Fashion-MNIST dataset to generate the magnitude-phase images. To be specific, we directly used the pre-processed images $\mathbf{x} \in [0, 1]$ as the magnitudes of the images, i.e., $\mathbf{x}_{mag} = \mathbf{x} \in [0, 1]$. Then we set the phase parts through $\mathbf{x}_{phase} = \exp(2\pi i \mathbf{x}_{mag})$ and combined them by $\mathbf{x} = \mathbf{x}_{mag} \circ \mathbf{x}_{phase}$ as the magnitude-phase training and testing images, where $\circ$ denotes the elementwise multiplication. The images, however, have linearly related magnitude and phase.
    
    As mentioned in Section \ref{Sec:Method}, we suggest to reduce the saturation problem by convolving the Fourier intensity measurement with a defocus function. Based on the convolution property of the Fourier transform, we can simulate it by multiplying the testing image with a defocus kernel $\mathbf{h}$ in the spatial domain. It can be generated through the Fresnel propagation \cite{goodman2017introduction}. Denote the original image as $x(p, q)$ and $X(u, v)$ as its  Fourier transform, then the optical field $X_L(u', v')$ in the defocus plane is expressed as:
    \begin{dmath}
    	\label{Eq:defocuskernel}
    	X_L(u', v') 
    	=  C \iint X(u, v) H(u - \frac{u'}{\lambda L}, v - \frac{v'}{\lambda L}) du\ dv,
    \end{dmath}
    where $\lambda$, $L$ and $H$ denotes the wavelength, distance between the lens after the object and the defocus plane, and the Fourier transform of the defocus kernel $\mathbf{h}$, respectively. $C$ is the constant and $h(p, q)$ is given as $e^{\frac{j\pi}{\lambda L}c(p^{2}+q^{2})}$. Based on the convolution theorem, Eq. \eqref{Eq:defocuskernel} is equivalent to the elementwise multiplication between $h$ and $x$ in the spatial domain with the scaling factor $\lambda L$. 
    
    Then, oversampled discrete Fourier transform (DFT) with size $ 762\times762 $ (the same size as in real experiments) was performed to generate the intensity measurements. For each image, the Fourier intensity measurement was capped at the value $ 4095 $ ($12$-bit) to simulate the dynamic range of the practical scientific cameras. The central $128\times 128$ patch was cropped as the input of the \textit{SiSPRNet} (the performance of different input sizes is discussed in Section \ref{Sec:ablainputsize}). The total epoch and mini-batch size were set as $1,000$ and $32$. The weight $\alpha$ of the TV loss function was set as $1$. The Adam optimizer \cite{kingma2017adam} with the learning rate of $10^{-4}$ was implemented to update the gradients in each training step. Note that the learning rate was updated through \textit{StepLR} scheduler in PyTorch every $100$ epoches with $\gamma = 0.9$.
    
    We adopted Peak Signal to Noise Ratio (PSNR, higher the better), Structural Similarity (SSIM, higher the better) and Mean Absolute Error (MAE, lower the better) as the performance metrics to measure the difference between the reconstructed images $\hat{\mathbf{x}}$ and ground-truth images $\mathbf{x}$. For each dataset, we selected the first $1,000$ images in the test set as the testing samples for evaluation. We converted the reconstructed real and imaginary parts to the magnitude and phase representation. The phase values of both images were shifted by $\pi$ to ensure non-negative values for the computation of PSNR. The average PSNR, SSIM and MAE were acquired among all $1,000$ testing images for evaluation. 
    
    \subsection{Ablation Study} \label{Sec:ablationstudy}
    
    To gain a deeper insight into the improvements obtained by \textit{SiSPRNet}, we conducted several ablation studies on different components of our framework and analyzed their impact on the speed and accuracy of the network.

    \subsubsection{\textbf{Impact of Input Size}} \label{Sec:ablainputsize}
    
\begin{figure}[htb]		
    	\centering
    	\includegraphics[width=0.7\linewidth]{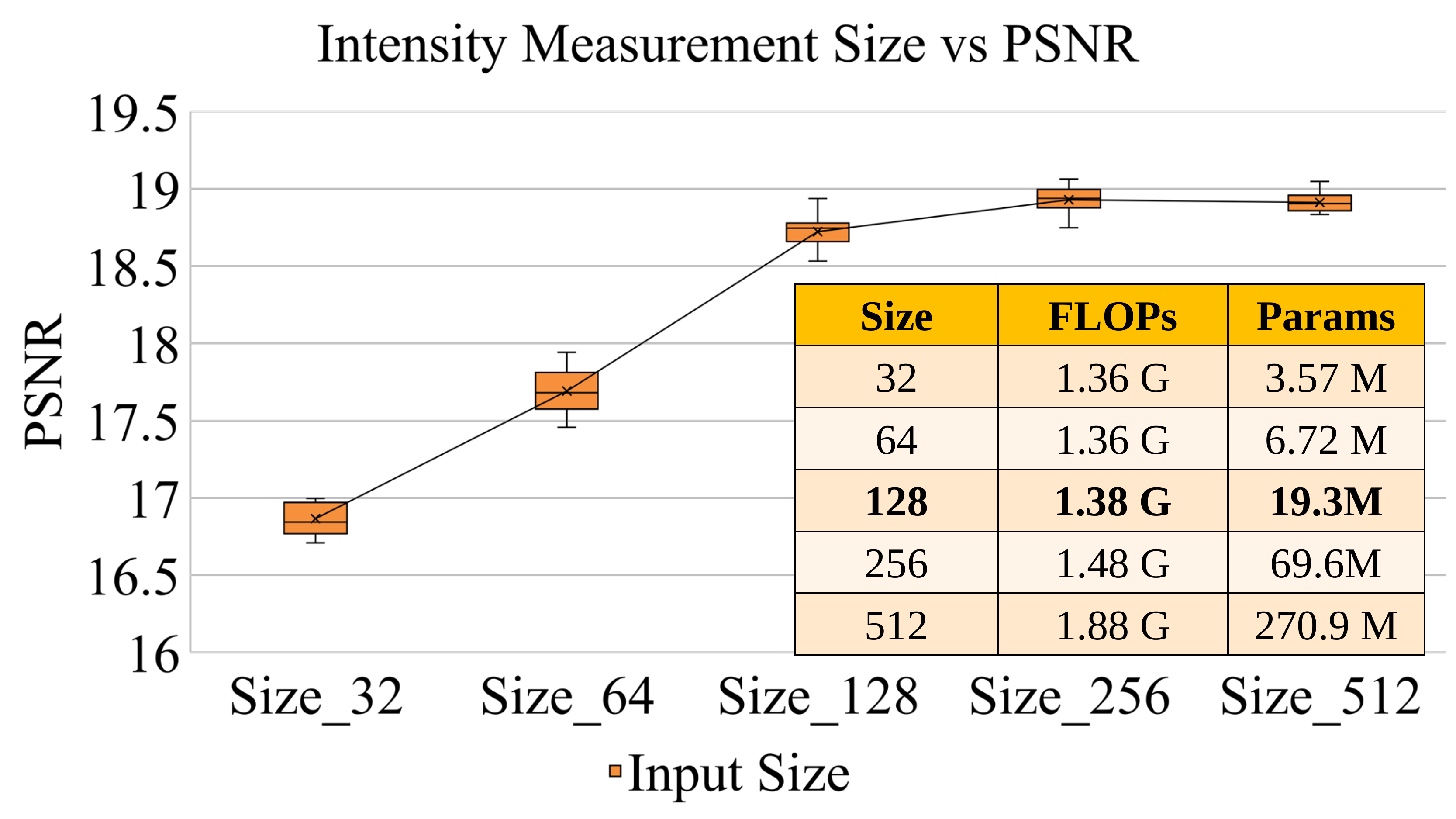}
    	\caption{Ablation study on the input sizes. The results are obtained with the RAF dataset \cite{li2017reliable}.}
    	\label{fig:Inputsize}
    \end{figure}

    For the two-dimensional Fourier spectrum, the central regions represent the low-frequency information while outer areas contain the high-frequency components. Structured signals (e.g., natural images) usually have their energy concentrated in the low-frequency regions; hence the high-frequency components have relatively lower intensity and are dominated by noise. An ablation analysis was performed to determine the size of the central regions to be used. In our experiments, the intensity measurements have the size of $762 \times 762$. We fed the central $[32 \times 32, 64 \times 64, 128 \times 128, 256 \times 256, 512 \times 512]$ of the intensity measurement to the same network and study the performance, as shown in Fig. \ref{fig:Inputsize}. The average testing PSNR from $30000th$ to $35000th$ training steps was chosen for comparison. The box plot of the performance, the floating point operations per second (FLOPs) and the numbers of trainable parameters with respect to the sizes are shown in Fig. \ref{fig:Inputsize}. Although the model performance keeps increasing along with the increase of input size, the number of the trainable parameters also increases quickly, which brings a huge computational burden. For example, it can be seen that the number of parameters dramatically increases from $ 19.3M $ to $ 69.6M $ when the input size changes from $128 \times 128$ to $256 \times 256$. While the growth speed of the performance is linear when the input size is smaller than $256 \times 256$, it saturates as the input size increases to about $128 \times 128$. The reason why the performance of \textit{SiSPRNet} cannot further improve with a larger input size is two-folded. When selecting the central $128 \times 128$ pixels, we have selected most of the significant data in the intensity measurement. When increasing the size to $256 \times 256$, some small magnitude data can be included, but they are much smaller than the data in the middle due to the large dynamic range of the intensity measurement. As deep neural networks work on statistics, these data with very small magnitudes can only slightly influence the overall performance. It is noted in Fig. \ref{fig:Inputsize} that only a slight improvement is achieved. If we further increase the size to $512 \times 512$, no improvement in performance is noted since most of the data included have very small magnitudes and are likely quantized to zeros when converting the data to $12$-bit integers. Even if they are not quantized to zero, they will have a very low SNR due to the quantization error. Based on the above, we believe it is a good choice to adopt $128 \times 128$ as the input size to balance performance and efficiency.
    
    \subsubsection{\textbf{Impact of Defocus Distance}} \label{Sec:abladefocusdist}
    
\begin{figure}[htb]		
    	\centering	\includegraphics[width=0.7\linewidth]{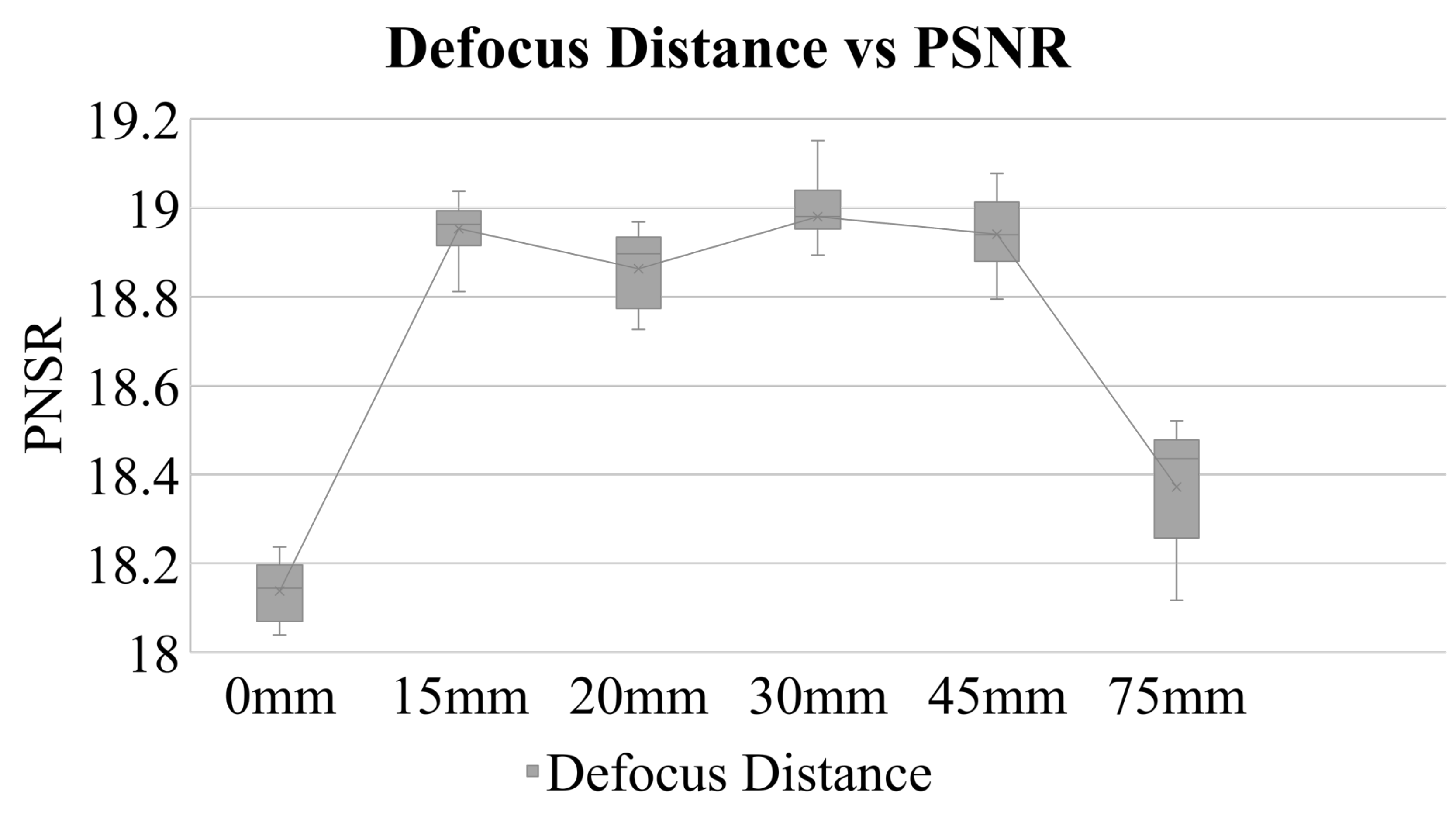}
    	\caption{Ablation study on the defocus distance. The results are obtained with the RAF dataset \cite{li2017reliable}. }
    	\label{fig:Defocusdist}
    \end{figure}

    The effect of the defocus distance $L$ is discussed in this section. We investigated $6$ situations of the defocus distance: $0mm$ (no defocusing), $15mm$, $20mm$, $30mm$, $45mm$, and $75mm$. The defocus kernel is a pure-phase object generated by Holoeye built-in function.  We used the same illumination strength and exposure time for all defocus distances. The size of the input intensity measurements is still $128 \times 128$. The box plot of the average PSNR value at each defocus distance is presented in Fig. \ref{fig:Defocusdist}. As shown in the figure, the reconstruction performance significantly degrades when there is no defocusing ($L = 0 mm$). It is because some pixels are severely saturated and give wrong information to the network. On the other hand, the performances are similar when $L = 15$ to $45 mm$ although the PSNR is slightly higher when $L = 30 mm$. It shows that the performance of the system is not sensitive to the choice of $L$ as long as it is not too small or too big.  When the defocus distance is large, i.e. $75 mm$, the PSNR significantly drops. It is because a portion of the Fourier intensity measurements (high-frequency areas) become very small due to the decrease in light intensity with the long defocus distance.  To balance the illumination efficiency and the reconstruction performances, the defocus distance $L$ was selected as $30 mm$ for all experiments.
    
    \subsubsection{\textbf{Impact of Dropout Layer}} \label{Sec:abladropout}
    
\begin{figure}[htb]		
    	\centering	\includegraphics[width=0.7\linewidth]{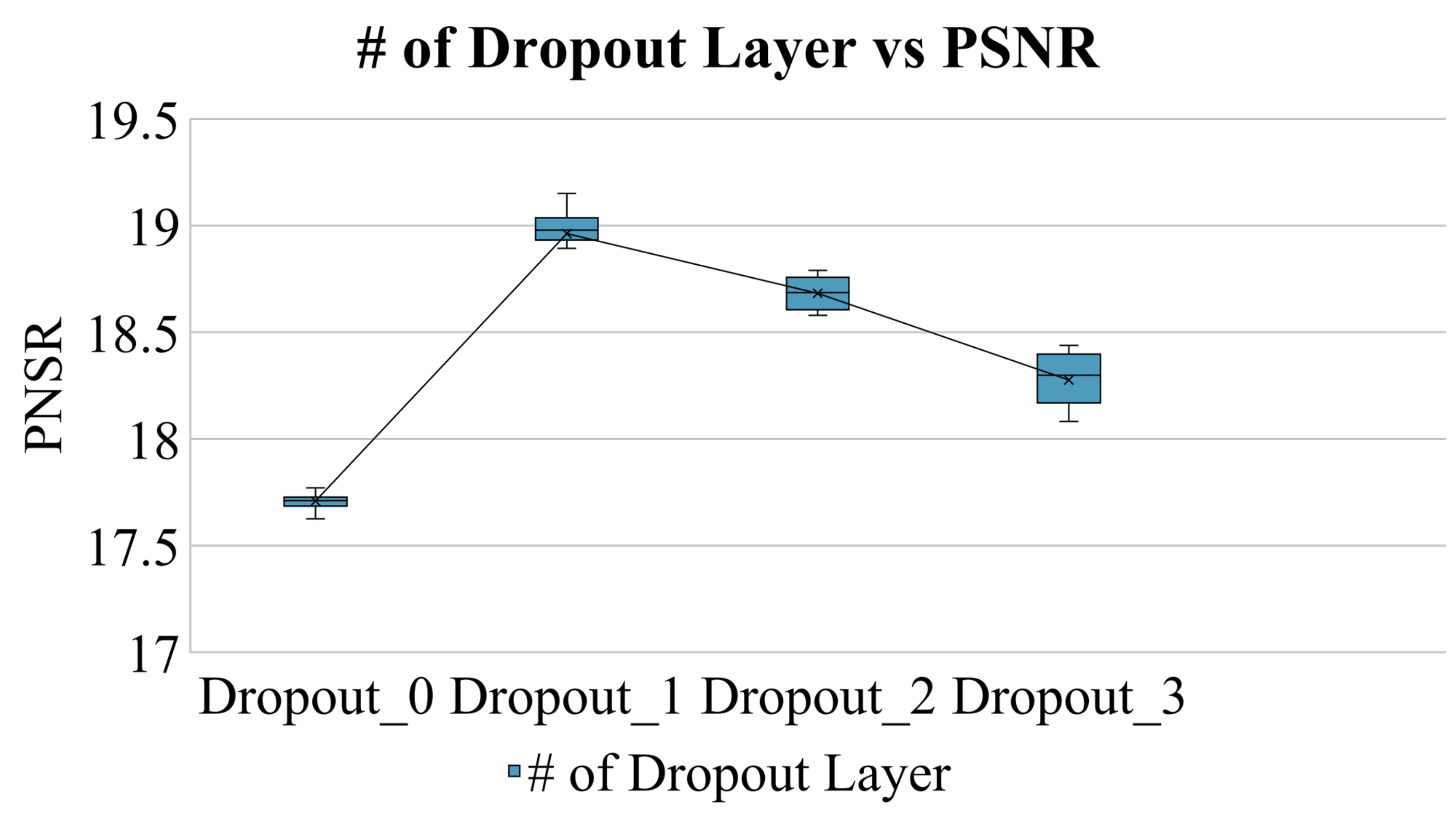}
    	\caption{Ablation study on the number of Dropout layers in feature extraction. The results are obtained with the RAF dataset \cite{li2017reliable}. }
    	\label{fig:Dropout}
    \end{figure}

    We investigated the effect of the number of Dropout layers in feature extraction part. We set the dropout rate as $0.2$ in every Dropout layer. Since $3$ FC layers are used in the MLP blocks, we investigated $4$ situations when introducing the Dropout layers: $0$ Dropout layer, $1$ Dropout layer after the first FC layer, $2$ Dropout layers after each of the first two FC layers, and $3$ Dropout layers after each of the FC layers. One Dropout layer is the default setting in our proposed \textit{SiSPRNet}. The size of the input intensity measurements is $128 \times 128$. The box plot of the average PSNR value in each situation is presented in Fig. \ref{fig:Dropout}. As shown in the figure, if there is no Dropout layer, the average PSNR of the reconstructed images is the lowest because of the over-fitting problem. The network is lazy to investigate the hidden frequency representations, and the reconstruction relies on a small number of inputs. The Dropout layer randomly deactivates a part of the inputs at the training stage, which inspires the network to find more global and robust representations. At the testing stage, all input signals are used to form an ensemble learning structure that increases the generalization power of the whole network. After implementing the Dropout layer, the model performance boosts as expected. Interestingly, with more dropout layers applied, there is too much information ignored in the training stage that hurts the learning progress of the network and decreases the performance. Consequently, only $1$ Dropout layer was adopted in the final model.
    
    \subsubsection{\textbf{Impact of Fully-Connected Layer}} \label{Sec:ablaMLPsize}
    
\begin{figure}[htb]		
    	\centering	\includegraphics[width=0.7\linewidth]{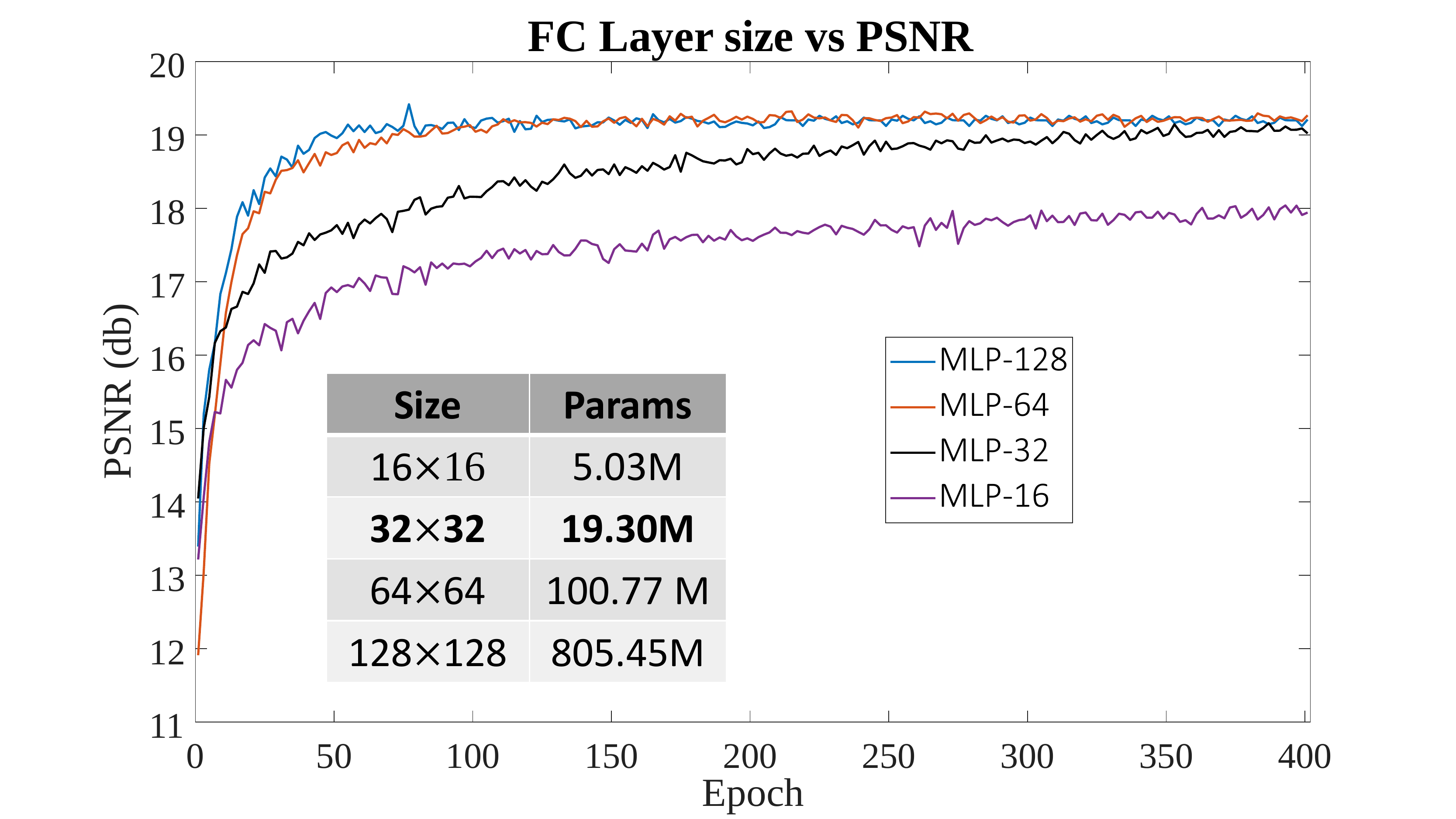}
    	\caption{Ablation study on the number of neurons of the FC layers. The results are obtained with the RAF dataset \cite{li2017reliable}. }
    	\label{fig:MLPSize}
    \end{figure}

    We investigated the impact of the size (numbers of neurons) of the FC layers of the MLP block on the overall performance. We investigated four situations: \textit{(i)} each FC layer has $256$ ($16\times16$) neurons; \textit{(ii)} each FC layer has $1024$ ($32\times32$) neurons; \textit{(iii)} each FC layer has $4096$ ($64\times64$) neurons, and \textit{(iv)} each FC layer has $16384$ ($128\times128$) neurons. The output in each case is reshaped to the corresponding 2-dimensional form before sending to the phase reconstruction part. In the phase reconstruction part, $256$-neuron FC has $3$ UR blocks, and other situations have $2$ UR blocks. $4096$-neuron FC only has only one upsampling layer in one of the UR blocks, and $16384$-neuron FC does not have upsampling layer. The average PSNR of the first $400$ training epochs is presented in Fig. \ref{fig:MLPSize}. As shown in the figure, the convergence speeds of the $4096$-neuron and $16384$-neuron FC are faster than those of the $1024$-neuron FC and $256$-neuron FC. However, $16384$-neuron, $4096$-neuron, and $1024$-neuron FCs achieve similar performance with the growth of training epochs. It suggests that $1024$-neuron FC should be sufficient to extract the representative features of the required image. On the other hand, the model size quickly grows with the increase in the number of neurons. Therefore, we adopted $1024$-neuron FC to balance performance and efficiency.
    
    \subsubsection{\textbf{Impact of Self-Attention Unit}} \label{Sec:ablaattention}
    
\begin{figure}[htb]		
    	\centering	\includegraphics[width=0.6\linewidth]{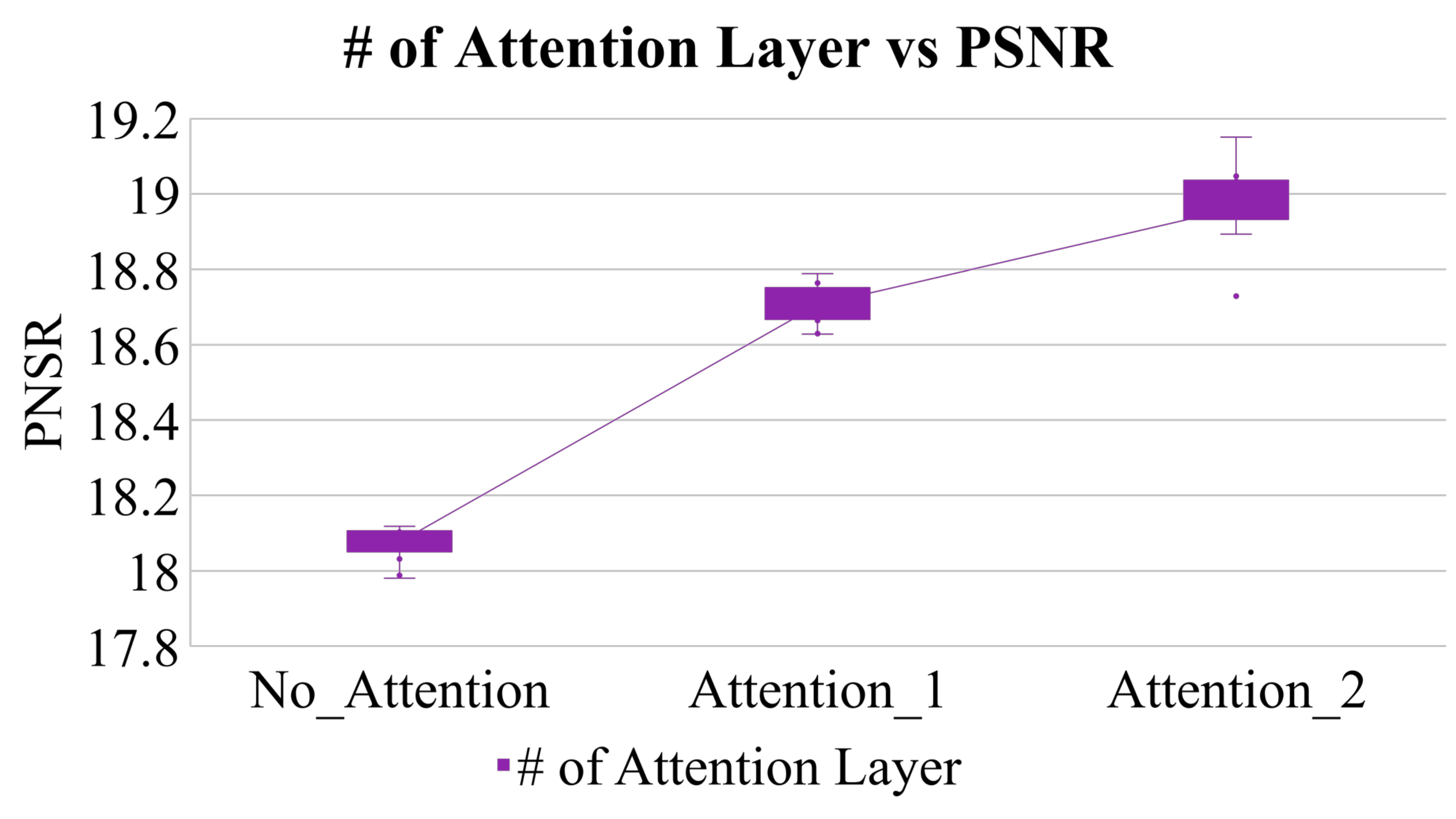}
    	\caption{Ablation study on the number of self-attention units in phase reconstruction part. The results are obtained with the RAF dataset \cite{li2017reliable}. }
    	\label{fig:Attention}
    \end{figure}

    We investigated the effect of the number of self-attention units in the phase reconstruction part. As shown in Fig. \ref{fig:structure}, there is a self-attention layer after each of the two convolution blocks in the UR block. We investigated $3$ situations when introducing the self-attention units: $0$ self-attention unit, $1$ self-attention unit after the first convolution block,  and $2$ self-attention units after each of the convolution blocks. The size of the input intensity measurements is still $128 \times 128$. The box plot of the average PSNR value in each situation is presented in Fig. \ref{fig:Attention}. As shown in the figure, if there is no self-attention unit, the average PSNR of the reconstructed images is the lowest. It is because the convolution blocks can only extract the local correlation of the image but ignore the global correlation that is commonly found in structured images. The self-attention unit computes the cross-correlation of any two positions in an image and generates the attention feature maps that strengthen the features of interest, which inspires the network to find more global and robust representations. After implementing the self-attention unit, the model performance increases as expected. Following the result of this ablation study, we adopted $2$ self-attention units in the UR block in our \textit{SiSPRNet}. 
    
    \subsubsection{\textbf{Impact of Residual Units}} \label{Sec:ablaresidual}
    
\begin{figure}[htb]		
    	\centering	\includegraphics[width=0.7\linewidth]{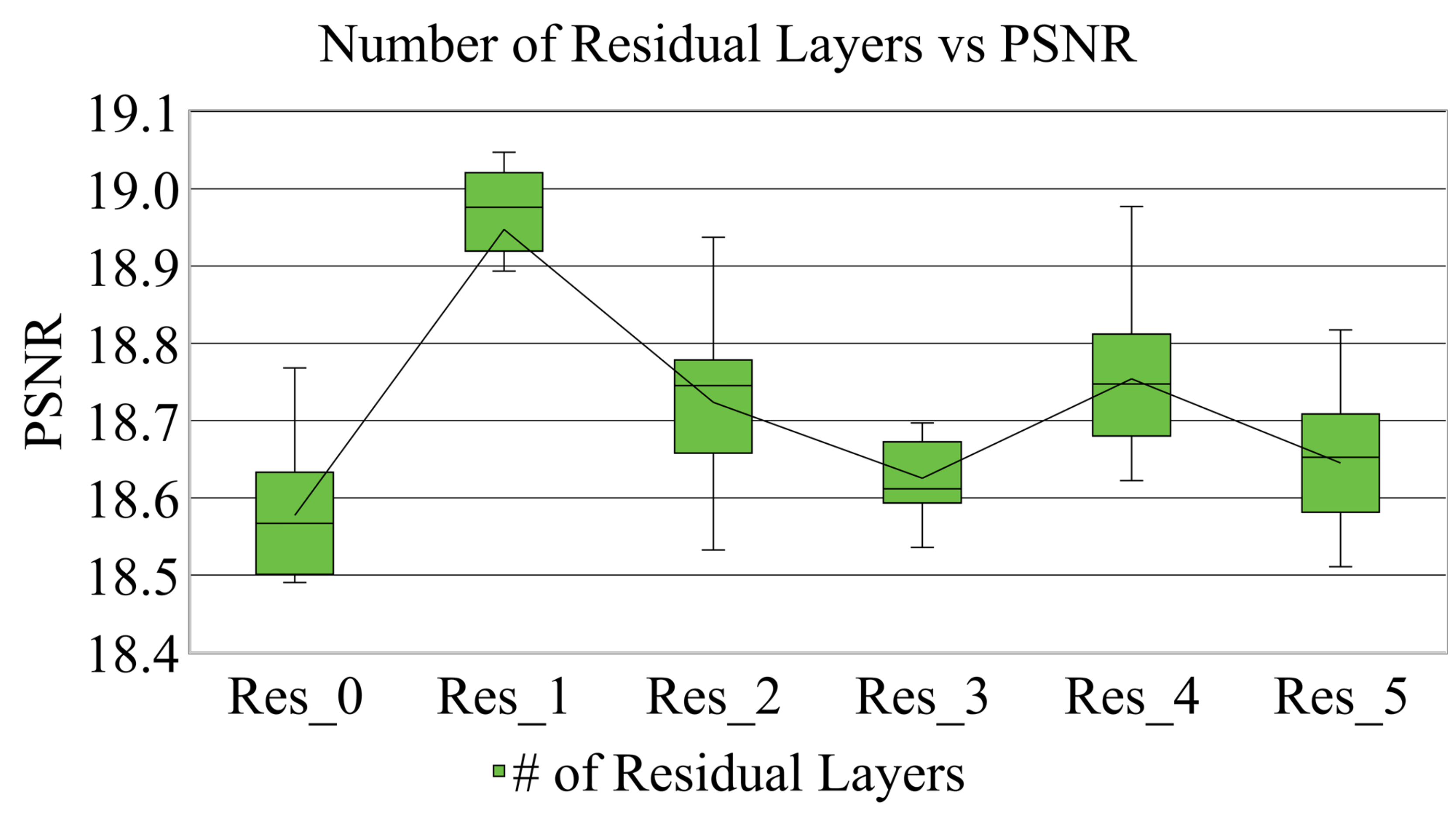}
    	\caption{Ablation study on the number of residual units. The results are obtained with the RAF dataset \cite{li2017reliable}.}
    	\label{fig:ResLayer}
    \end{figure}

    We also investigated the effects of the number of residual units in the phase reconstruction part. As shown in Fig. \ref{fig:structure}, each residual unit contains two layers of Conv Blocks interlaced with two self-attention layers plus the PReLU at the end. The size of the input images to the network is still $128 \times 128$. With everything remaining the same but just changing the number of residual units in each UR block, the average PSNRs of the estimated images are obtained and shown in Fig. \ref{fig:ResLayer}. It can be seen that using the residual units brings in better PSNR performance. The model with one residual unit has similar or even better performance than the models with more residual units. To pursue better PSNR and efficiency, we chose to use one residual unit in the UR block.
    
    \subsubsection{\textbf{Cross-Validation: Model Robustness}} \label{Sec:CrossValidation}
\begin{figure}[htb]
    	\centering
    	\begin{subfigure}[t]{\textwidth}
    		\begin{tabular}{*{2}{c@{\extracolsep{0em}}} }
    			\includegraphics[width=0.5\textwidth]{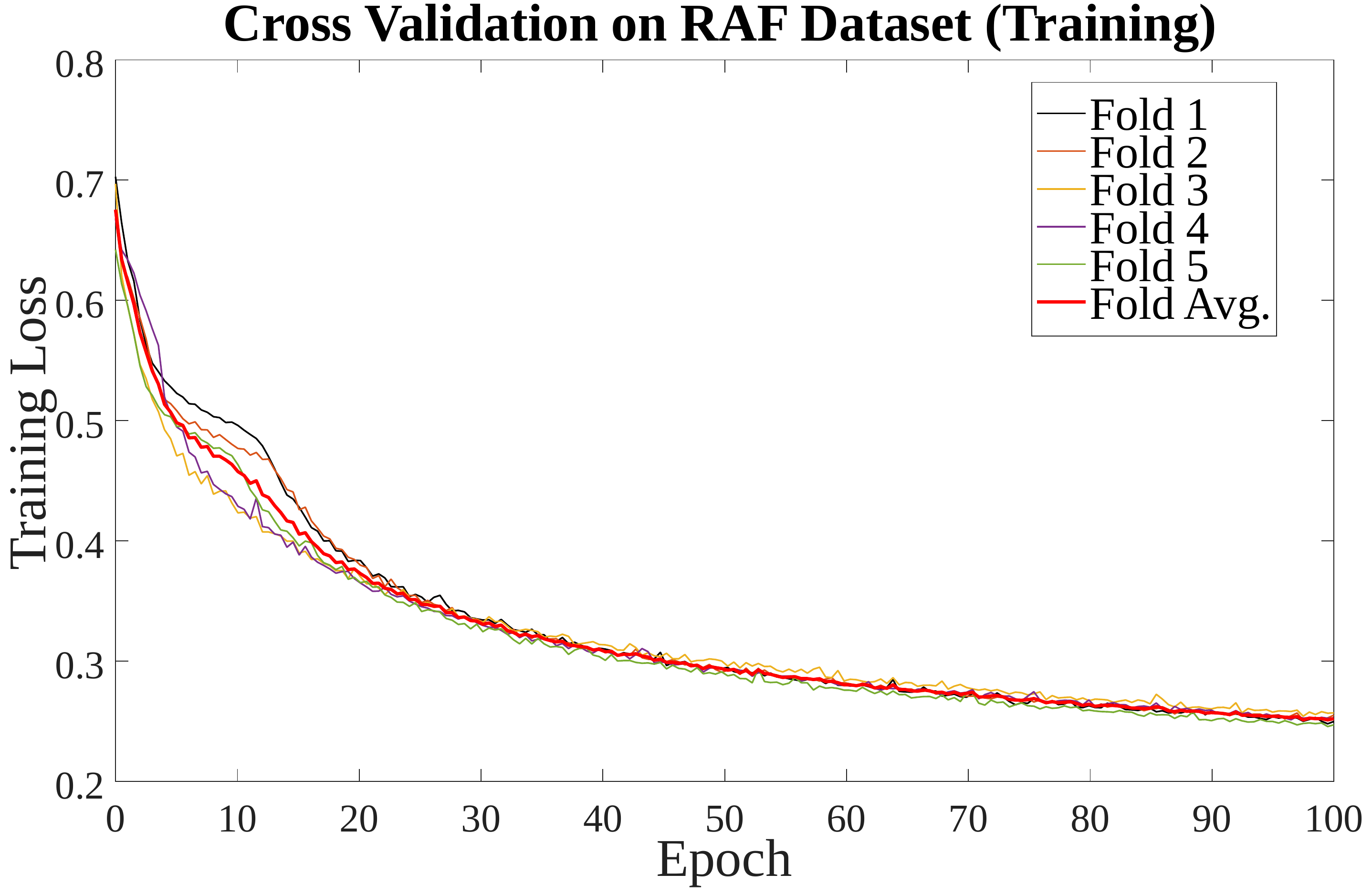}
    			\includegraphics[width=0.5\textwidth]{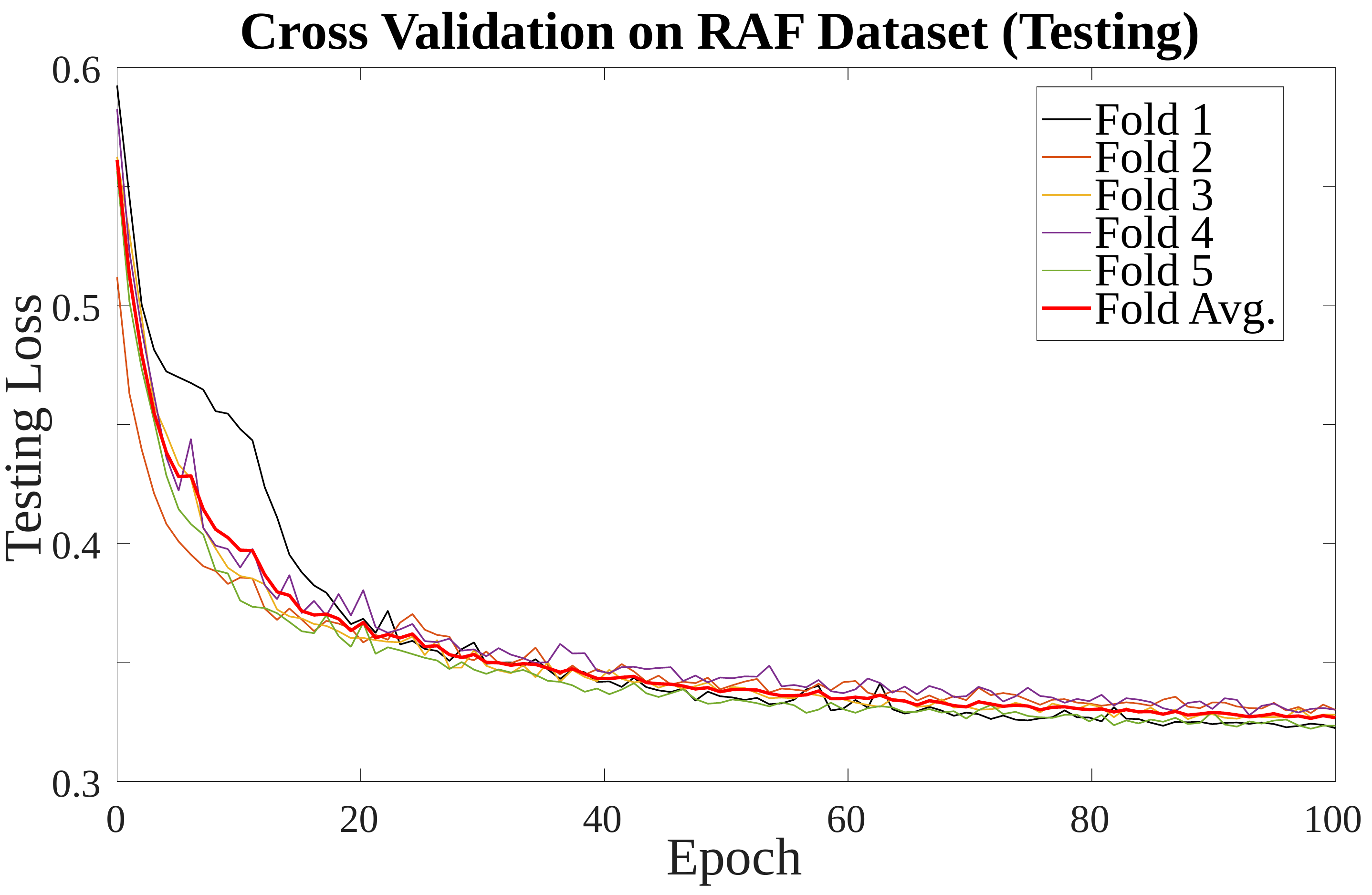}
    		\end{tabular}
    		\caption{}
    	\end{subfigure}
    	
    	\begin{subfigure}[t]{\textwidth}
    		\begin{tabular}{*{2}{c@{\extracolsep{0em}}} }
    			\includegraphics[width=0.5\textwidth]{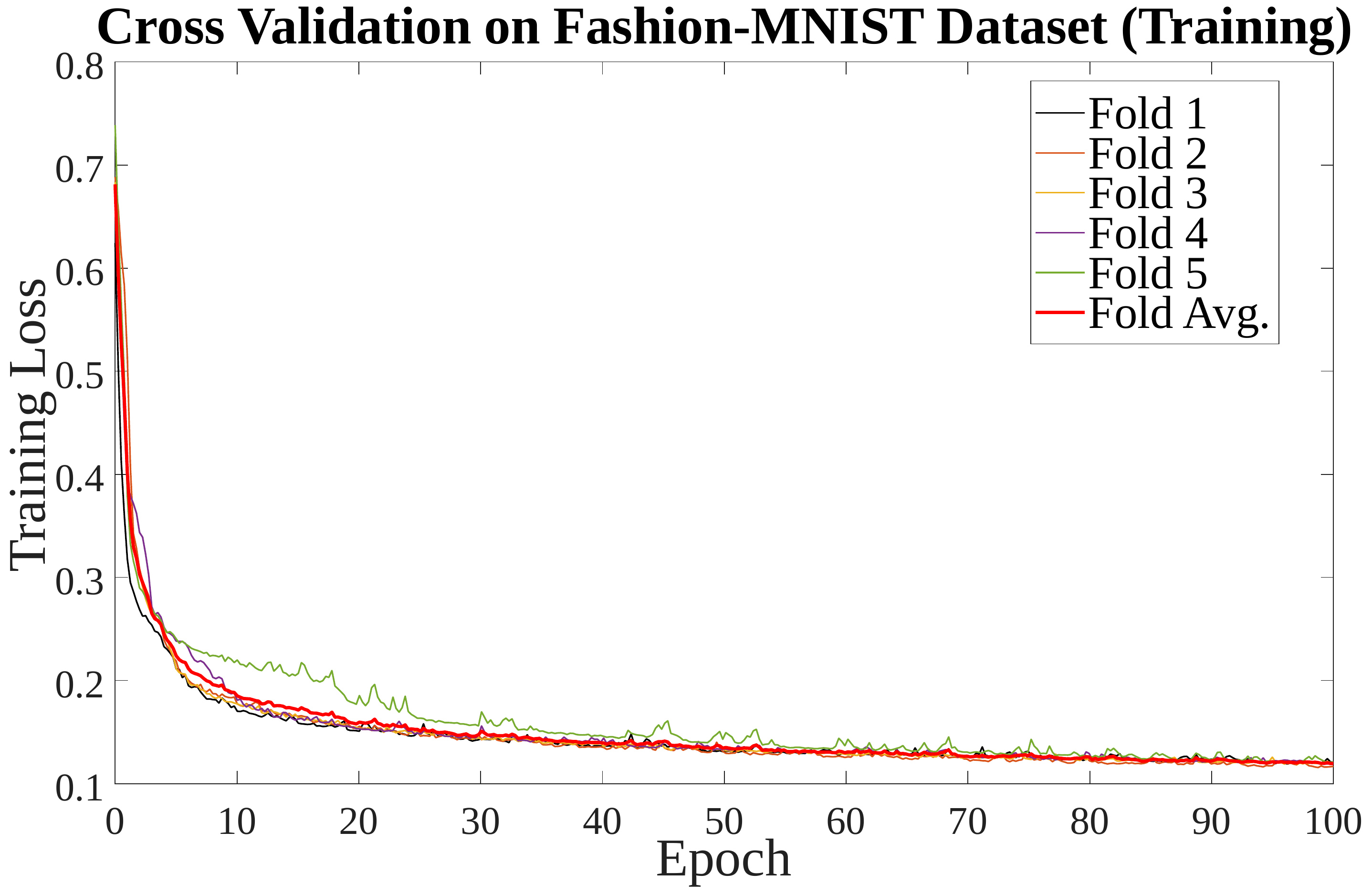}
    			\includegraphics[width=0.5\textwidth]{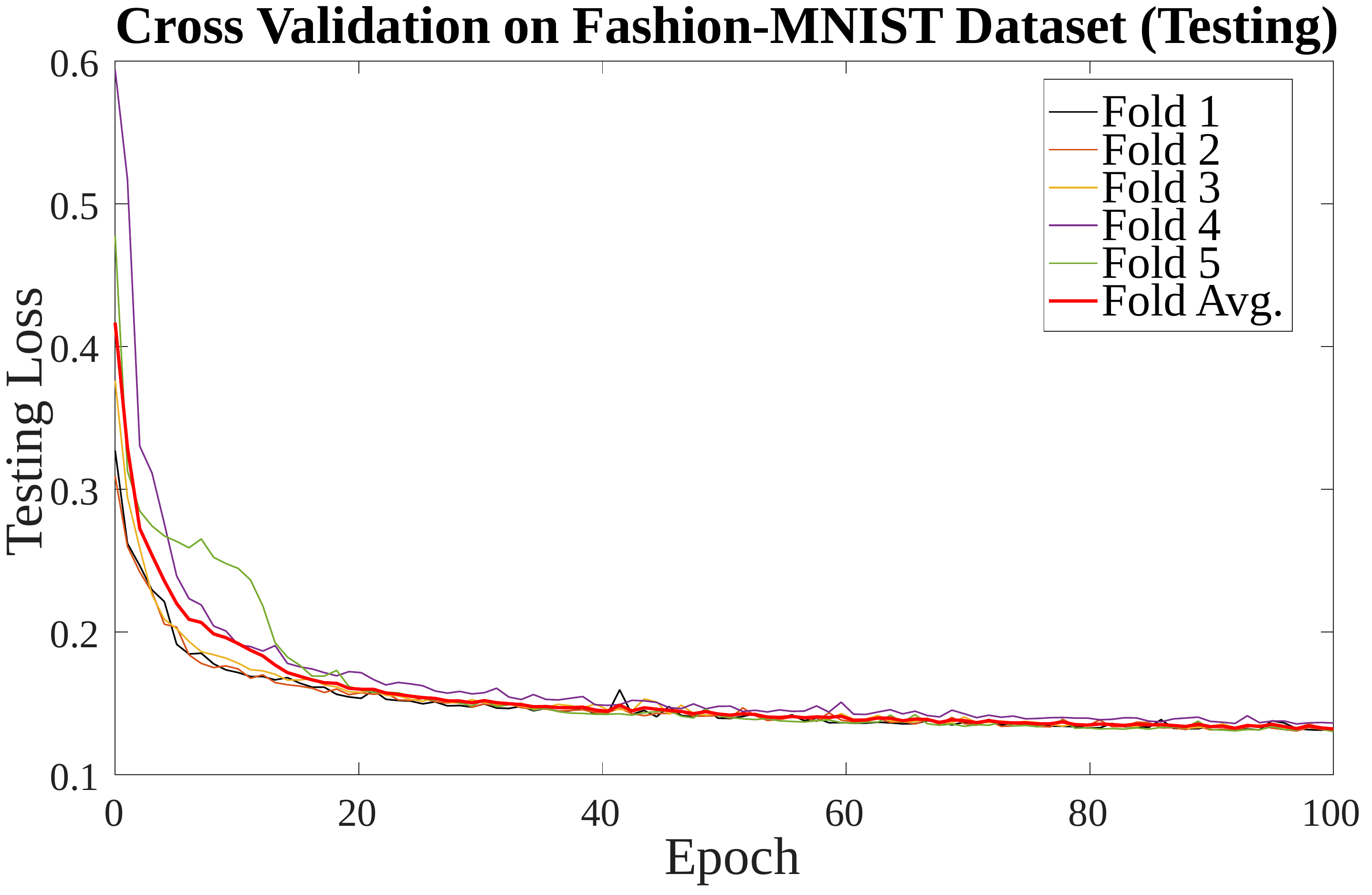}
    		\end{tabular}
    		\caption{}
    	\end{subfigure}
    	
    	\caption{Training \& Testing errors of $5$-Fold cross validation on (a) RAF dataset, and (b) Fashion-MNIST dataset, respectively. Fold Avg. (colored in red) denotes the mean of $5$ folds.}
    	\label{Fig:KFold}
    \end{figure}

    Besides the generalization ability, the robustness is also a crucial metric to evaluate neural networks. The more robust the network, the less probable the overfitting problem would happen in the network. In order to demonstrate the robustness of the proposed \textit{ SiSPRNet}, we utilized the cross-validation, a statistical method of evaluating learning-based approaches through dividing the whole datasets into two portions: one for training and the other for validating \cite{Refaeilzadeh2016}. In this ablation study, we adopted the K-Fold cross-validation which is the most popular cross-validation approach \cite{Stone_1974}. The procedure of the K-Fold cross validation can be briefly described as follows: \textit{1.} Randomly shuffle the dataset. \textit{2.} Split the dataset into $K$ groups.  \textit{3.} In each fold, take a group as the testing dataset and the remaining groups as the training dataset. Then, fit the deep learning model on the training dataset and do the evaluation on the testing dataset. \textit{4.} Reset the model parameters and repeat step \textit{3} until $K$ folds are accomplished. \textit{5.} Summarize the model evaluation scores obtained in each fold. To be specific, we adopted $K=5$ folds for evaluation and we ran $100$ epochs in each fold. In each fold, $80\%$ of the samples were treated as the training data and the rest $20\%$ were utilized for testing. We used the training (for the training dataset) and testing (for the testing dataset) losses as the evaluation scores. The training loss was recorded every $20$ training batches and the testing loss was collected every epoch. The simulated results on the RAF and Fashion-MNIST datasets are shown in Fig.\ref{Fig:KFold}. As can be seen, both losses decrease quickly with the growth of the training epoch for both datasets. The curves of the losses are similar in each fold. Although there are small variations at the beginning of the training stage ($5-15$ epochs in Fig.\ref{Fig:KFold}(a) and  $10-25$ epochs in Fig.\ref{Fig:KFold}(b)), all folds converge to the optimal direction and thus are close to the average fold curve. In conclusion, the K-fold cross validation results demonstrate the robustness and effectiveness of \textit{ SiSPRNet}. Note that cross validation was only performed in this ablation study. The evaluation results in Section \ref{sec:simuresult} and \ref{sec:expresult} were obtained directly from the $1000$ samples in the testing dataset.
    
    \subsection{Simulation Results} \label{sec:simuresult}
    
    In this paper, we compare our \textit{SiSPRNet} with the recent deep learning Fourier phase retrieval methods including: Plain Network (denoted as PlainNet) \cite{He_2016_CVPR}: the basic CNN structure that stacks convolutional layers one by one; Residual Network (denoted as ResNet) \cite{He_2016_CVPR, Nishizaki_2020}: a popular deep-learning structure and has several skip connections among the convolutional layers; residual dense network (denoted as ResDenseNet) \cite{Zhang_2018_CVPR}: a network that adds more skip connections to achieve better efficiency; Lensless imaging network (denoted as LenlessNet) \cite{Sinha17}: a residual UNet to recover the phase-only images in a lensless imaging system; Conditional Generative Adversarial Phase Retrieval Network (denoted as PRCGAN) \cite{uelwer2021phase}: a network that utilizes the conditional adversarial learning strategy  for phase retrieval task; and NNPhase \cite{Wu_cw5029}: a network to recover the complex-valued images from Fraunhofer diffraction patterns. As a reference, we also compare with a few traditional optimization-based algorithms such as Gerchberg–Saxton (GS) \cite{Gerchberg1972APA}, Hybrid Input-Output (HIO) \cite{Fienup:82}, and total variation ADMM (ADMM-TV) \cite{chang2018total}.
    
    All the deep learning methods were trained with the same settings on the RAF and Fashion-MNIST datasets. For each image in the dataset, we converted it to the size of $128\times128$ and then padded zeros to make it have the size $762\times762$. The 2D-DFT was applied to generate the Fourier intensity images. Finally, the central $128\times128$ region of the intensity image was used for the training and testing of all models. All trained models were acquired at the last epoch of the training. Note that the hyper-parameters of all compared deep learning methods were either set to the defined values in their original papers or optimally fine-tuned and fixed before the training. For the traditional optimization-based methods, the full frame $762\times762$ intensity images were used as the input for their evaluation. We did three different trials of evaluation for each testing image, where each trial had  maximum $ 2000 $ iterations. The best result of the three trials was used as the final performance. For each reconstructed phase image, we removed the global phase offset by adding a global phase shift from $0$ to $2\pi$ to the image. Then, the image having the highest PSNR as compared with the ground-truth was used as the final result. Note that most of these existing deep learning-based phase retrieval methods can only accept very small-size intensity measurements. For example, PRCGAN and NNPhase can only accept intensity measurements of size $28\times28$ and $64\times64$ pixels, respectively, as shown in their original papers. We need to convert the input $762\times762$ pixels intensity measurement to the sizes these networks can handle. Downsampling does not work since it violates the oversampling requirement and the number of significant data will be substantially reduced. Similar to the proposed \textit{SiSPRNet}, we extract the central $128\times128$ pixels of the intensity measurements as the input to these networks, which is easier for them to adapt.
    
\begin{figure}[htbp]
    	\begin{adjustbox}{width=1\textwidth,center}
    		\begin{tabular}{c@{\extracolsep{0em}}c@{\extracolsep{0em}}c@{\extracolsep{0em}}c@{\extracolsep{0em}}c@{\extracolsep{0em}}c@{\extracolsep{0em}}c@{\extracolsep{0em}}c@{\extracolsep{0em}}c@{\extracolsep{0em}}c@{\extracolsep{0em}}c@{\extracolsep{0em}}}
    			
    			\includegraphics[height=0.09\textwidth, valign=t]{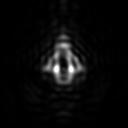}~
    			&\includegraphics[height=0.09\textwidth, valign=t]{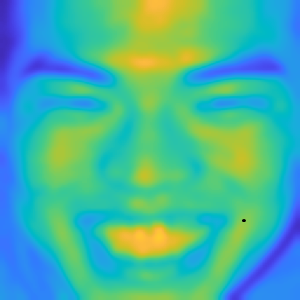}~
    			&\includegraphics[width=0.09\textwidth, valign=t]{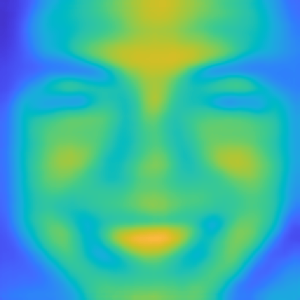}~
    			&\includegraphics[width=0.09\textwidth, valign=t]{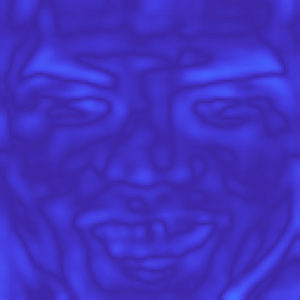}~
    			&\includegraphics[width=0.09\textwidth, valign=t]{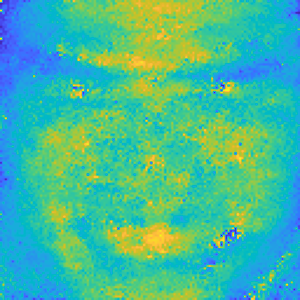}~
    			&\includegraphics[width=0.09\textwidth, valign=t]{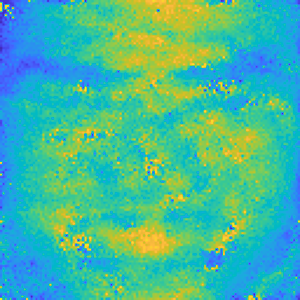}~
    			&\includegraphics[width=0.09\textwidth, valign=t]{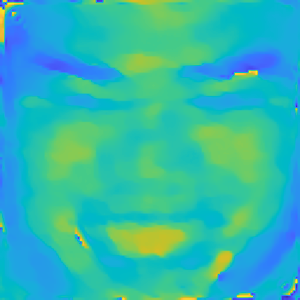}~
    			&\includegraphics[width=0.09\textwidth, valign=t]{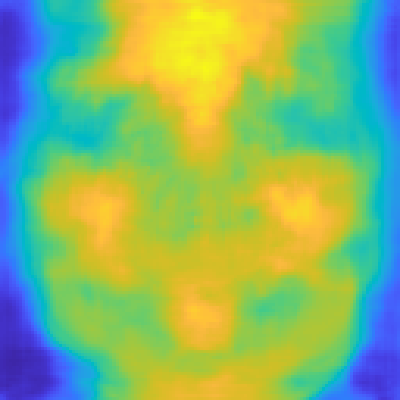}~
    			&\includegraphics[width=0.09\textwidth, valign=t]{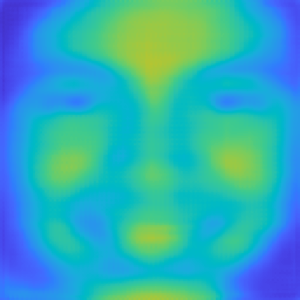}~
    			&\includegraphics[width=0.09\textwidth, valign=t]{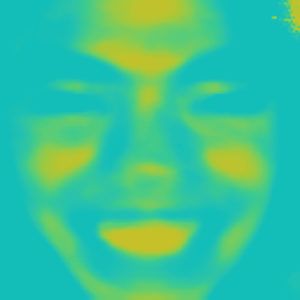}~
    			&\includegraphics[width=0.09\textwidth, valign=t]{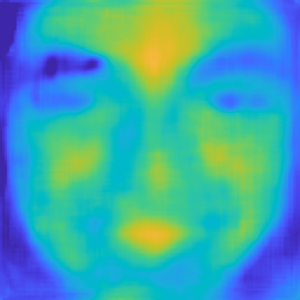}
    			\\
    			
    			\includegraphics[height=0.09\textwidth, valign=t]{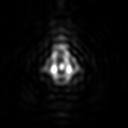}~
    			&\includegraphics[height=0.09\textwidth, valign=t]{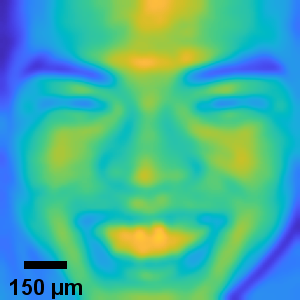}~
    			&\includegraphics[width=0.09\textwidth, valign=t]{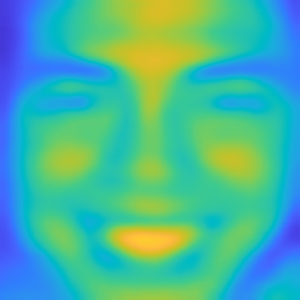}~
    			&\includegraphics[width=0.09\textwidth, valign=t]{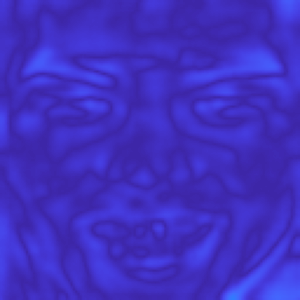}~	
    			&\includegraphics[width=0.09\textwidth, valign=t]{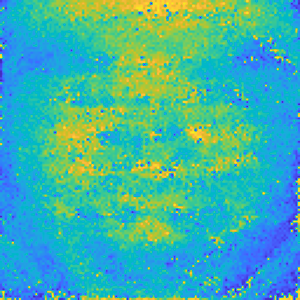}~
    			&\includegraphics[width=0.09\textwidth, valign=t]{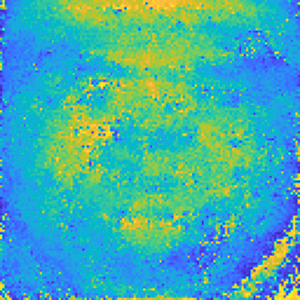}~
    			&\includegraphics[width=0.09\textwidth, valign=t]{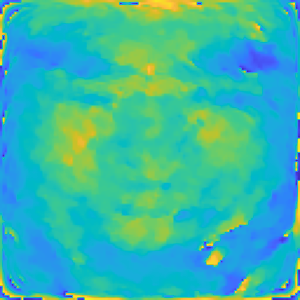}~
    			&\includegraphics[width=0.09\textwidth, valign=t]{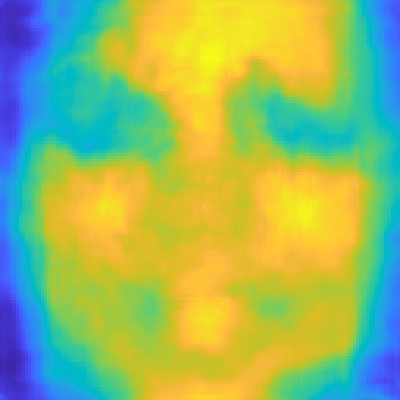}~
    			&\includegraphics[width=0.09\textwidth, valign=t]{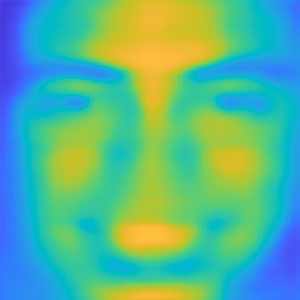}~
    			&\includegraphics[width=0.09\textwidth, valign=t]{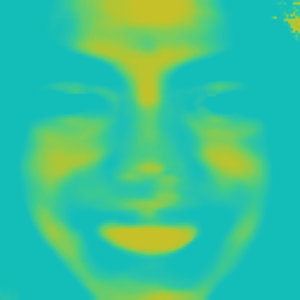}~
    			&\includegraphics[width=0.09\textwidth, valign=t]{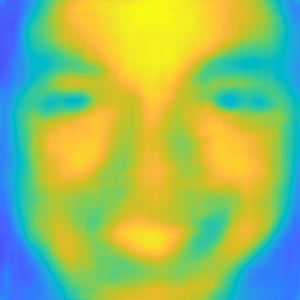}
    			
    			\\
    			
    			\includegraphics[height=0.09\textwidth, valign=t]{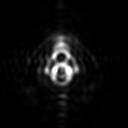}~
    			&\includegraphics[height=0.09\textwidth, valign=t]{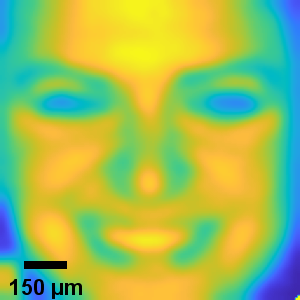}~	&\includegraphics[width=0.09\textwidth, valign=t]{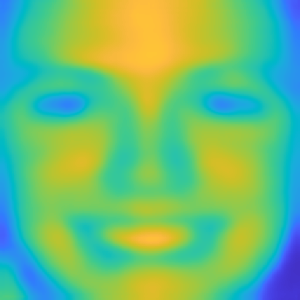}~
    			&\includegraphics[width=0.09\textwidth, valign=t]{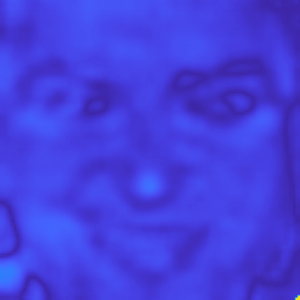}~
    			&\includegraphics[width=0.09\textwidth, valign=t]{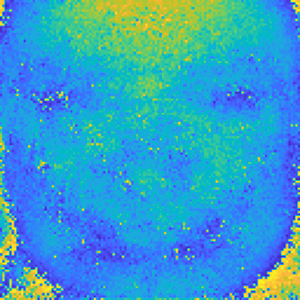}~
    			&\includegraphics[width=0.09\textwidth, valign=t]{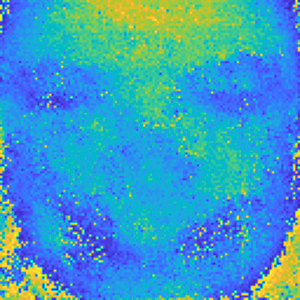}~
    			&\includegraphics[width=0.09\textwidth, valign=t]{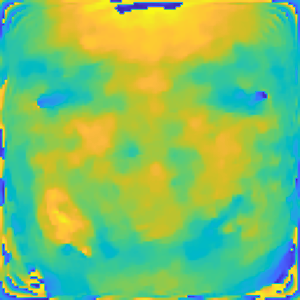}~
    			&\includegraphics[width=0.09\textwidth, valign=t]{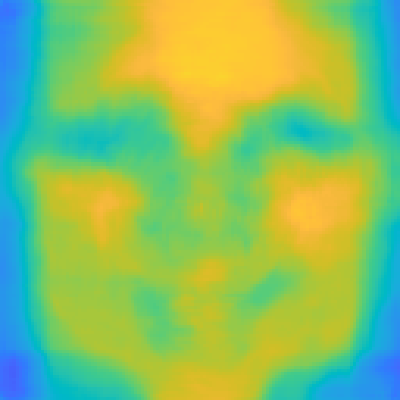}~
    			&\includegraphics[width=0.09\textwidth, valign=t]{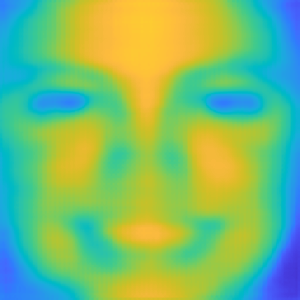}~
    			&\includegraphics[width=0.09\textwidth, valign=t]{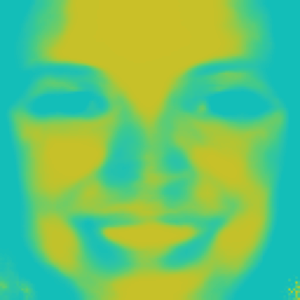}~
    			&\includegraphics[width=0.09\textwidth, valign=t]{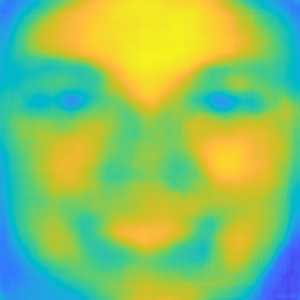}
    			\\
    			
    			\includegraphics[height=0.09\textwidth, valign=t]{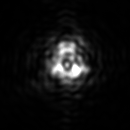}
    			&\includegraphics[height=0.09\textwidth, valign=t]{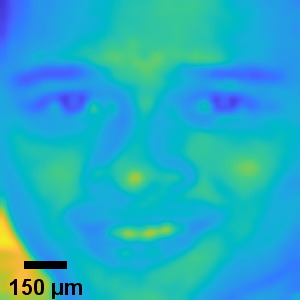}~
    			&\includegraphics[width=0.09\textwidth, valign=t, valign=t]{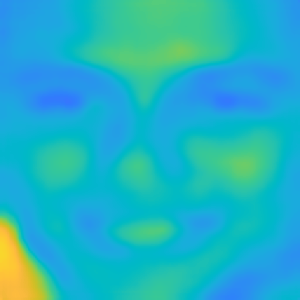}~
    			&\includegraphics[width=0.09\textwidth, valign=t, valign=t]{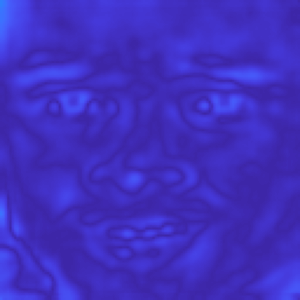}~
    			&\includegraphics[width=0.09\textwidth, valign=t, valign=t]{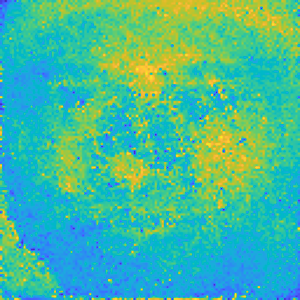}~
    			&\includegraphics[width=0.09\textwidth, valign=t, valign=t]{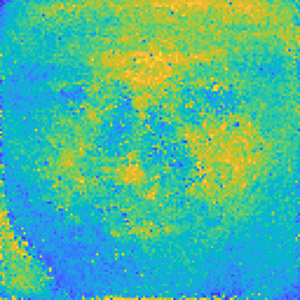}~
    			&\includegraphics[width=0.09\textwidth, valign=t, valign=t]{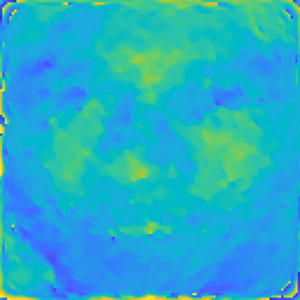}~
    			&\includegraphics[width=0.09\textwidth, valign=t, valign=t]{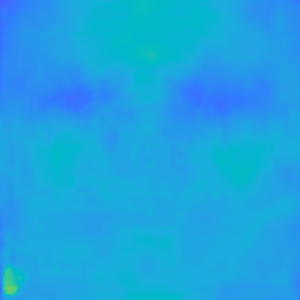}~
    			&\includegraphics[width=0.09\textwidth, valign=t, valign=t]{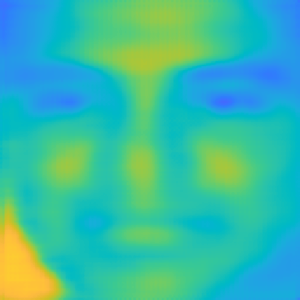}~
    			&\includegraphics[width=0.09\textwidth, valign=t, valign=t]{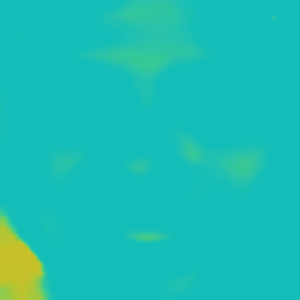}~
    			&\includegraphics[width=0.09\textwidth, valign=t, valign=t]{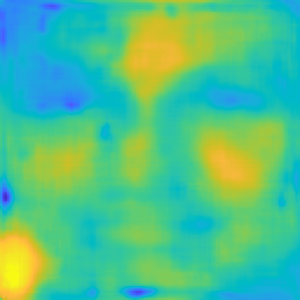}
    			\\    			
    			
    			\includegraphics[height=0.09\textwidth, valign=t]{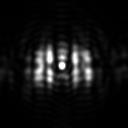}
    			&\includegraphics[height=0.09\textwidth, valign=t]{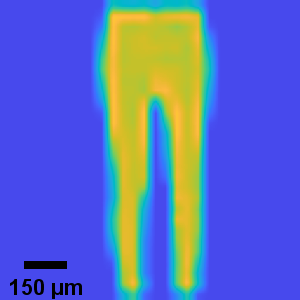}~
    			&\includegraphics[width=0.09\textwidth, valign=t]{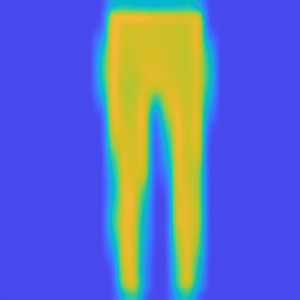}~
    			&\includegraphics[width=0.09\textwidth, valign=t]{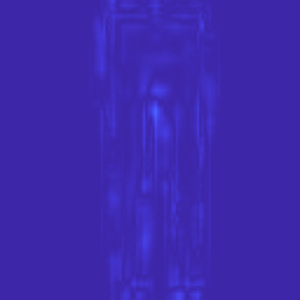}~
    			&\includegraphics[width=0.09\textwidth, valign=t]{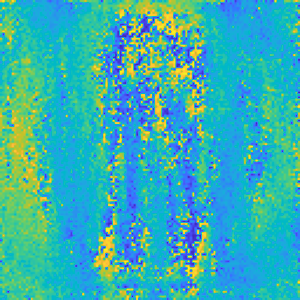}~
    			&\includegraphics[width=0.09\textwidth, valign=t]{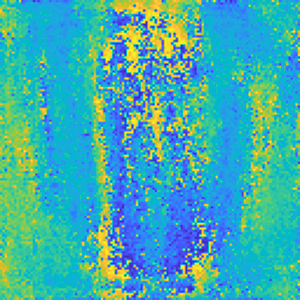}~
    			&\includegraphics[width=0.09\textwidth, valign=t]{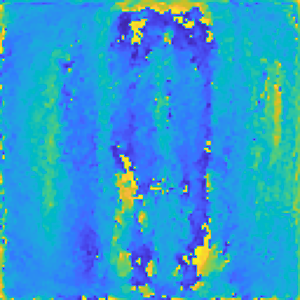}~
    			&\includegraphics[width=0.09\textwidth, valign=t]{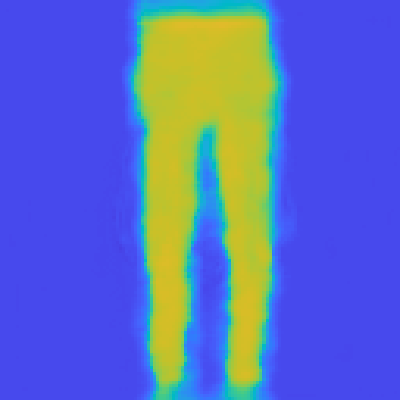}~
    			&\includegraphics[width=0.09\textwidth, valign=t]{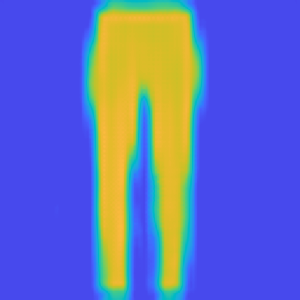}~
    			&\includegraphics[width=0.09\textwidth, valign=t]{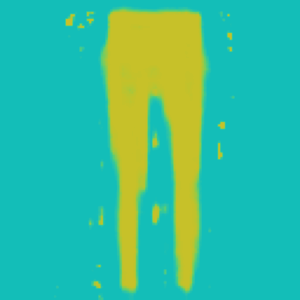}~
    			&\includegraphics[width=0.09\textwidth, valign=t]{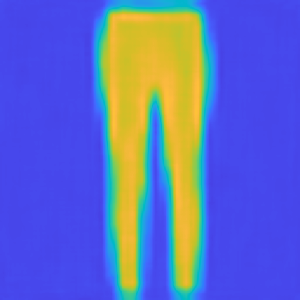}
    			\\
    			
    			\includegraphics[height=0.09\textwidth, valign=t]{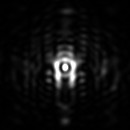}
    			&\includegraphics[height=0.105\textwidth, width=0.09\textwidth, valign=t]{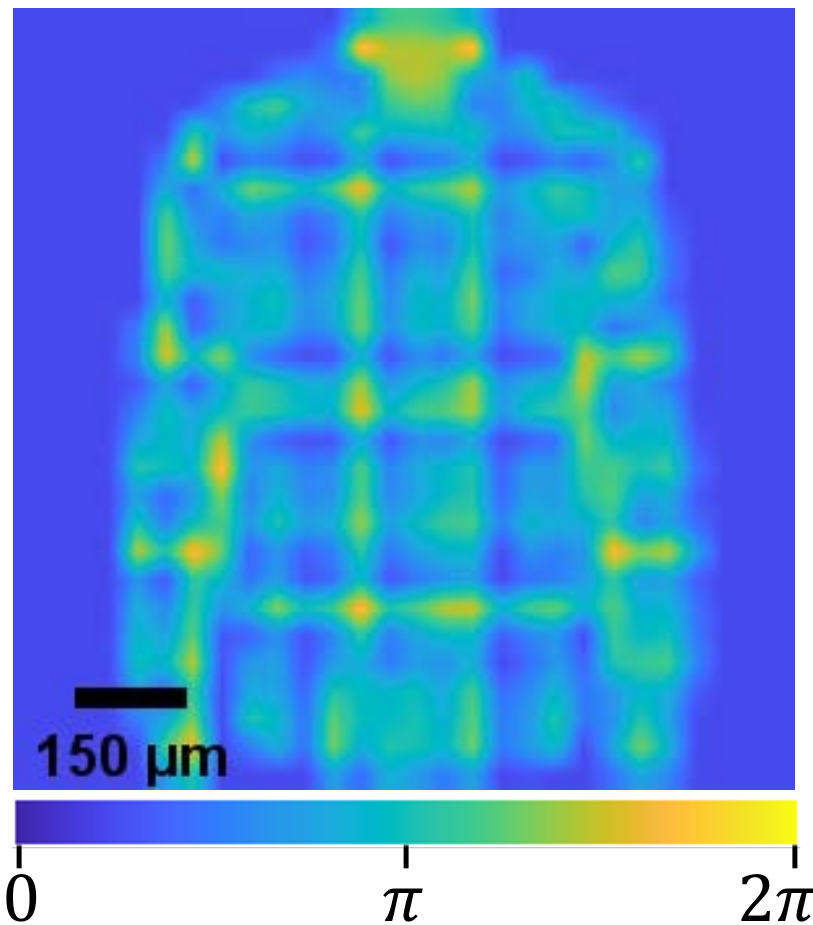}~         &\includegraphics[width=0.09\textwidth, valign=t]{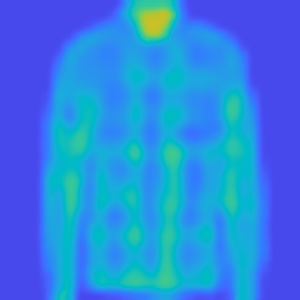}~
    			&\includegraphics[width=0.09\textwidth, valign=t]{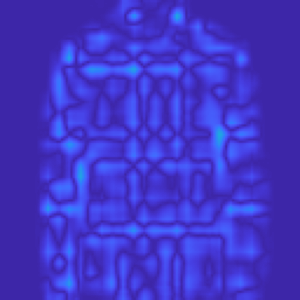}~
    			&\includegraphics[width=0.09\textwidth, valign=t]{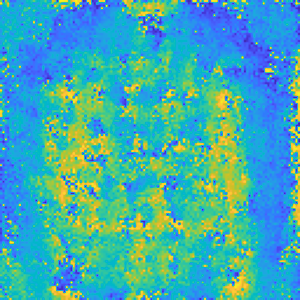}~
    			&\includegraphics[width=0.09\textwidth, valign=t]{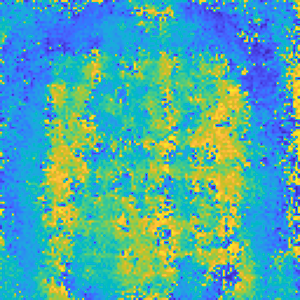}~
    			&\includegraphics[width=0.09\textwidth, valign=t]{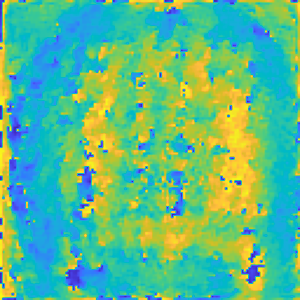}~
    			&\includegraphics[width=0.09\textwidth, valign=t]{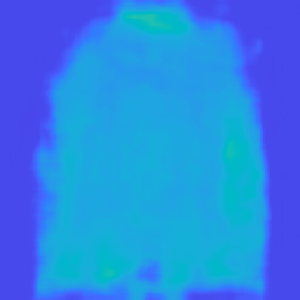}~
    			&\includegraphics[width=0.09\textwidth, valign=t]{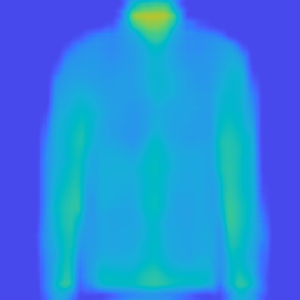}~
    			&\includegraphics[width=0.09\textwidth, valign=t]{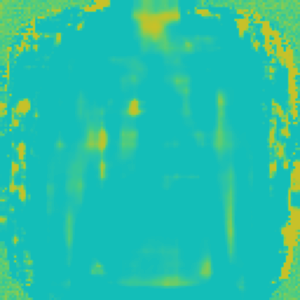}~
    			&\includegraphics[width=0.09\textwidth, valign=t]{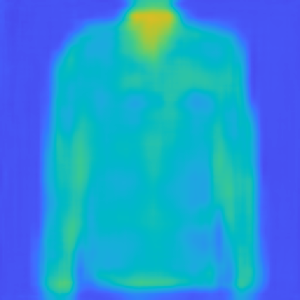}
    			\\    
    			
    			{\scriptsize \makecell[c]{Fourier\\measurement}} & {\scriptsize \makecell[c]{Ground-truth \\ (phase)}} & {\scriptsize \makecell[c]{\textbf{ \textit{SiSPRNet} }\\(ours)} }& {\scriptsize \makecell[c]{\textbf{ \textit{SiSPRNet}} \\(Error Map)} }  & {\scriptsize GS } & {\scriptsize HIO} & {\scriptsize ADMM-TV} & {\scriptsize ResNet}
    			& {\scriptsize LenslessNet}  & {\scriptsize PRCGAN}  & {\scriptsize NNPhase}  
    			\\
    		\end{tabular}
    	\end{adjustbox}
    	\vspace{-0.2cm}
    	\caption{Simulation and experimental results of different phase retrieval methods on the RAF \cite{li2017reliable} and the Fashion-MNIST datasets \cite{xiao2017/online}. The first \textbf{column} denotes the Fourier intensity measurements (pixel values: $0 \rightarrow 4095$). The second \textbf{column} shows the corresponding phase parts of the complex-valued images with scale bars (for experimental results) at the bottom left corners. The fourth \textbf{column} denotes the error maps of \textit{SiSPRNet}. The other \textbf{columns} present the reconstructed images through different methods. Except for the Fourier intensity measurements (the first \textbf{column}), the colormap of the rest \textbf{columns} ranges from $0$ to $2\pi$. The first and second \textbf{rows} are the simulation and experimental results of the same image from the RAF dataset. The third and the fourth \textbf{rows} show the experimental performances of other images in the RAF dataset. The fifth and the sixth \textbf{rows} indicate the experimental results of images from the Fashion-MNIST dataset. Please zoom in for better view. More results can be found in \nameref{Sec:Moreresults}. }
    	\label{fig:compvisual}
    \end{figure}

    \textbf{\textit{Phase-only datasets:}} The phase retrieval results with the phase-only datasets are shown in Table \ref{table:RAFresult}, Table \ref{table:Fashionresult}, and Fig. \ref{fig:compvisual}. To save space, the qualitative results of PlainNet and ResDenseNet are not included in Fig. \ref{fig:compvisual} due to their inferior performances. As can be seen in Table \ref{table:RAFresult} and Table \ref{table:Fashionresult}, the proposed \textit{SiSPRNet} achieves much better performance than all compared methods. Compared with other state-of-the-art deep learning methods, SiSPRNet achieves PSNR and SSIM gains of at least $1.631$ dB and $0.0656$, respectively, on the two datasets. It can be seen in Fig. \ref{fig:compvisual} that the reconstructed images of the proposed \textit{SiSPRNet} outperform other deep learning methods qualitatively. In addition, the error maps ($4$th column) also show that the estimated images are close to the ground-truth images. As can be seen, the reconstructed images only differ from the original ones in some local areas with minor errors, while most regions are close to $0$. On the other hand, SiSPRNet also achieves a significantly better performance than all compared traditional optimization-based methods as shown in Table \ref{table:RAFresult}, Table \ref{table:Fashionresult} and Fig. \ref{fig:compvisual}. 
    
    Note that we train all the networks with our quantized defocused Fraunhofer diffraction patterns datasets. Although the reconstructed images via PRCGAN \cite{uelwer2021phase} seem to have better high-frequency information (facial details), the reconstructed image does not share similarities with the original image (last second column).  In fact, the performance of PRCGAN in our simulation is poorer than that reported in the original paper of PRCGAN. It is because the dataset used in the original simulation of PRCGAN does not consider the saturation problem, which is commonly found when imaging the Fourier intensity. Besides, the authors assume the object images are magnitude-only (no phase information), which is far from real situations.  As for NNPhase \cite{Wu_2021}, it aims to deal with the crystal data that the application is quite specific. Besides, LenslessNet \cite{Sinha17} is initially designed for the phase retrieval task based on Fresnel diffraction patterns with lensless imaging systems. When testing with the more general and realistic datasets used in our simulations, the performances of these networks drop significantly compared with those reported in their original papers. We will show in the next section that these networks also suffer from the same problems in the real experiments. The above simulation results show that the proposed \textit{SiSPRNet} outperforms the traditional optimization-based and deep learning methods both quantitatively and qualitatively. More simulation results can be found in \nameref{Sec:Moreresults}. 
    
    We also evaluate the inference speed of different methods by collecting the mean execution time of one hundred samples. The results are presented in Table \ref{table:RAFresult} (last column). Since our approach is designed in an end-to-end manner that can effectively utilize the huge computational power of the current advanced GPU devices, it can be seen in Table \ref{table:RAFresult} that the proposed \textit{SiSPRNet} is very fast and can achieve real-time performance ($3.569$ ms/img). It is comparable with other learning-based approaches. In contrast, the inference speed of the traditional algorithms is quite slow (in the order of seconds) since lots of iterations are required, and they can only be implemented with the CPU. In addition, we assessed the number of learnable parameters and model complexity of all compared deep learning methods. The results are shown in Table \ref{table:Fashionresult}. As can be seen, the proposed \textit{SiSPRNet} has a moderate number of learnable parameters and low model complexity (measured in terms of Giga floating-point operations per second, GFLOPs) among all approaches. Compared with the deep learning approaches having similar or more parameters, \textit{SiSPRNet} has better reconstruction performance. It should be noted that the generator of the PRCGAN contains more than $100M$ parameters but could not give high accuracy, which might be the result of overfitting. From the perspective of complexity, \textit{SiSPRNet} is lower than most of the networks but has much better performance.
    
\begin{figure}[htb]
    	
    	\begin{subfigure}{\textwidth}	
    		\centering
    		\begin{adjustbox}{width=1\textwidth, center}        \begin{tabular}{c@{\extracolsep{0em}}c@{\extracolsep{0em}}c@{\extracolsep{0em}}c@{\extracolsep{0em}}c@{\extracolsep{0em}}c@{\extracolsep{0em}}c@{\extracolsep{0em}}c@{\extracolsep{0em}}c@{\extracolsep{0em}}c@{\extracolsep{0em}}c@{\extracolsep{0em}}}
    				
    				\includegraphics[width=0.1\textwidth, valign=t]{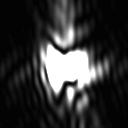}~
    				&\includegraphics[width=0.1\textwidth, valign=t]{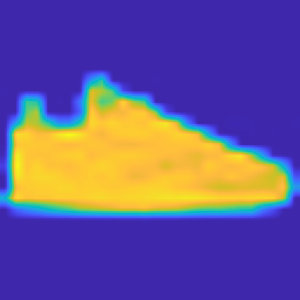}~
    				&\includegraphics[width=0.1\textwidth, valign=t]{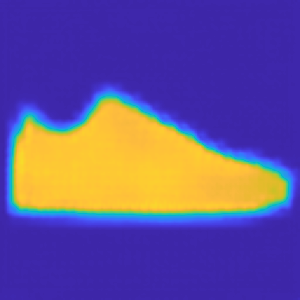}~
    				&\includegraphics[width=0.1\textwidth, valign=t]{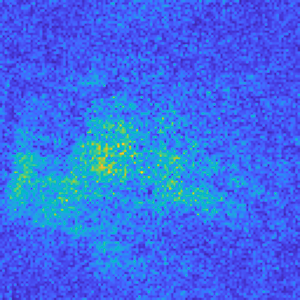}~
    				&\includegraphics[width=0.1\textwidth, valign=t]{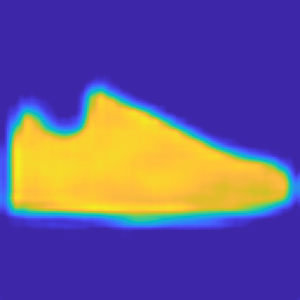}~
    				&\includegraphics[width=0.1\textwidth, valign=t]{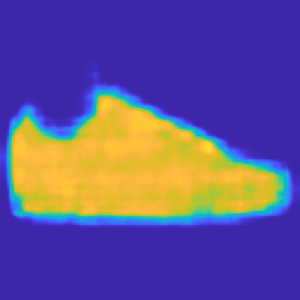}~
    				&\includegraphics[width=0.1\textwidth, valign=t]{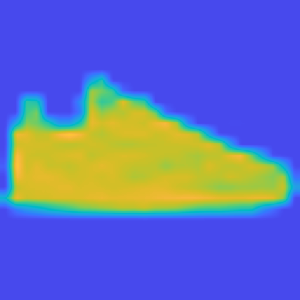}~
    				&\includegraphics[width=0.1\textwidth, valign=t]{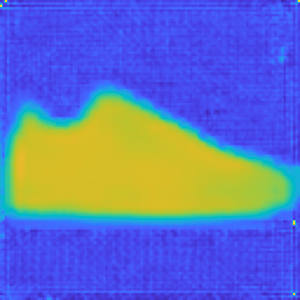}~
    				&\includegraphics[width=0.1\textwidth, valign=t]{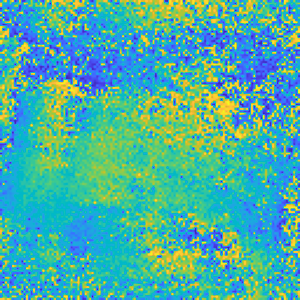}~
    				&\includegraphics[width=0.1\textwidth, valign=t]{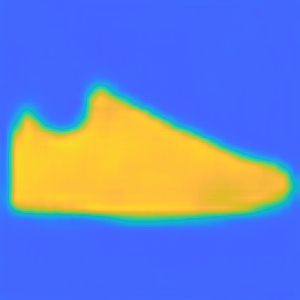}~
    				&\includegraphics[width=0.1\textwidth, valign=t]{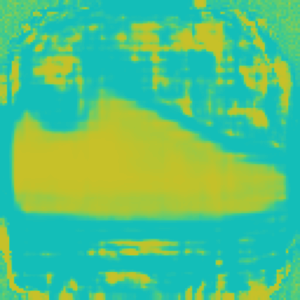}~
    				\\
    				
    				\includegraphics[width=0.1\textwidth, valign=t]{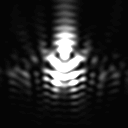}~
    				&\includegraphics[width=0.1\textwidth, valign=t]{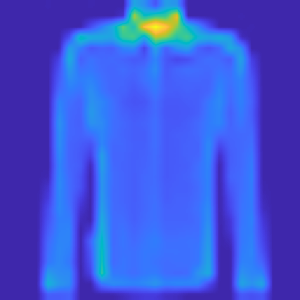}~
    				&\includegraphics[width=0.1\textwidth, valign=t]{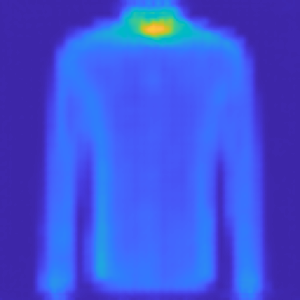}~
    				&\includegraphics[width=0.1\textwidth, valign=t]{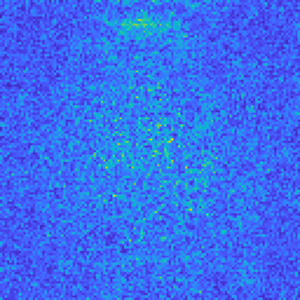}~
    				&\includegraphics[width=0.1\textwidth, valign=t]{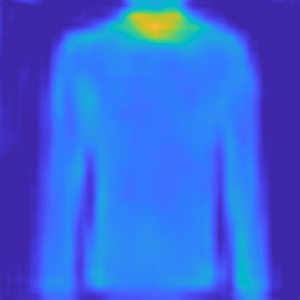}~
    				&\includegraphics[width=0.1\textwidth, valign=t]{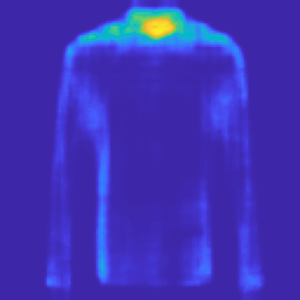}~
    				&\includegraphics[width=0.1\textwidth, valign=t]{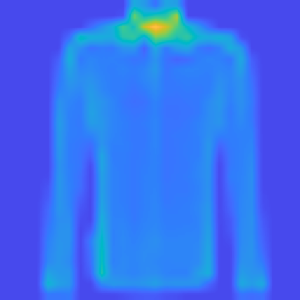}~
    				&\includegraphics[width=0.1\textwidth, valign=t]{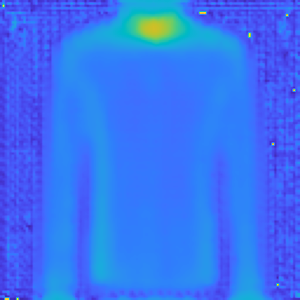}~
    				&\includegraphics[width=0.1\textwidth, valign=t]{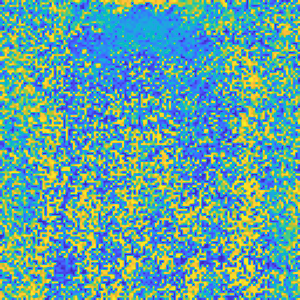}~
    				&\includegraphics[width=0.1\textwidth, valign=t]{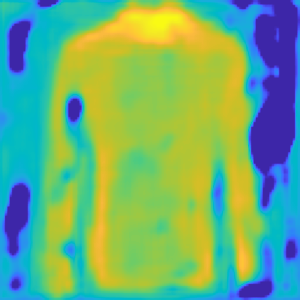}~
    				&\includegraphics[width=0.1\textwidth, valign=t]{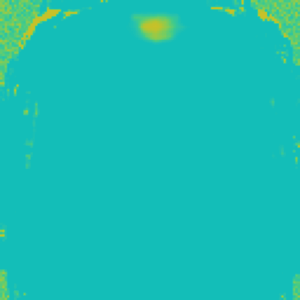}~
    				\\
    				
    				\includegraphics[width=0.1\textwidth, valign=t]{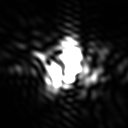}~
    				&\includegraphics[height = 0.118\textwidth,width=0.1\textwidth, valign=t]{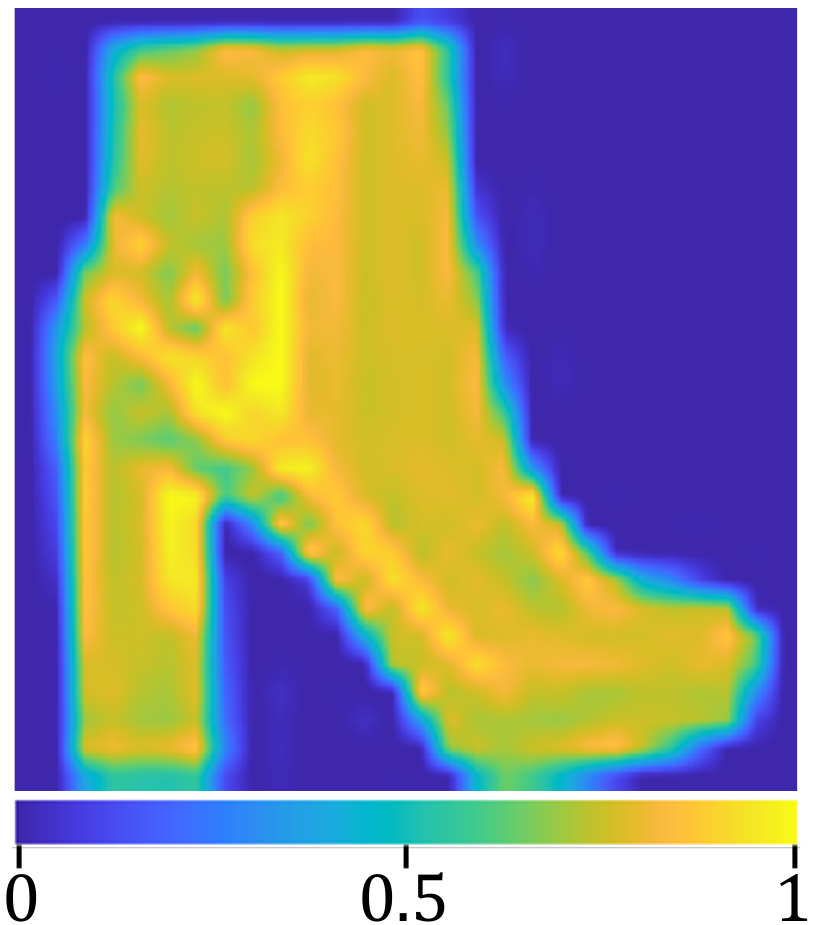}~
    				&\includegraphics[width=0.1\textwidth, valign=t]{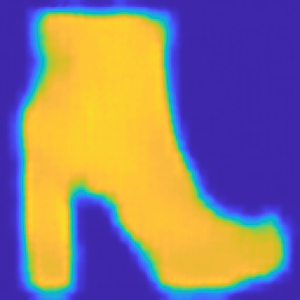}~
    				&\includegraphics[width=0.1\textwidth, valign=t]{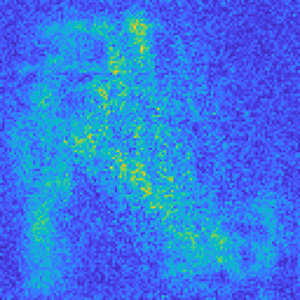}~
    				&\includegraphics[width=0.1\textwidth, valign=t]{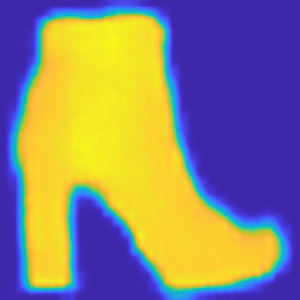}~
    				&\includegraphics[width=0.1\textwidth, valign=t]{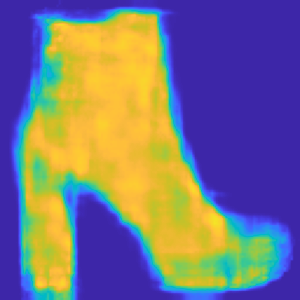}~
    				&\includegraphics[height = 0.118\textwidth,width=0.1\textwidth, valign=t]{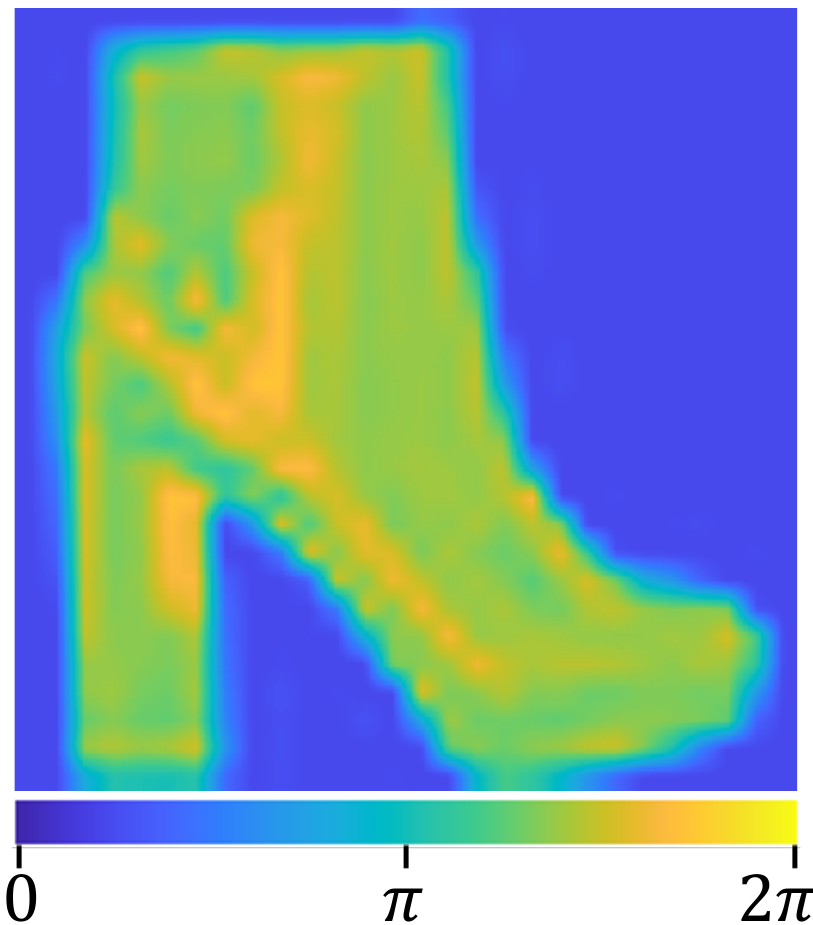}~
    				&\includegraphics[width=0.1\textwidth, valign=t]{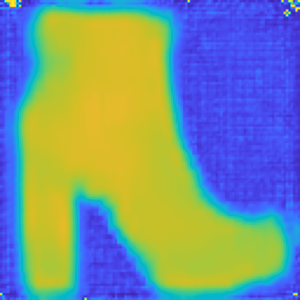}~		&\includegraphics[width=0.1\textwidth, valign=t]{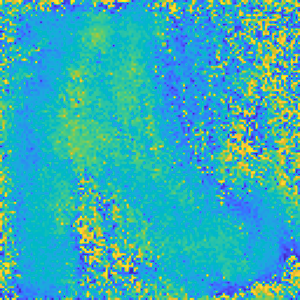}~
    				&\includegraphics[width=0.1\textwidth, valign=t]{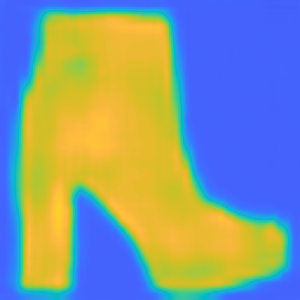}~
    				&\includegraphics[width=0.1\textwidth, valign=t]{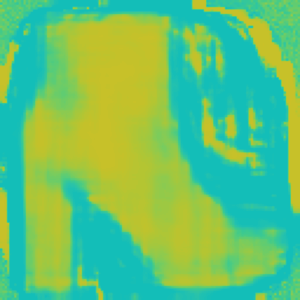}~
    				\\
    				
    				{\scriptsize \makecell[c]{Fourier\\measurement}} & {\scriptsize \makecell[c]{Ground-truth \\ (mag.)}} & {\scriptsize \makecell[c]{\textbf{ \textit{SiSPRNet} }  \\ (mag.)}} & {\scriptsize \makecell[c]{HIO \\ (mag.)}} & {\scriptsize \makecell[c]{NNPhase \\ (mag.)}} & {\scriptsize \makecell[c]{PRCGAN \\ (mag.)}} & {\scriptsize \makecell[c]{Ground-truth \\ (phase)}} & {\scriptsize \makecell[c]{\textbf{ \textit{SiSPRNet}} \\ (phase)}} & {\scriptsize \makecell[c]{HIO \\ (phase)}} & {\scriptsize \makecell[c]{NNPhase \\ (phase)}} & {\scriptsize \makecell[c]{PRCGAN \\ (phase)}} 
    				\\
    			\end{tabular}
    		\end{adjustbox}
    		\vspace*{-2mm}
    		\caption{}
    	\end{subfigure}
    	
    	\begin{subfigure}{\textwidth}	
    		\centering
    		\begin{adjustbox}{width=0.7\columnwidth, center}
    			\begin{tabular}{|c|c|c|c|c|c|c|}
    				\hline \multirow{2}{*}{ Methods } & \multicolumn{2}{c|}{ MAE $\downarrow$} & \multicolumn{2}{c|}{ PSNR $\uparrow$} & \multicolumn{2}{c|}{ SSIM $\uparrow$} \\
    				\cline { 2 - 7 } & Mag. & Phase & Mag. & Phase & Mag. & Phase \\
    				\hline
    				\hline HIO \cite{Fienup:82} & $0.297$ & $1.894$ &  $7.710$ & $8.338$ & $0.329$ & $0.044$ \\
    				PRCGAN \cite{uelwer2021phase} & $0.073$ & $1.654$ &  $18.672$ & $9.924$ & $0.588$ & $0.387$ \\
    				NNPhase \cite{Wu_cw5029} & $0.109$ & $0.687$ & $18.169$ & $18.752$ & $0.395$ & $0.46$ \\
    				\textbf{ \textit{SiSPRNet}  (Ours)} & $\mathbf{0.053}$ & $\mathbf{0.359}$ & $\mathbf{20.817}$ & $\mathbf{21.966}$ & $\mathbf{0.744}$ & $\mathbf{0.594}$   \\
    				\hline
    			\end{tabular}
    		\end{adjustbox}
    		\caption{}
    	\end{subfigure}
    	
    	\vspace{-0.2cm}
    	\caption{(a) Qualitative simulation results of different phase retrieval methods on the magnitude-phase Fashion-MNIST datasets \cite{xiao2017/online}. The pixel values of the Fourier intensity measurements (first column) range from $0$ to $4095$. The colormap of the magnitude parts (second to the sixth columns) ranges from $0$ to $1$, while the colormap of the phase parts (seventh to the last columns) ranges from $0$ to $2\pi$. Please zoom in for better view. (b) Quantitative comparisons (average MAE/PSNR/SSIM) on the same dataset. Best performances are denoted in \textbf{bold} font.}
    	\label{fig:ampphase}
    	
    \end{figure}

    \textbf{\textit{Magnitude-phase dataset:}} The simulation results are shown in Fig. \ref{fig:ampphase}. We select HIO, NNPhase, and PRCGAN for comparison since they all perform Fourier phase retrieval. It can be seen that the proposed \textit{ SiSPRNet} outperforms all compared approaches. In particular, the results show that PRCGAN cannot deal with complex-valued phase retrieval tasks, and it is difficult for HIO to reconstruct complex-valued images with single-shot intensity measurement. It is particularly the case since, in the images, the magnitude surrounding the objects is very small (or very close to zero). It introduces ambiguities to the spatial constraint that affects the estimation process of HIO. The magnitude and phase images given by these approaches have poor quality. As presented in Fig. \ref{fig:ampphase}(b), the proposed \textit{SiSPRNet} achieves much better performance than all compared methods evaluated by different metrics. For the magnitude part, the proposed \textit{SiSPRNet} achieves average PSNR and SSIM gains of at least $2.145dB$ and $0.156$, respectively. For the phase part, the increases in PSNR and SSIM gains can reach $3.214 dB$ and $0.181$, respectively. The above simulation results show that the proposed \textit{SiSPRNet} can be generalized to complex-valued image datasets (containing linearly related magnitude and phase parts) and outperform the existing methods.
    
        \begin{table}[ht]
    	\centering
    	\caption{Quantitative comparison (average MAE/PSNR/SSIM/Inference Time) with the state-of-the-art methods for phase retrieval on the RAF dataset. Best and second best performances are in \textcolor{red}{red} and \textcolor{blue}{blue} colors, respectively. }
    	\begin{adjustbox}{width=0.8\columnwidth, center}
    		\begin{tabular}{|c|c|c|c|c|c|c|c|}
    			\hline \multirow{2}{*}{ Methods } & \multicolumn{2}{c|}{ MAE $\downarrow$} & \multicolumn{2}{c|}{ PSNR $\uparrow$} & \multicolumn{2}{c|}{ SSIM $\uparrow$} & \multirow{2}{*}{\parbox{1.2cm}{\centering Inference Time (ms/img) }}\\
    			\cline { 2 - 7 } & Sim. & Exp. & Sim. & Exp. & Sim. & Exp. & \\
    			\hline
    			\hline 	GS  \cite{Gerchberg1972APA} & $0.961$ & $1.136$ & $15.252$ & $12.816$ & $0.247$ & $0.126$ & $1.273\times 10^{3}$\\
    			HIO \cite{Fienup:82} & $1.024$ & $1.221$ & $14.718$ & $12.188$ & $0.233$ & $0.127$ & $1.116\times 10^{3}$\\
    			ADMM-TV  \cite{chang2018total} &  $0.783$ & $1.066$ & $16.574$ & $13.637$ & $0.341$ & $0.128$ & $1.573\times 10^{4}$\\
    			PlainNet \cite{He_2016_CVPR} & $0.975$ & $1.180$ & $13.841$ & $12.479$ & $0.425$ & $0.377$ & $1.583$\\
    			ResNet \cite{He_2016_CVPR, Nishizaki_2020} & $0.894$ & $0.812$ & $15.178$ & $15.831$ & $0.456$ & $0.471$ & $2.438$\\
    			ResDenseNet \cite{Zhang_2018_CVPR} & $1.084$ & $1.010$ & $13.344$ & $13.800$ & $0.380$ & $0.387$ & $1.732$\\
    			LenslessNet \cite{Sinha17} & $\textcolor{blue}{0.671}$ & $\textcolor{blue}{0.704}$ & $\textcolor{blue}{17.573}$ & $\textcolor{blue}{17.132}$ & $\textcolor{blue}{0.598}$ & $\textcolor{blue}{0.596}$ & $7.396$\\
    			PRCGAN \cite{uelwer2021phase} & $0.951$ & $0.945$ & $14.738$ & $14.833$ & $0.511$ & $0.508$ & $0.983$\\
    			NNPhase \cite{Wu_cw5029} & $0.755$ & $0.931$ & $16.753$ & $15.416$ & $0.558$ & $0.544$ & $5.834$\\
    			\textbf{\textit{SiSPRNet}  (Ours)} & $\textcolor{red}{0.570}$ & $\textcolor{red}{0.582}$ & $\textcolor{red}{19.204}$ & $\textcolor{red}{18.905}$ & $\textcolor{red}{0.663}$ & $\textcolor{red}{0.663}$ & $3.569$\\
    			\hline
    		\end{tabular}
    	\end{adjustbox}
    	\label{table:RAFresult}
    \end{table}
    
        \begin{table}[ht]
    	\centering
    	\caption{Quantitative comparison (average MAE/PSNR/SSIM/Parameters/Complexity) with the state-of-the-art methods for phase retrieval on the Fashion-MNIST dataset. Best and second best performances are in \textcolor{red}{red} and \textcolor{blue}{blue} colors, respectively.}
    	\begin{adjustbox}{width=0.9\columnwidth,center}
    		\begin{tabular}{|c|c|c|c|c|c|c|c|c|}
    			\hline \multirow{2}{*}{ Methods } & \multicolumn{2}{c|}{ MAE $\downarrow$} & \multicolumn{2}{c|}{ PSNR $\uparrow$} & \multicolumn{2}{c|}{ SSIM $\uparrow$} & \multirow{2}{*}{\parbox{1.6cm}{\centering Parameters (Millions) }} & \multirow{2}{*}{\parbox{1.6cm}{\centering Complexity (GFLOPs) }} \\
    			\cline { 2 - 7 } & Sim. & Exp. & Sim. & Exp. & Sim. & Exp. & & \\
    			\hline
    			GS \cite{Gerchberg1972APA} & $1.567$ & $1.678$ & $9.841$ & $9.162$ & $0.088$ & $0.052$ & NA & NA \\
    			HIO \cite{Fienup:82} & $1.601$ & $1.719$ & $9.583$ & $8.915$ & $0.086$ & $0.054$ & NA & NA\\
    			ADMM-TV \cite{chang2018total} & $1.086$ & $1.504$ & $13.404$ & $9.747$ & $0.221$ & $0.125$ & NA & NA \\
    			PlainNet \cite{He_2016_CVPR} & $0.589$ & $0.601$ & $16.916$ & $16.672$ & $0.629$ & $0.623$ & $0.17$ & $2.81$\\
    			ResNet \cite{He_2016_CVPR, Nishizaki_2020} & $0.289$ & $0.304$ & $21.394$ & $20.806$ & $0.739$ & $0.726$ & $0.17$ & $2.88$ \\
    			ResDenseNet \cite{Zhang_2018_CVPR} & $0.381$ & $0.419$ & $19.406$ & $18.252$ & $0.675$ & $0.657$  & $0.17$ & $2.83$\\
    			LenslessNet \cite{Sinha17} & $\textcolor{blue}{0.179}$ & $\textcolor{blue}{0.221}$ & $\textcolor{blue}{26.784}$ & $\textcolor{blue}{23.833}$ & $\textcolor{blue}{0.842}$ & $\textcolor{blue}{0.802}$ & $20.54$ & $2.49$\\
    			PRCGAN \cite{uelwer2021phase} & $1.598$ & $1.602$ &  $10.102$ & $10.126$ & $0.422$ & $0.417$ & $111.20$ & $0.11$\\
    			NNPhase \cite{Wu_cw5029} & $0.404$ & $0426$ & $23.956$ & $22.778$ & $0.739$ & $0.443$ & $18.85$ & $6.49$\\
    			\textbf{ \textit{SiSPRNet} (Ours)} & $\textcolor{red}{0.127}$ & $\textcolor{red}{0.144}$ & $\textcolor{red}{29.209}$ & $\textcolor{red}{28.132}$ & $\textcolor{red}{0.884}$ & $\textcolor{red}{0.872}$ & $19.30$ & $1.38$  \\
    			\hline
    		\end{tabular}
    	\end{adjustbox}
    	\label{table:Fashionresult}
    \end{table}
    
    
        \section{Experimental Results} \label{Sec:experiment}
    
    \subsection{Experimental Setup}
    
\begin{figure}[ht]               
    	\centering\includegraphics[width=0.8\linewidth]{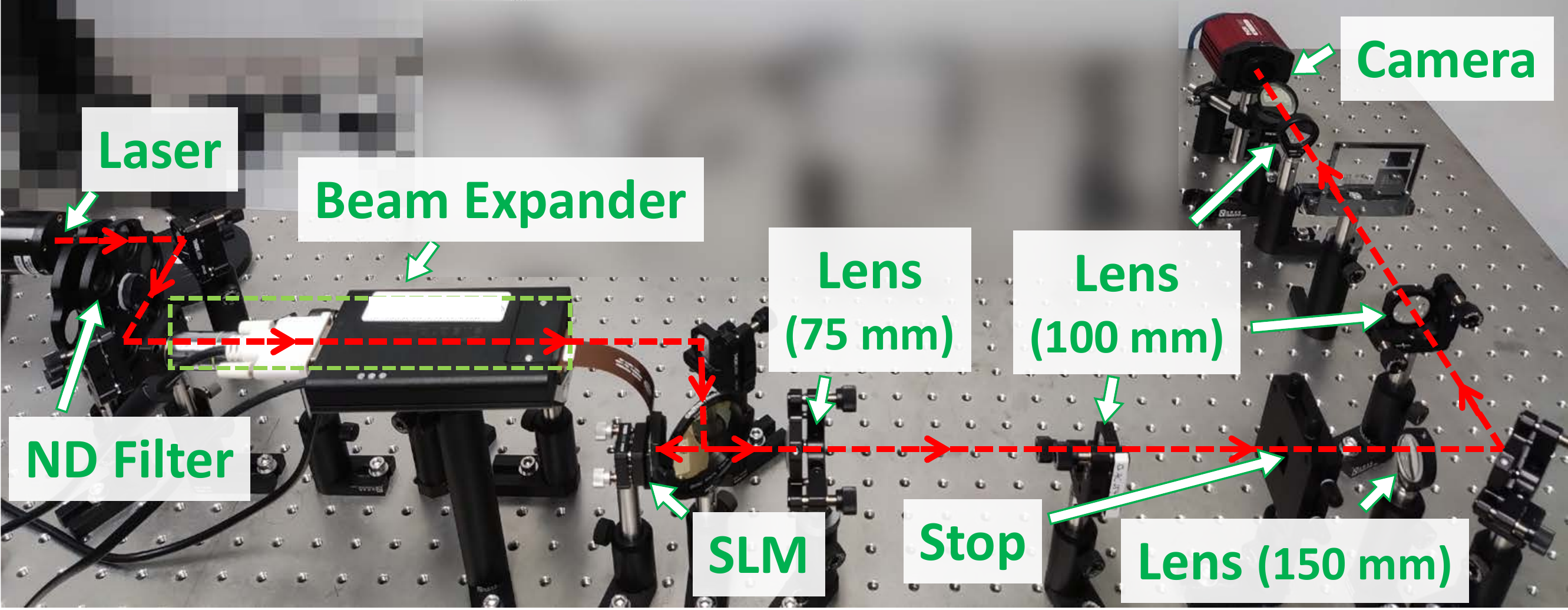}
    	\caption{Experimental setup of the defocus phase retrieval system.}
    	\label{fig:setup}
    	
    \end{figure}

    We also evaluated our proposed \textit{ SiSPRNet} with a practical optical system. A picture of the experimental setup is shown in Fig. \ref{fig:setup}.  It comprises a Thorlabs $ 10 mW$ HeNe laser with wavelength $ \lambda = 632.8 nm$ ; and a $ 12 $-bit $ 1920\times1200 $ Kiralux CMOS camera with pixel pitch $ 5.86 \mu m$ . Besides, the optical system includes one $75 mm$, three $100 mm$, and one $150 mm$ lenses to form a standard imaging system. The choice of the lenses is for generating Fourier intensity images of appropriate size to match the size of the CMOS sensor. In practice, these lenses can be compactly arranged such that the whole imaging system can be made very small. However, the lens design is out of the scope of this study. A $ 1920\times1080 $ Holoeye Pluto phase-only SLM with pixel pitch $ \delta_{SLM}=8\mu m $ was used to generate the testing objects in the experiments. Specifically, the images in the two datasets mentioned above were loaded to the SLM to impose multiple-level phase changes. The images were pre-multiplied with a defocus kernel; thus, there was no need to move the camera beyond the focal plane. It is to ensure that the defocusing process, which is not our main focus, will be perfectly implemented so that we only need to focus on the reconstruction performances of different methods. The defocus kernel is a pure-phase object generated by Holoeye built-in function corresponding to $L = 30 mm$. The reason for choosing $L = 30mm$ has been explained in Section \ref{Sec:abladefocusdist}. We utilized a square stop to crop the central $ 128\times 128 $ SLM pixels in all experiments. The resulting Fourier intensity measurements were captured by the camera and formed the training and testing datasets for \textit{SiSPRNet}. The size of the acquired measurement is $\frac{\lambda L}{d\delta_{SLM}} = 762 \times 762$ where $d$ denotes the pixel pitch of the camera. Thus, the scaled measurement is equivalent to the Fourier intensity of zero-padding images. Similar to the simulation, the central $128\times128$ regions of the intensity images were used as the training and testing samples for all deep learning methods. For the optimization-based methods, the full $762\times762$ intensity images were used as the input. 
    
    \subsection{Experimental Results} \label{sec:expresult}
    
    With the training dataset, we fine-tuned all the pre-trained models acquired in the simulation with the experimental measurements. That is, we re-trained the models with the trained parameters from the simulation and used the experimental measurements as the new training inputs. We used the same training and testing datasets (RAF and Fashion-MNIST) used in simulation. Total epoch of the re-training procedure was set as $150$. The quantitative and qualitative experimental results are presented in Table \ref{table:RAFresult}, Table \ref{table:Fashionresult}, and Fig. \ref{fig:compvisual}, respectively. 
    
    It is shown in Table \ref{table:RAFresult} and Table \ref{table:Fashionresult} that the experimental results of \textit{SiSPRNet} only decline a little compared with the simulation results. It proves the robustness of the proposed method. Compared with other state-of-the-art deep learning methods, \textit{SiSPRNet} can achieve at least $1.773dB$ and $0.067$ gains in PSNR and SSIM, respectively, on the two datasets. As can be clearly seen in Fig. \ref{fig:compvisual}, the proposed \textit{SiSPRNet} can reconstruct the most similar contours as the ground-truth and preserve more details compared with other methods. The above experimental results show that the proposed \textit{SiSPRNet} outperforms all {deep learning} methods quantitatively and qualitatively when using in practical phase retrieval systems.  On the other hand, \textit{SiSPRNet} also improves significantly over the compared traditional optimization-based methods. However, there is always room for improvement. It is shown in the results of all learning-based approaches that the reconstructed images are a bit blurry compared with the ground-truth. We note that for the captured intensity images, the values of the high-frequency components are much smaller than those in the low-frequency areas. The noise in the images further makes the high-frequency components difficult to identify. It introduces much difficulty to the neural networks to extract the features of the high-frequency components. Further investigation is needed to improve the sharpness of the reconstructed images. Extra denoising and attention modules may be applied to better recover the high-frequency components.
    
        \section{Conclusion} 
    This paper proposed an end-to-end deep neural network structure for single-shot maskless phase retrieval. We have explained how we constructed our Fourier phase retrieval optical path  to ensure the feasibility of the solution and reduce the saturation problem due to the high dynamic range of Fourier intensity measurements. We proposed a novel deep neural network structure named \textit{SiSPRNet} to reconstruct phase images using the central areas of the intensity measurements. To fully extract the representative features, we proposed a new feature extraction unit using MLP as the front end. It was tailor-designed for extracting the representative features of the object and improving the efficiency in training. Besides, a self-attention mechanism under a residual learning structure was included to enhance the global correlation in the reconstructed phase images. We evaluated the proposed \textit{SiSPRNet} both by computer simulation and on an optical experimentation platform (with defocusing to mitigate the saturation problem) and compared it with a number of existing deep neural network approaches.From the simulation and experiment results we presented, we can conclude that the proposed SiSPRNet consistently outperforms other deep neural network structures in reconstructing phase-only images and images with linearly related magnitude and phase. It is a promising solution to practical phase retrieval applications. At the moment, we only conduct simulations and experiments on synthesized objects. Further research is underway to enhance the network structure for working with more complex image datasets and realistic objects. 
    
        \section*{Appendix}  \label{Sec:Moreresults}
    
\begin{figure}[htb]
    	\begin{adjustbox}{width=1\textwidth,center}        \begin{tabular}{c@{\extracolsep{0em}}c@{\extracolsep{0em}}c@{\extracolsep{0em}}c@{\extracolsep{0em}}c@{\extracolsep{0em}}c@{\extracolsep{0em}}c@{\extracolsep{0em}}c@{\extracolsep{0em}}c@{\extracolsep{0em}}c@{\extracolsep{0em}}c@{\extracolsep{0em}}}
    			
    			\includegraphics[width=0.1\textwidth, valign=t]{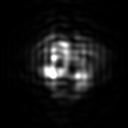}~
    			&\includegraphics[width=0.1\textwidth, valign=t]{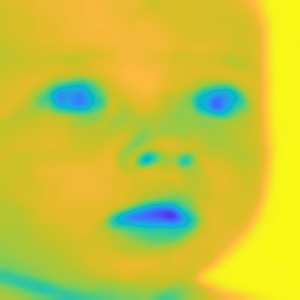}~
    			&\includegraphics[width=0.1\textwidth, valign=t]{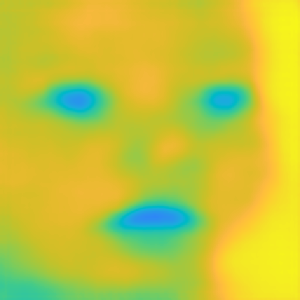}~
    			&\includegraphics[width=0.1\textwidth, valign=t]{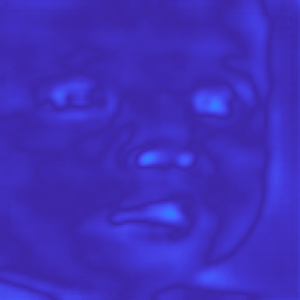}~
    			&\includegraphics[width=0.1\textwidth, valign=t]{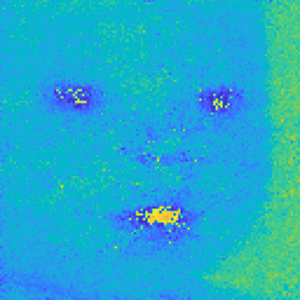}~
    			&\includegraphics[width=0.1\textwidth, valign=t]{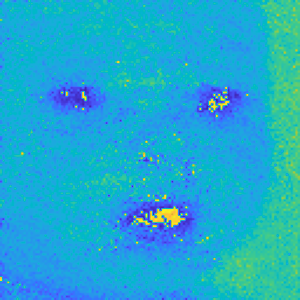}~
    			&\includegraphics[width=0.1\textwidth, valign=t]{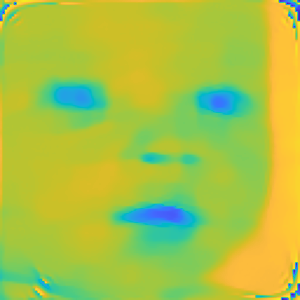}~
    			&\includegraphics[width=0.1\textwidth, valign=t]{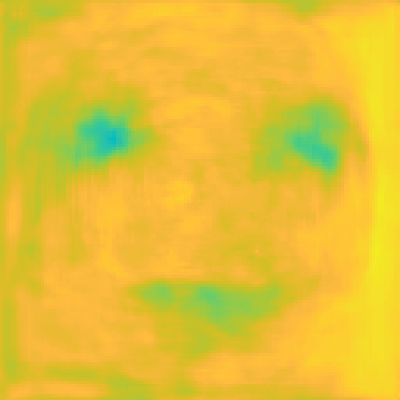}~
    			&\includegraphics[width=0.1\textwidth, valign=t]{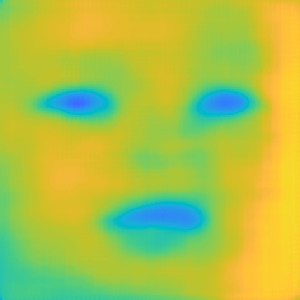}~
    			&\includegraphics[width=0.1\textwidth, valign=t]{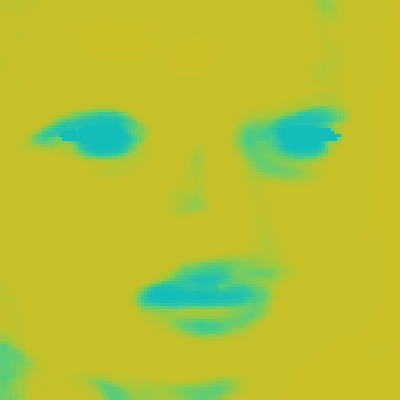}~
    			&\includegraphics[width=0.1\textwidth, valign=t]{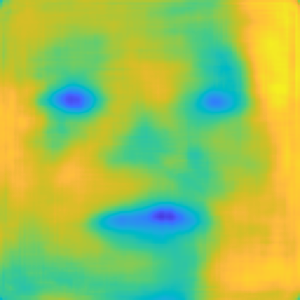}
    			
    			\\
    			
    			\includegraphics[width=0.1\textwidth, valign=t]{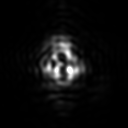}~
    			&\includegraphics[width=0.1\textwidth, valign=t]{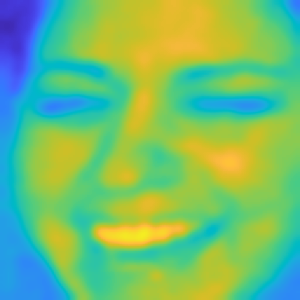}~
    			&\includegraphics[width=0.1\textwidth, valign=t]{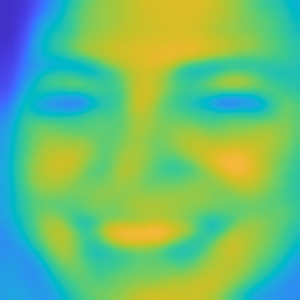}~
    			&\includegraphics[width=0.1\textwidth, valign=t]{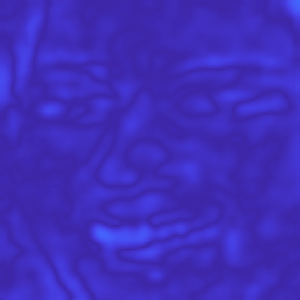}~
    			&\includegraphics[width=0.1\textwidth, valign=t]{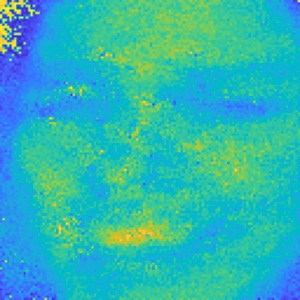}~
    			&\includegraphics[width=0.1\textwidth, valign=t]{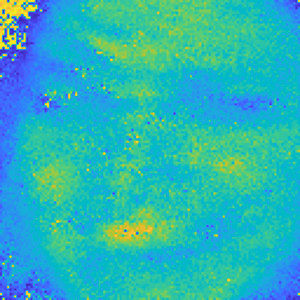}~
    			&\includegraphics[width=0.1\textwidth, valign=t]{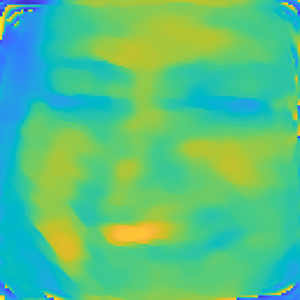}~
    			&\includegraphics[width=0.1\textwidth, valign=t]{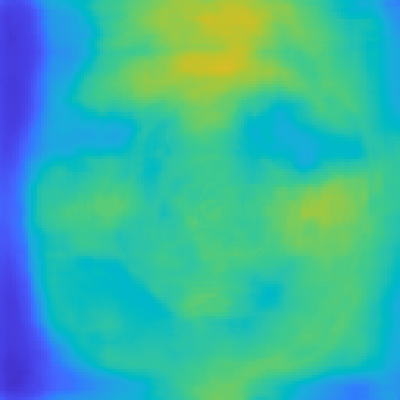}~
    			&\includegraphics[width=0.1\textwidth, valign=t]{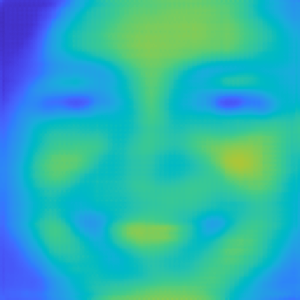}~
    			&\includegraphics[width=0.1\textwidth, valign=t]{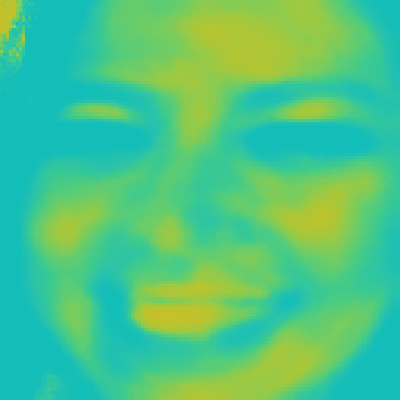}~
    			&\includegraphics[width=0.1\textwidth, valign=t]{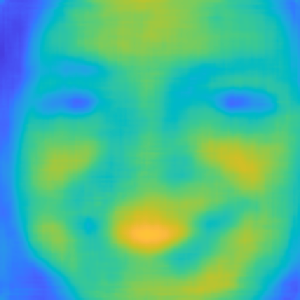}
    			\\
    			
    			\includegraphics[width=0.1\textwidth, valign=t]{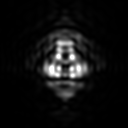}~
    			&\includegraphics[height=0.118\textwidth,width=0.1\textwidth, valign=t]{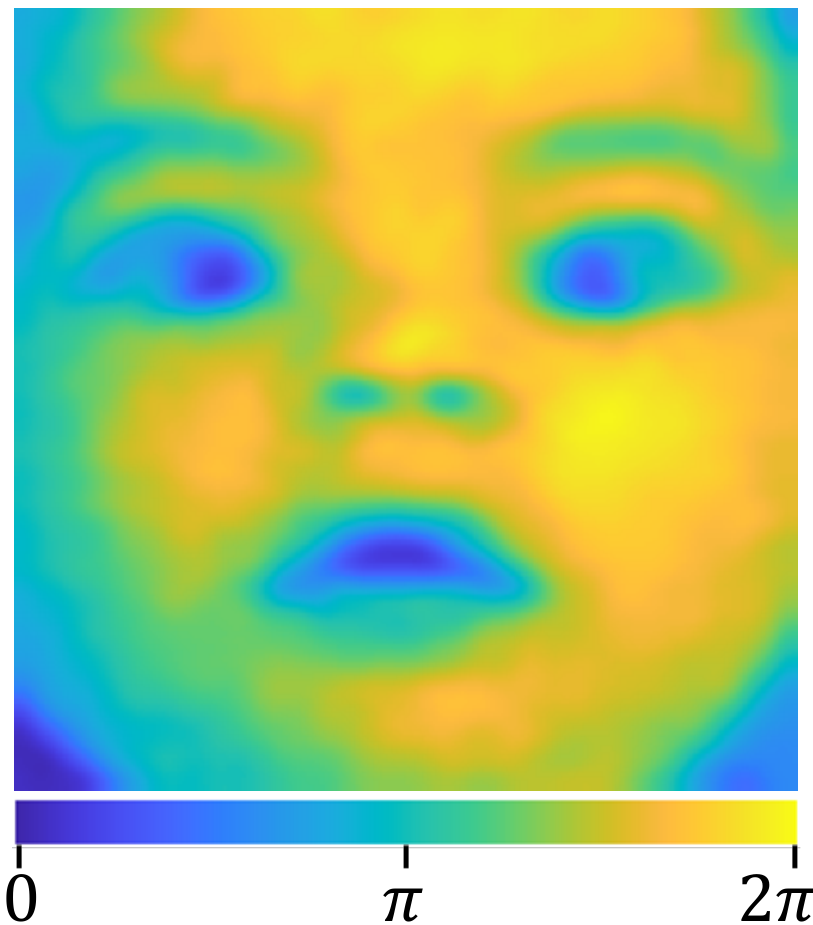}~
    			&\includegraphics[width=0.1\textwidth, valign=t]{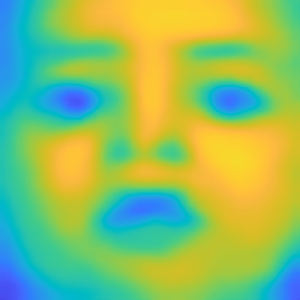}~
    			&\includegraphics[width=0.1\textwidth, valign=t]{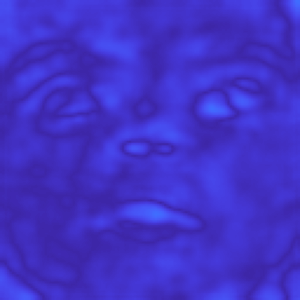}~
    			&\includegraphics[width=0.1\textwidth, valign=t]{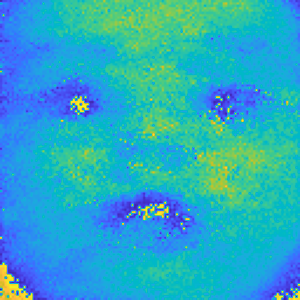}~
    			&\includegraphics[width=0.1\textwidth, valign=t]{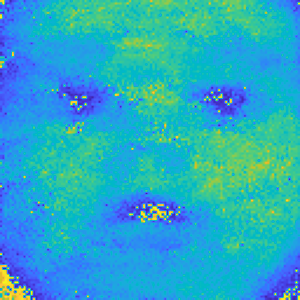}~
    			&\includegraphics[width=0.1\textwidth, valign=t]{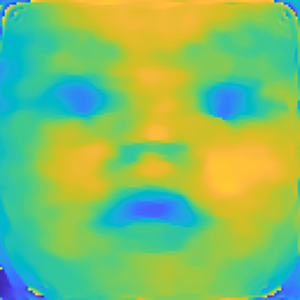}~
    			&\includegraphics[width=0.1\textwidth, valign=t]{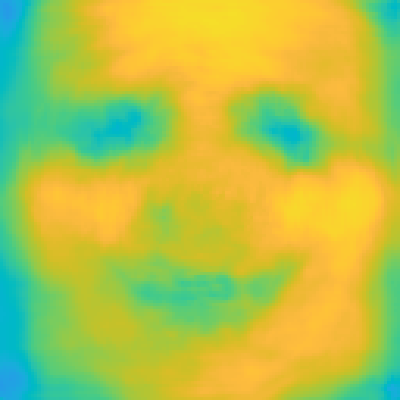}~
    			&\includegraphics[width=0.1\textwidth, valign=t]{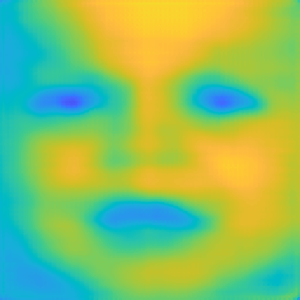}~
    			&\includegraphics[width=0.1\textwidth, valign=t]{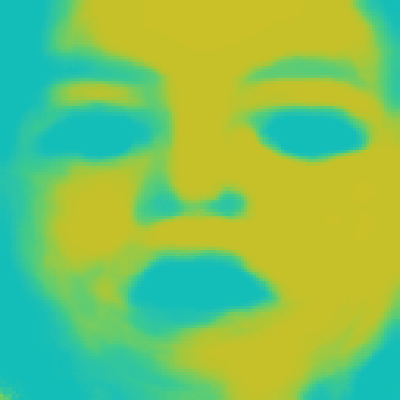}~
    			&\includegraphics[width=0.1\textwidth, valign=t]{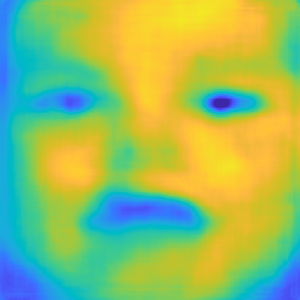}
    			\\
    			
    			{\scriptsize \makecell[c]{Fourier\\measurement}} & {\scriptsize \makecell[c]{Ground-truth \\ (phase)}} & {\scriptsize \makecell[c]{\textbf{ \textit{SiSPRNet}}\\(ours)} }& {\scriptsize \makecell[c]{\textbf{ \textit{SiSPRNet} } \\(Error Map)}}  & {\scriptsize GS } & {\scriptsize HIO} & {\scriptsize ADMM-TV} & {\scriptsize ResNet}
    			& {\scriptsize LenslessNet}  & {\scriptsize PRCGAN}  & {\scriptsize NNPhase}  
    			\\
    		\end{tabular}
    	\end{adjustbox}
    	\vspace{-0.2cm}
    	\caption{Simulation results of different phase retrieval methods on the RAF dataset \cite{li2017reliable}. The first \textbf{column} denotes the Fourier intensity measurements (pixel values: $0 \rightarrow 4095$). The second \textbf{column} shows the corresponding phase parts of the complex-valued images. The fourth \textbf{column} denotes the error maps of the  \textit{SiSPRNet}. The other \textbf{columns} present the reconstructed images through different methods.  Except for the Fourier intensity measurements (the first \textbf{column}), the colormap of the rest \textbf{columns} ranges from $0$ to $2\pi$. Each \textbf{row} presents the simulation results of a testing image.  Please zoom in for better view.}
    	\label{fig:firstimage}
    	
    \end{figure}

\begin{figure}[htb]
    	\begin{adjustbox}{width=1\textwidth,center}          \begin{tabular}{c@{\extracolsep{0em}}c@{\extracolsep{0em}}c@{\extracolsep{0em}}c@{\extracolsep{0em}}c@{\extracolsep{0em}}c@{\extracolsep{0em}}c@{\extracolsep{0em}}c@{\extracolsep{0em}}c@{\extracolsep{0em}}c@{\extracolsep{0em}}c@{\extracolsep{0em}}}
    			
    			\includegraphics[width=0.1\textwidth, valign=t]{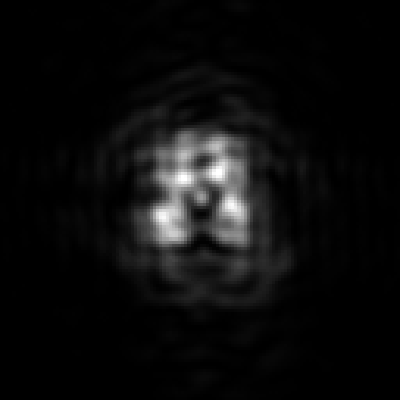}~
    			&\includegraphics[width=0.1\textwidth, valign=t]{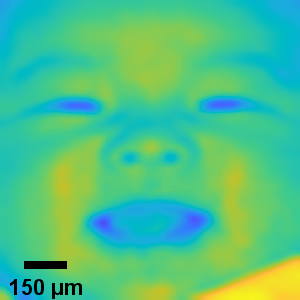}~
    			&\includegraphics[width=0.1\textwidth, valign=t]{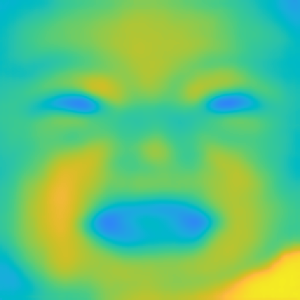}~
    			&\includegraphics[width=0.1\textwidth, valign=t]{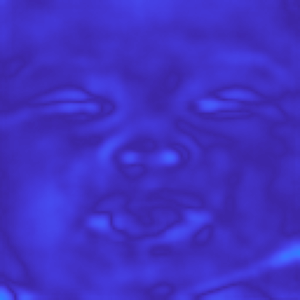}~
    			&\includegraphics[width=0.1\textwidth, valign=t]{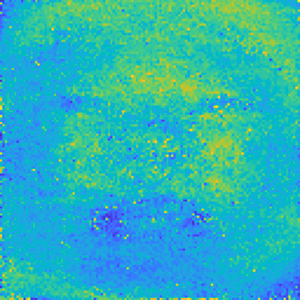}~
    			&\includegraphics[width=0.1\textwidth, valign=t]{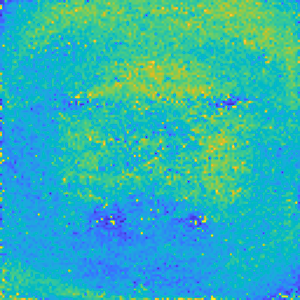}~
    			&\includegraphics[width=0.1\textwidth, valign=t]{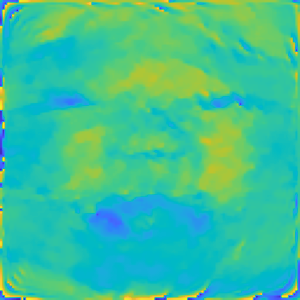}~
    			&\includegraphics[width=0.1\textwidth, valign=t]{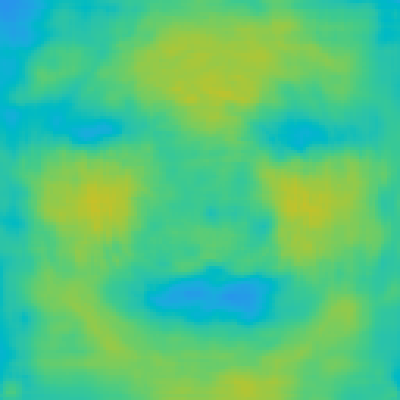}~
    			&\includegraphics[width=0.1\textwidth, valign=t]{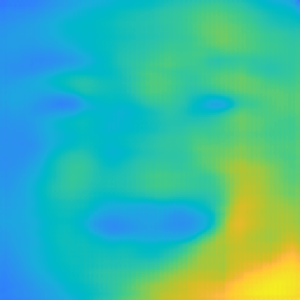}~
    			&\includegraphics[width=0.1\textwidth, valign=t]{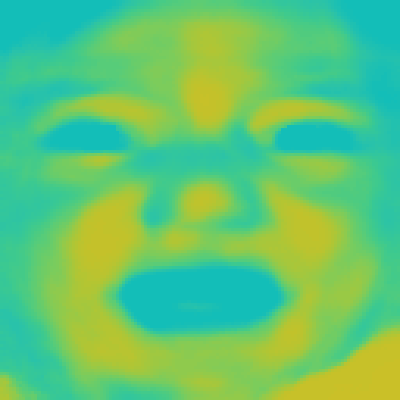}~
    			&\includegraphics[width=0.1\textwidth, valign=t]{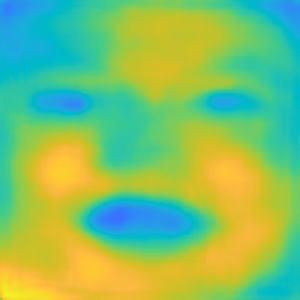}
    			\\
    			
    			\includegraphics[width=0.1\textwidth, valign=t]{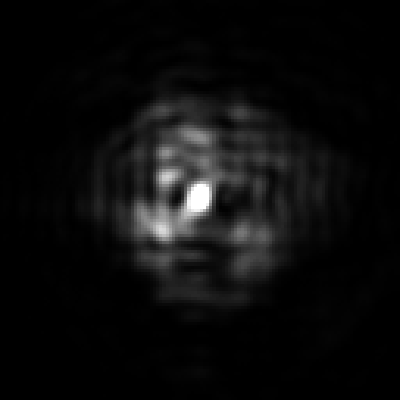}~
    			&\includegraphics[width=0.1\textwidth, valign=t]{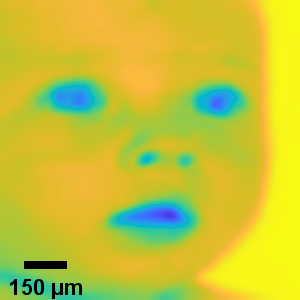}~
    			&\includegraphics[width=0.1\textwidth, valign=t]{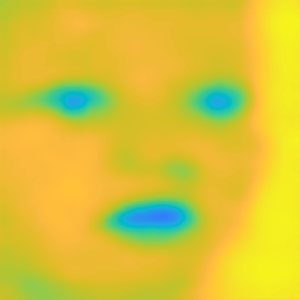}~
    			&\includegraphics[width=0.1\textwidth, valign=t]{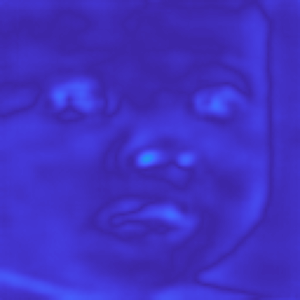}~
    			&\includegraphics[width=0.1\textwidth, valign=t]{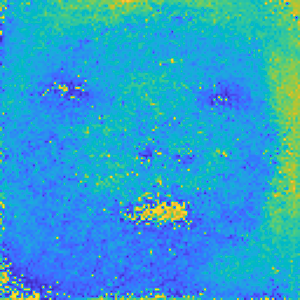}~
    			&\includegraphics[width=0.1\textwidth, valign=t]{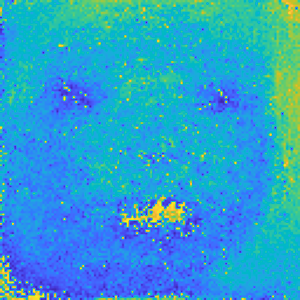}~
    			&\includegraphics[width=0.1\textwidth, valign=t]{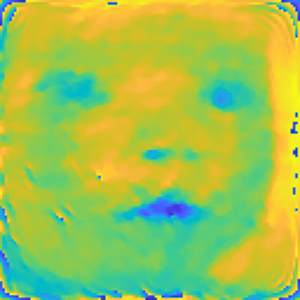}~
    			&\includegraphics[width=0.1\textwidth, valign=t]{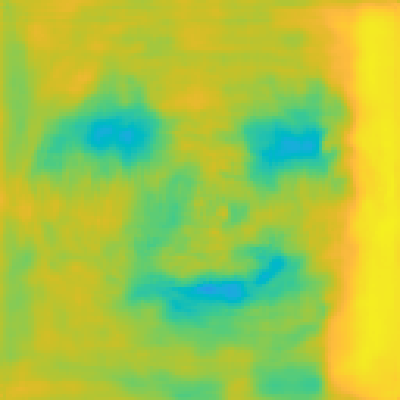}~
    			&\includegraphics[width=0.1\textwidth, valign=t]{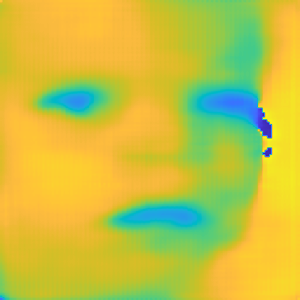}~
    			&\includegraphics[width=0.1\textwidth, valign=t]{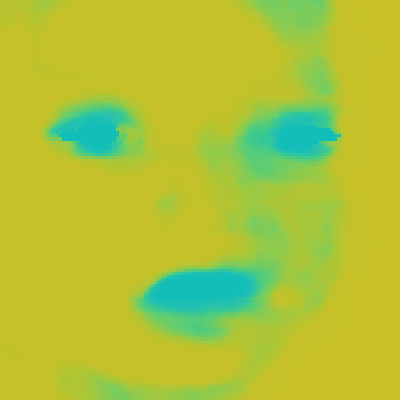}~
    			&\includegraphics[width=0.1\textwidth, valign=t]{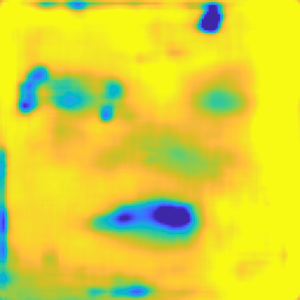}
    			\\
    			
    			\includegraphics[width=0.1\textwidth, valign=t]{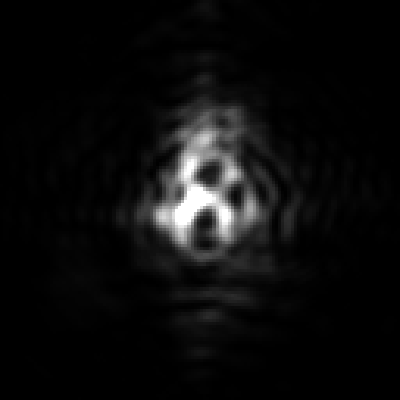}~
    			&\includegraphics[height=0.118\textwidth, width=0.1\textwidth, valign=t]{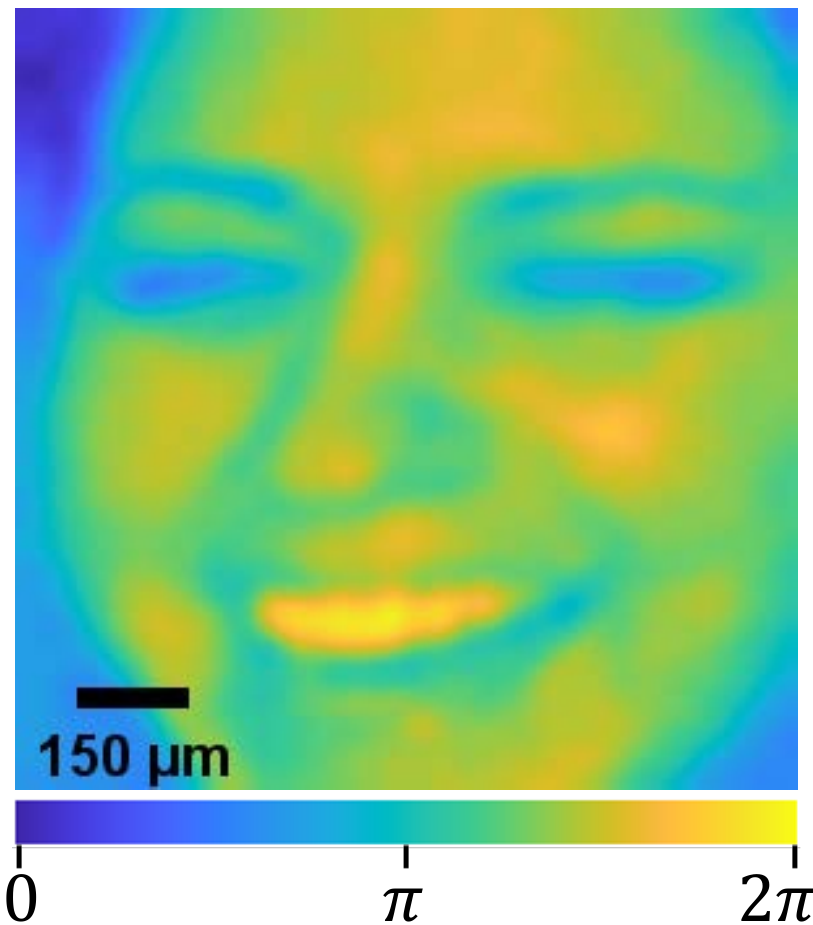}~
    			&\includegraphics[width=0.1\textwidth, valign=t]{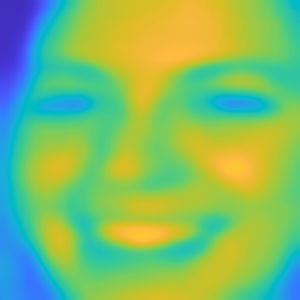}~
    			&\includegraphics[width=0.1\textwidth, valign=t]{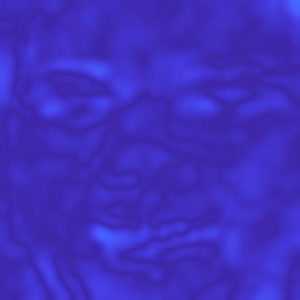}~
    			&\includegraphics[width=0.1\textwidth, valign=t]{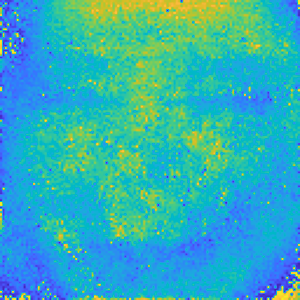}~
    			&\includegraphics[width=0.1\textwidth, valign=t]{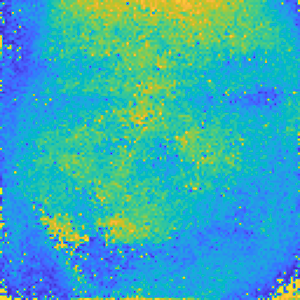}~
    			&\includegraphics[width=0.1\textwidth, valign=t]{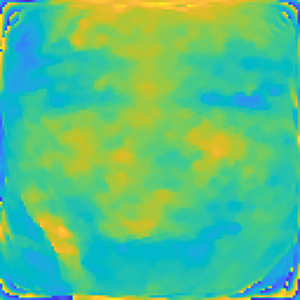}~
    			&\includegraphics[width=0.1\textwidth, valign=t]{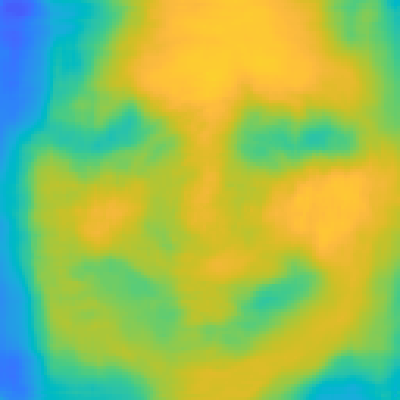}~
    			&\includegraphics[width=0.1\textwidth, valign=t]{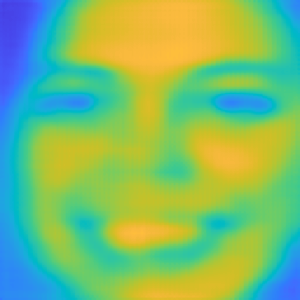}~
    			&\includegraphics[width=0.1\textwidth, valign=t]{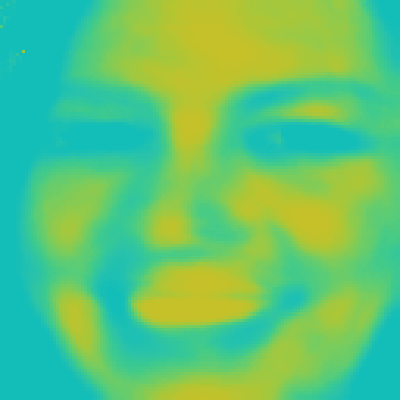}~
    			&\includegraphics[width=0.1\textwidth, valign=t]{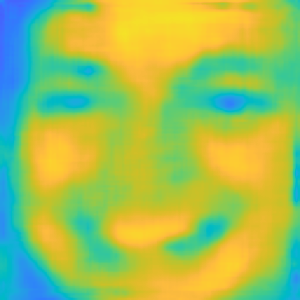}
    			\\
    			
    			{\scriptsize \makecell[c]{Fourier\\measurement}} & {\scriptsize \makecell[c]{Ground-truth \\ (phase)}} & {\scriptsize \makecell[c]{\textbf{ \textit{SiSPRNet}}\\(ours)} }& {\scriptsize \makecell[c]{\textbf{ \textit{SiSPRNet}} \\(Error Map)} }  & {\scriptsize GS } & {\scriptsize HIO} & {\scriptsize ADMM-TV} & {\scriptsize ResNet}
    			& {\scriptsize LenslessNet}  & {\scriptsize PRCGAN}  & {\scriptsize NNPhase}  
    			\\
    		\end{tabular}
    	\end{adjustbox}
    	\vspace{-0.2cm}
    	\caption{Experimental results of different phase retrieval methods on the RAF dataset \cite{li2017reliable}. The first \textbf{column} denotes the Fourier intensity measurements (pixel values: $0 \rightarrow 4095$). The second \textbf{column} shows the corresponding phase parts of the complex-valued images with scale bars at the bottom left corners.  The fourth \textbf{column} denotes the error maps of \textit{SiSPRNet}. The other \textbf{columns} present the reconstructed images through different methods. Except for the Fourier intensity measurements (the first \textbf{column}), the colormap of the rest \textbf{columns} ranges from $0$ to $2\pi$. Each \textbf{row} presents the experimental results of a testing image. Please zoom in for better view.}
    	\label{fig:secondimage}
    	
    \end{figure}

\begin{figure}[htb]
    	\begin{adjustbox}{width=1\textwidth, center}          \begin{tabular}{c@{\extracolsep{0em}}c@{\extracolsep{0em}}c@{\extracolsep{0em}}c@{\extracolsep{0em}}c@{\extracolsep{0em}}c@{\extracolsep{0em}}c@{\extracolsep{0em}}c@{\extracolsep{0em}}c@{\extracolsep{0em}}c@{\extracolsep{0em}}c@{\extracolsep{0em}}}
    			
    			\includegraphics[width=0.1\textwidth, valign=t]{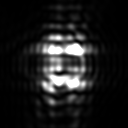}~
    			&\includegraphics[width=0.1\textwidth, valign=t]{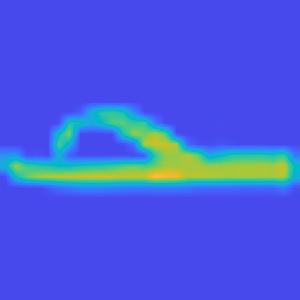}~
    			&\includegraphics[width=0.1\textwidth, valign=t]{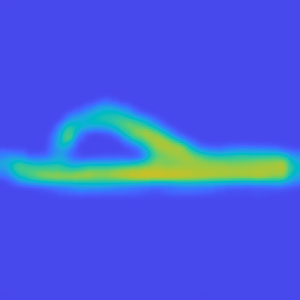}~
    			&\includegraphics[width=0.1\textwidth, valign=t]{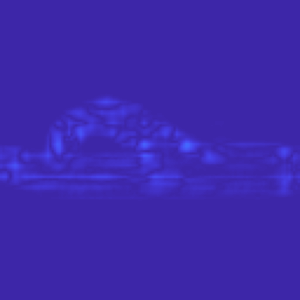}~
    			&\includegraphics[width=0.1\textwidth, valign=t]{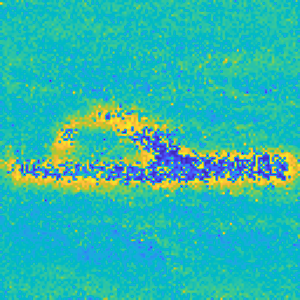}~
    			&\includegraphics[width=0.1\textwidth, valign=t]{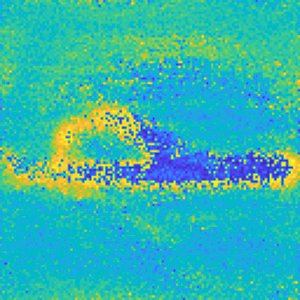}~
    			&\includegraphics[width=0.1\textwidth, valign=t]{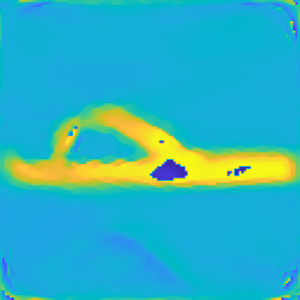}~
    			&\includegraphics[width=0.1\textwidth, valign=t]{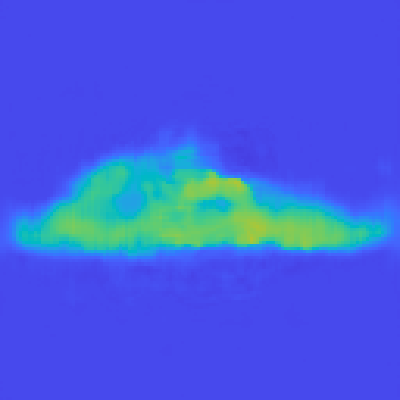}~
    			&\includegraphics[width=0.1\textwidth, valign=t]{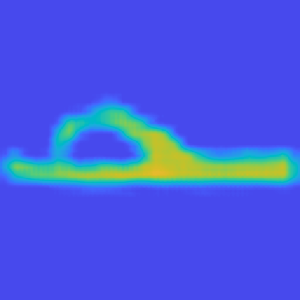}~
    			&\includegraphics[width=0.1\textwidth, valign=t]{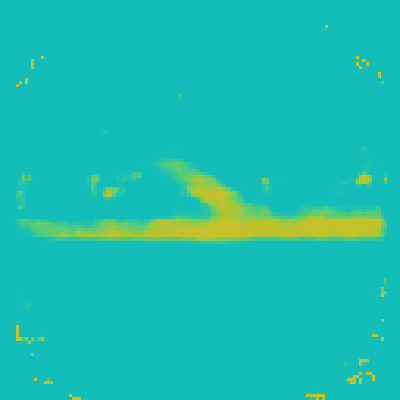}~
    			&\includegraphics[width=0.1\textwidth, valign=t]{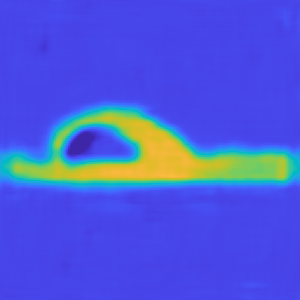}
    			\\
    			
    			\includegraphics[width=0.1\textwidth, valign=t]{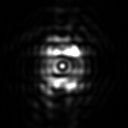}~
    			&\includegraphics[width=0.1\textwidth, valign=t]{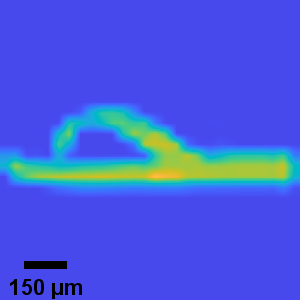}~
    			&\includegraphics[width=0.1\textwidth, valign=t]{Figures/third/SiPRNet_img_0106_angle_gt_exp.png}~
    			&\includegraphics[width=0.1\textwidth, valign=t]{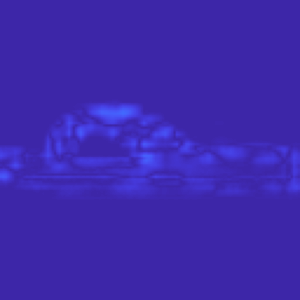}~
    			&\includegraphics[width=0.1\textwidth, valign=t]{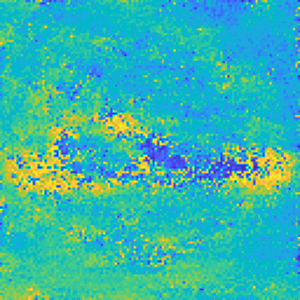}~
    			&\includegraphics[width=0.1\textwidth, valign=t]{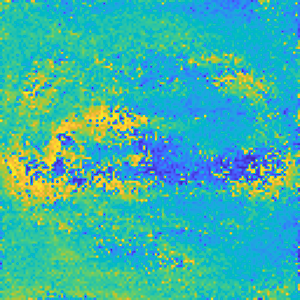}~
    			&\includegraphics[width=0.1\textwidth, valign=t]{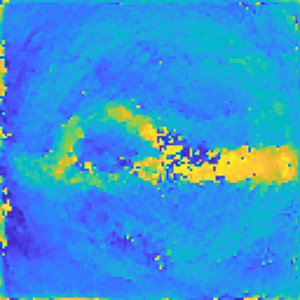}~
    			&\includegraphics[width=0.1\textwidth, valign=t]{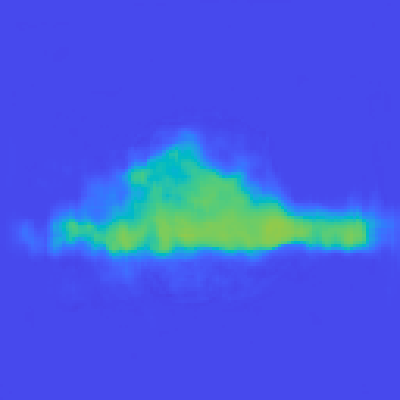}~
    			&\includegraphics[width=0.1\textwidth, valign=t]{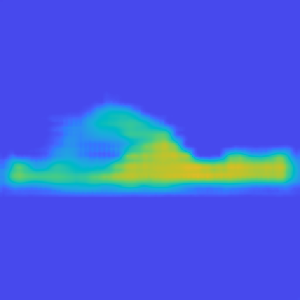}~
    			&\includegraphics[width=0.1\textwidth, valign=t]{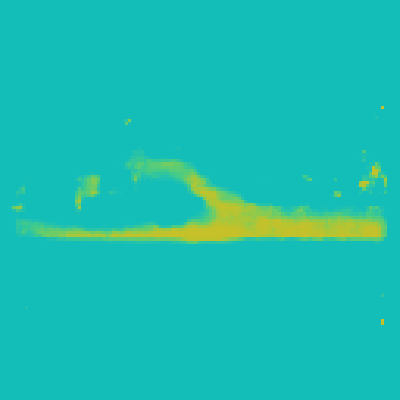}~
    			&\includegraphics[width=0.1\textwidth, valign=t]{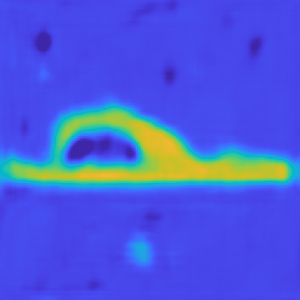}
    			\\
    			
    			\includegraphics[width=0.1\textwidth, valign=t]{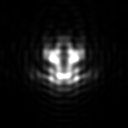}~
    			&\includegraphics[width=0.1\textwidth, valign=t]{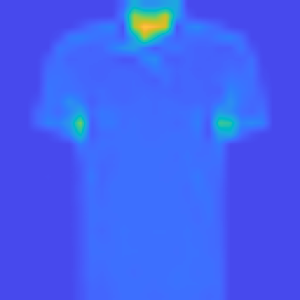}~
    			&\includegraphics[width=0.1\textwidth, valign=t]{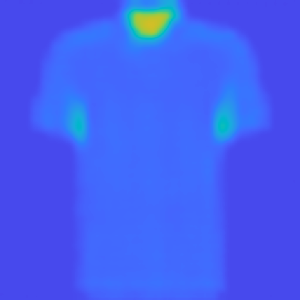}~
    			&\includegraphics[width=0.1\textwidth, valign=t]{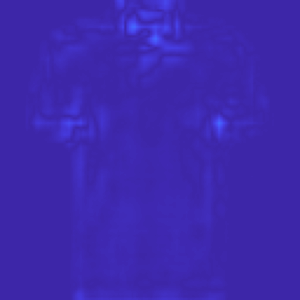}~
    			&\includegraphics[width=0.1\textwidth, valign=t]{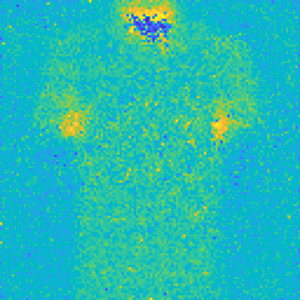}~
    			&\includegraphics[width=0.1\textwidth, valign=t]{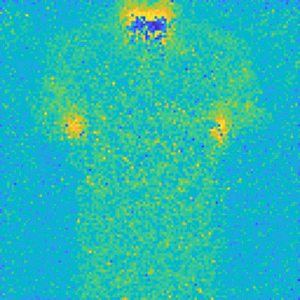}~
    			&\includegraphics[width=0.1\textwidth, valign=t]{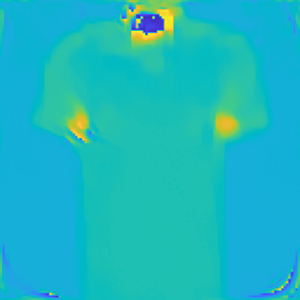}~
    			&\includegraphics[width=0.1\textwidth, valign=t]{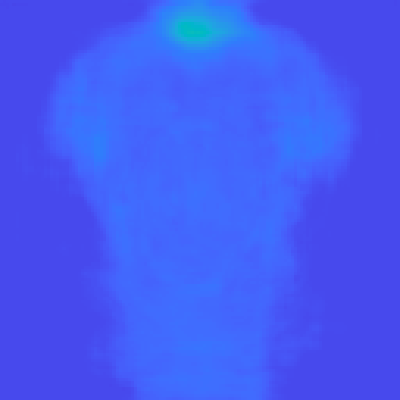}~
    			&\includegraphics[width=0.1\textwidth, valign=t]{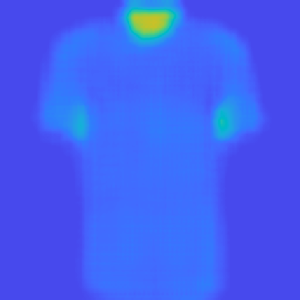}~
    			&\includegraphics[width=0.1\textwidth, valign=t]{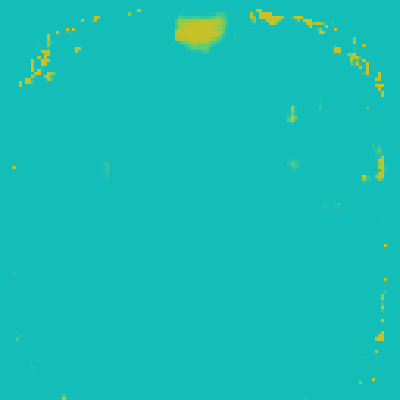}~
    			&\includegraphics[width=0.1\textwidth, valign=t]{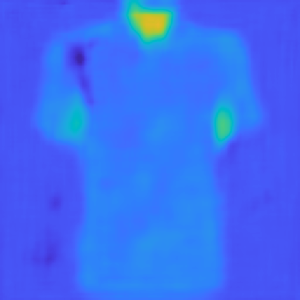}
    			\\
    			
    			\includegraphics[width=0.1\textwidth, valign=t]{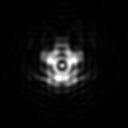}~
    			&\includegraphics[height=0.118\textwidth,width=0.1\textwidth, valign=t]{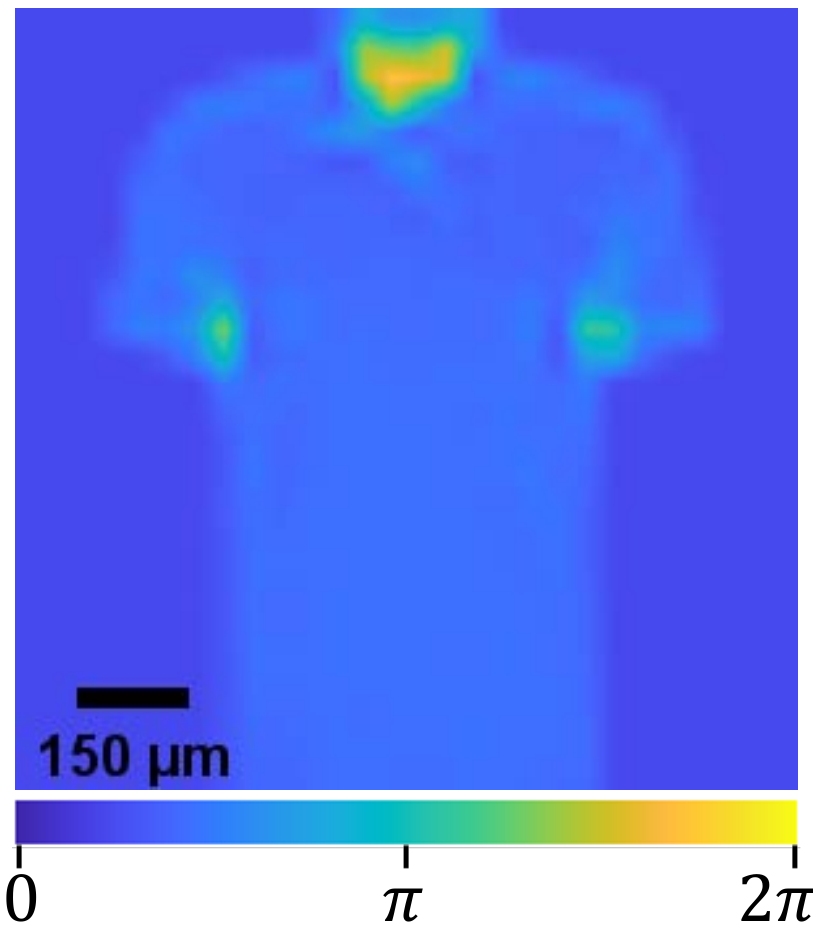}~
    			&\includegraphics[width=0.1\textwidth, valign=t]{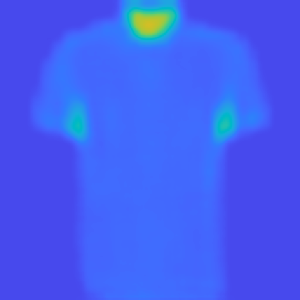}~
    			&\includegraphics[width=0.1\textwidth, valign=t]{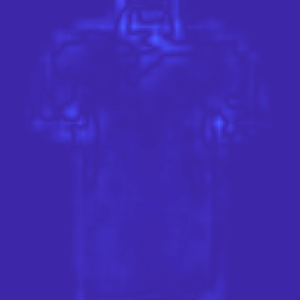}~
    			&\includegraphics[width=0.1\textwidth, valign=t]{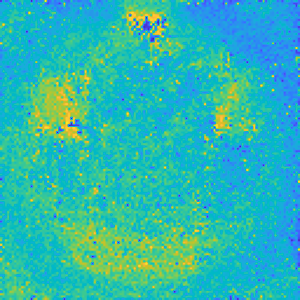}~
    			&\includegraphics[width=0.1\textwidth, valign=t]{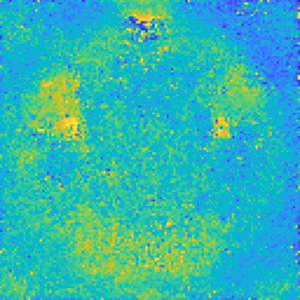}~
    			&\includegraphics[width=0.1\textwidth, valign=t]{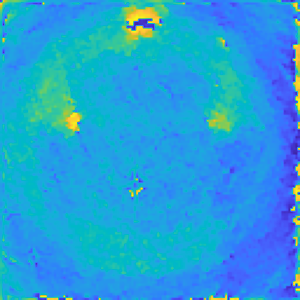}~
    			&\includegraphics[width=0.1\textwidth, valign=t]{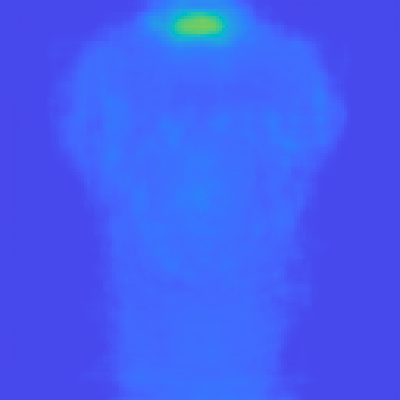}~
    			&\includegraphics[width=0.1\textwidth, valign=t]{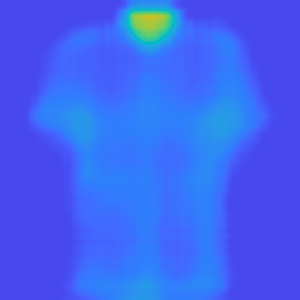}~
    			&\includegraphics[width=0.1\textwidth, valign=t]{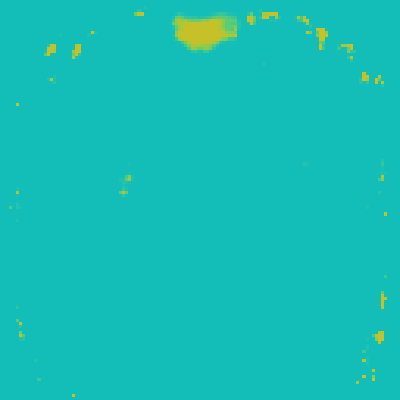}~
    			&\includegraphics[width=0.1\textwidth, valign=t]{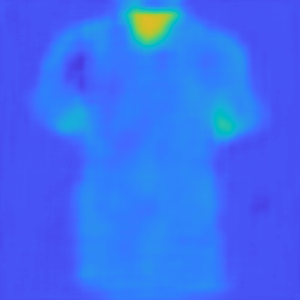}
    			\\
    			
    			{\scriptsize \makecell[c]{Fourier\\measurement}} & {\scriptsize \makecell[c]{Ground-truth \\ (phase)}} & {\scriptsize \makecell[c]{\textbf{ \textit{SiSPRNet}}\\(ours)} }& {\scriptsize \makecell[c]{\textbf{ \textit{SiSPRNet}} \\(Error Map)} }  & {\scriptsize GS } & {\scriptsize HIO} & {\scriptsize ADMM-TV} & {\scriptsize ResNet}
    			& {\scriptsize LenslessNet}  & {\scriptsize PRCGAN}  & {\scriptsize NNPhase}  
    			\\
    			
    		\end{tabular}
    	\end{adjustbox}
    	\vspace{-0.2cm}
    	\caption{Simulation and experimental results of different phase retrieval methods on the Fashion-MNIST dataset \cite{xiao2017/online}. The first \textbf{column} denotes the Fourier intensity measurements (pixel values: $0 \rightarrow 4095$). The second \textbf{column} shows the corresponding phase parts of the complex-valued images with scale bars (for experimental results) at the bottom left corners. The fourth \textbf{column} denotes the error maps of \textit{SiSPRNet}. The other \textbf{columns} present the reconstruction images through different methods. Except for the Fourier intensity measurements (the first \textbf{column}), the colormap of the rest \textbf{columns} ranges from $0$ to $2\pi$. The first and second \textbf{rows} present the simulation and experimental results of the same image. The third and the fourth \textbf{rows} show the simulation and experimental results of another image. Please zoom in for better view.}
    	\label{fig:thirdimage}
    	
    \end{figure}

    In this Appendix, we show additional qualitative comparison results.  Fig. \ref{fig:firstimage}, \ref{fig:secondimage}, \ref{fig:thirdimage} present the simulation and experimental results. The detailed explanations are attached with each figure.
    
    \section*{Data availability} The datasets used in this paper are available in  \cite{li2017reliable, xiao2017/online}. Data underlying the results presented in this paper are not publicly available at this time but may be obtained from the authors upon reasonable request.
   	
   	\section*{Funding} General Research Fund (PolyU 15225321), Centre for Advances in Reliability and Safety (CAiRS)
    
    \section*{Acknowledgments} This work was supported by the Hong Kong Research Grant Council under General Research Fund no. PolyU 15225321, and the Centre for Advances in Reliability and Safety (CAiRS) admitted under AIR@InnoHK Research Cluster.
    
    \section*{Disclosures} The authors declare no conflicts of interest.

	\bibliography{Reference_All}

\begin{thebibliography}{10}
\newcommand{\enquote}[1]{``#1''}

\bibitem{Ye_SiSPRNet}
Q.~Ye, L.-W. Wang, and D.~P.-K. Lun, \enquote{{SiSPRNet: End-to-End Learning
  for Single-Shot Phase Retrieval},}
  \url{https://github.com/Qiustander/SiPRNet} (2022).

\bibitem{Gerchberg1972APA}
R.~W. Gerchberg, \enquote{A practical algorithm for the determination of phase
  from image and diffraction plane pictures,} {\protect\JournalTitle{Optik}}
  \textbf{35}, 237--246 (1972).

\bibitem{Fienup:82}
J.~R. Fienup, \enquote{Phase retrieval algorithms: a comparison,}
  {\protect\JournalTitle{Appl. Opt.}} \textbf{21}, 2758--2769 (1982).

\bibitem{rodenburg2008ptychography}
J.~M. Rodenburg, \enquote{Ptychography and related diffractive imaging
  methods,} {\protect\JournalTitle{Advances in Imaging and Electron Physics}}
  \textbf{150}, 87--184 (2008).

\bibitem{Anand07}
A.~Anand, G.~Pedrini, W.~Osten, and P.~Almoro, \enquote{Wavefront sensing with
  random amplitude mask and phase retrieval,} {\protect\JournalTitle{Optics
  Letters}} \textbf{32}, 1584--1586 (2007).

\bibitem{Horisaki2014SingleshotPI}
R.~Horisaki, Y.~Ogura, M.~Aino, and J.~Tanida, \enquote{Single-shot phase
  imaging with a coded aperture,} {\protect\JournalTitle{Optics letters}}
  \textbf{39 22}, 6466--9 (2014).

\bibitem{candes2015phase}
E.~J. Candes, X.~Li, and M.~Soltanolkotabi, \enquote{Phase retrieval via
  wirtinger flow: Theory and algorithms,} {\protect\JournalTitle{IEEE
  Transactions on Information Theory}} \textbf{61}, 1985--2007 (2015).

\bibitem{YE2022106808}
Q.~Ye, Y.-H. Chan, M.~G. Somekh, and D.~P. Lun, \enquote{Robust phase retrieval
  with green noise binary masks,} {\protect\JournalTitle{Optics and Lasers in
  Engineering}} \textbf{149}, 106808 (2022).

\bibitem{Zheng2017DigitalMD}
C.~Zheng, R.~Zhou, C.~Kuang, G.~Zhao, Z.~Yaqoob, and P.~So, \enquote{Digital
  micromirror device-based common-path quantitative phase imaging,}
  {\protect\JournalTitle{Optics Letters}} \textbf{42 7}, 1448--1451 (2017).

\bibitem{candes2015code}
E.~J. Candes, X.~Li, and M.~Soltanolkotabi, \enquote{Phase retrieval from coded
  diffraction patterns,} {\protect\JournalTitle{Applied and Computational
  Harmonic Analysis}} \textbf{39}, 277--299 (2015).

\bibitem{LeCun_2015}
Y.~LeCun, Y.~Bengio, and G.~Hinton, \enquote{Deep learning,}
  {\protect\JournalTitle{Nature}} \textbf{521}, 436--444 (2015).

\bibitem{He_2016_CVPR}
K.~He, X.~Zhang, S.~Ren, and J.~Sun, \enquote{Deep residual learning for image
  recognition,} in \emph{Proceedings of the IEEE Conference on Computer Vision
  and Pattern Recognition (CVPR),}  (2016).

\bibitem{zhang2017beyond}
K.~Zhang, W.~Zuo, Y.~Chen, D.~Meng, and L.~Zhang, \enquote{Beyond a gaussian
  denoiser: Residual learning of deep cnn for image denoising,}
  {\protect\JournalTitle{IEEE transactions on image processing}} \textbf{26},
  3142--3155 (2017).

\bibitem{wang2020lightening}
L.-W. Wang, Z.-S. Liu, W.-C. Siu, and D.~P. Lun, \enquote{Lightening network
  for low-light image enhancement,} {\protect\JournalTitle{IEEE Transactions on
  Image Processing}} \textbf{29}, 7984--7996 (2020).

\bibitem{attention_17}
A.~Vaswani, N.~Shazeer, N.~Parmar, J.~Uszkoreit, L.~Jones, A.~N. Gomez, L.~u.
  Kaiser, and I.~Polosukhin, \enquote{Attention is all you need,} in
  \emph{Advances in Neural Information Processing Systems,}  vol.~30 I.~Guyon,
  U.~V. Luxburg, S.~Bengio, H.~Wallach, R.~Fergus, S.~Vishwanathan, and
  R.~Garnett, eds. (Curran Associates, Inc., 2017).

\bibitem{deng_li_goy_kang_barbastathis_2020}
M.~Deng, S.~Li, A.~Goy, I.~Kang, and G.~Barbastathis, \enquote{Learning to
  synthesize: robust phase retrieval at low photon counts,}
  {\protect\JournalTitle{Light: Science \& Applications}} \textbf{9} (2020).

\bibitem{White:19}
J.~White and Z.~Chang, \enquote{Attosecond streaking phase retrieval with
  neural network,} {\protect\JournalTitle{Optics Express}} \textbf{27},
  4799--4807 (2019).

\bibitem{Shi:19}
J.~Shi, X.~Zhu, H.~Wang, L.~Song, and Q.~Guo, \enquote{Label enhanced and patch
  based deep learning for phase retrieval from single frame fringe pattern in
  fringe projection 3d measurement,} {\protect\JournalTitle{Opt. Express}}
  \textbf{27}, 28929--28943 (2019).

\bibitem{Sinha17}
A.~Sinha, J.~Lee, S.~Li, and G.~Barbastathis, \enquote{Lensless computational
  imaging through deep learning,} {\protect\JournalTitle{Optica}} \textbf{4},
  1117--1125 (2017).

\bibitem{Bai:19}
C.~Bai, M.~Zhou, J.~Min, S.~Dang, X.~Yu, P.~Zhang, T.~Peng, and B.~Yao,
  \enquote{Robust contrast-transfer-function phase retrieval via flexible deep
  learning networks,} {\protect\JournalTitle{Optics Letters}} \textbf{44},
  5141--5144 (2019).

\bibitem{Kumar:21}
S.~Kumar, \enquote{Phase retrieval with physics informed zero-shot network,}
  {\protect\JournalTitle{Optics Letters}} \textbf{46}, 5942--5945 (2021).

\bibitem{Uelwer_PhaseRetrieval}
T.~Uelwer, T.~Hoffmann, and S.~Harmeling, \enquote{Non-iterative phase
  retrieval with cascaded neural networks,} in \emph{Artificial Neural Networks
  and Machine Learning -- ICANN 2021,}  I.~Farka{\v{s}}, P.~Masulli, S.~Otte,
  and S.~Wermter, eds. (Springer International Publishing, Cham, 2021), pp.
  295--306.

\bibitem{cond_gan_14}
M.~Mirza and S.~Osindero, \enquote{Conditional generative adversarial nets,}
  (2014).

\bibitem{uelwer2021phase}
T.~Uelwer, A.~Oberstra$\beta$, and S.~Harmeling, \enquote{Phase retrieval using
  conditional generative adversarial networks,} in \emph{2020 25th
  International Conference on Pattern Recognition (ICPR),}  (IEEE, 2021), pp.
  731--738.

\bibitem{Wu_cw5029}
L.~Wu, P.~Juhas, S.~Yoo, and I.~Robinson, \enquote{{Complex imaging of phase
  domains by deep neural networks},} {\protect\JournalTitle{IUCrJ}} \textbf{8},
  12--21 (2021).

\bibitem{Wu_2021}
L.~Wu, S.~Yoo, A.~F. Suzana, T.~A. Assefa, J.~Diao, R.~J. Harder, W.~Cha, and
  I.~K. Robinson, \enquote{Three-dimensional coherent x-ray diffraction imaging
  via deep convolutional neural networks,} {\protect\JournalTitle{npj
  Computational Materials}} \textbf{7} (2021).

\bibitem{Fus_18}
F.~Fus, Y.~Yang, A.~Pacureanu, S.~Bohic, and P.~Cloetens, \enquote{Unsupervised
  solution for in-line holography phase retrieval using bayesian inference,}
  {\protect\JournalTitle{Opt. Express}} \textbf{26}, 32847--32865 (2018).

\bibitem{Zhang_21}
Y.~Zhang, M.~A. Noack, P.~Vagovic, K.~Fezzaa, F.~Garcia-Moreno, T.~Ritschel,
  and P.~Villanueva-Perez, \enquote{Phasegan: a deep-learning phase-retrieval
  approach for unpaired datasets,} {\protect\JournalTitle{Opt. Express}}
  \textbf{29}, 19593--19604 (2021).

\bibitem{hayes1982reconstruction}
M.~{Hayes}, \enquote{The reconstruction of a multidimensional sequence from the
  phase or magnitude of its fourier transform,} {\protect\JournalTitle{IEEE
  Transactions on Acoustics, Speech, and Signal Processing}} \textbf{30},
  140--154 (1982).

\bibitem{goodman2017introduction}
J.~Goodman, \emph{Introduction to Fourier Optics} (W. H. Freeman, 2017).

\bibitem{srivastava2014dropout}
N.~Srivastava, G.~Hinton, A.~Krizhevsky, I.~Sutskever, and R.~Salakhutdinov,
  \enquote{Dropout: a simple way to prevent neural networks from overfitting,}
  {\protect\JournalTitle{The journal of machine learning research}}
  \textbf{15}, 1929--1958 (2014).

\bibitem{non-local}
X.~Wang, R.~Girshick, A.~Gupta, and K.~He, \enquote{Non-local neural networks,}
  in \emph{Proceedings of the IEEE conference on computer vision and pattern
  recognition,}  (2018), pp. 7794--7803.

\bibitem{Wang:21}
Y.~Wang, Z.~Lin, H.~Wang, C.~Hu, H.~Yang, and M.~Gu,
  \enquote{High-generalization deep sparse pattern reconstruction: feature
  extraction of speckles using self-attention armed convolutional neural
  networks,} {\protect\JournalTitle{Opt. Express}} \textbf{29}, 35702--35711
  (2021).

\bibitem{Wang_2017_CVPR}
F.~Wang, M.~Jiang, C.~Qian, S.~Yang, C.~Li, H.~Zhang, X.~Wang, and X.~Tang,
  \enquote{Residual attention network for image classification,} in
  \emph{Proceedings of the IEEE Conference on Computer Vision and Pattern
  Recognition (CVPR),}  (2017).

\bibitem{zhang2021explainable}
X.~Zhang, L.~Han, W.~Zhu, L.~Sun, and D.~Zhang, \enquote{An explainable 3d
  residual self-attention deep neural network for joint atrophy localization
  and alzheimer's disease diagnosis using structural mri,}
  {\protect\JournalTitle{IEEE journal of biomedical and health informatics}}
  (2021).

\bibitem{park2021dynamic}
K.~Park, J.~W. Soh, and N.~I. Cho, \enquote{Dynamic residual self-attention
  network for lightweight single image super-resolution,}
  {\protect\JournalTitle{IEEE Transactions on Multimedia}}  (2021).

\bibitem{li2017reliable}
S.~Li, W.~Deng, and J.~Du, \enquote{Reliable crowdsourcing and deep
  locality-preserving learning for expression recognition in the wild,} in
  \emph{2017 IEEE Conference on Computer Vision and Pattern Recognition
  (CVPR),}  (IEEE, 2017), pp. 2584--2593.

\bibitem{xiao2017/online}
H.~Xiao, K.~Rasul, and R.~Vollgraf, \enquote{Fashion-mnist: a novel image
  dataset for benchmarking machine learning algorithms,}  (2017).

\bibitem{kingma2017adam}
D.~P. Kingma and J.~Ba, \enquote{Adam: A method for stochastic optimization,}
  (2017).

\bibitem{Refaeilzadeh2016}
P.~Refaeilzadeh, L.~Tang, and H.~Liu, \emph{Cross-Validation} (Springer New
  York, New York, NY, 2016), pp. 1--7.

\bibitem{Stone_1974}
M.~Stone, \enquote{Cross-validatory choice and assessment of statistical
  predictions,} {\protect\JournalTitle{Journal of the Royal Statistical
  Society: Series B (Methodological)}} \textbf{36}, 111--133 (1974).

\bibitem{Nishizaki_2020}
Y.~Nishizaki, R.~Horisaki, K.~Kitaguchi, M.~Saito, and J.~Tanida,
  \enquote{Analysis of non-iterative phase retrieval based on machine
  learning,} {\protect\JournalTitle{Optical Review}} \textbf{27}, 136--141
  (2020).

\bibitem{Zhang_2018_CVPR}
Y.~Zhang, Y.~Tian, Y.~Kong, B.~Zhong, and Y.~Fu, \enquote{Residual dense
  network for image super-resolution,} in \emph{Proceedings of the IEEE
  Conference on Computer Vision and Pattern Recognition (CVPR),}  (2018).

\bibitem{chang2018total}
H.~Chang, Y.~Lou, Y.~Duan, and S.~Marchesini, \enquote{Total variation--based
  phase retrieval for poisson noise removal,} {\protect\JournalTitle{SIAM
  Journal on Imaging Sciences}} \textbf{11}, 24--55 (2018).

\end{thebibliography}

\end{document}